\documentclass[preprint,12pt]{elsarticle}

\usepackage{amssymb}
\usepackage{amsmath}
\usepackage{bbm}
\usepackage{hyperref}
\usepackage{tikz}
\usepackage{ifthen}
\usepackage{comment}
\usepackage{booktabs}
\usepackage{pifont}% http://ctan.org/pkg/pifont
\newcommand{\cmark}{\ding{51}}%
\newcommand{\xmark}{\ding{55}}%

\newcommand{\mau}{\mathbf{u}}

\newcommand{\mauh}{\mathbf{u}_h}
\newcommand{\bmauh}{\bar{\mathbf{u}}_H}
\usepackage[acronym,nonumberlist]{glossaries}

\makenoidxglossaries

\newacronym{LES}{LES}{large eddy simulation}
\newacronym{NC}{NC}{no closure}
\newacronym{FDNS}{FDNS}{filtered direct numerical simulation}
\newacronym{SMAG}{SMAG}{Smagorinsky model}
\newacronym{DIV}{DIV}{divergence of stress tensor predicted by a neural network}
\newacronym{SKEW}{SKEW}{skew-symmetric neural network architecture}
\newacronym{BC}{BC}{boundary condition}
\newacronym{PDE}{PDE}{partial differential equation}
\newacronym{CNN}{CNN}{convolutional neural network}
\newacronym{CNN-C}{CNN-C}{convolutional neural network with backscatter clipping}
\newacronym{RHS}{RHS}{right-hand side}
\newacronym{DNS}{DNS}{direct numerical simulation}
\newacronym{FFT}{FFT}{fast Fourier transform}
\newacronym{MSE}{MSE}{mean squared error}

\newcommand{\R}[1]{}
\newcommand{\revone}[1]{\textcolor{black}{#1}}
\newcommand{\revtwo}[1]{\textcolor{black}{#1}}
\newcommand{\revthree}[1]{\textcolor{black}{#1}}

\journal{Computers \& Fluids}

\begin{document}

\begin{frontmatter}

%% Title, authors and addresses

%% use the tnoteref command within \title for footnotes;
%% use the tnotetext command for theassociated footnote;
%% use the fnref command within \author or \affiliation for footnotes;
%% use the fntext command for theassociated footnote;
%% use the corref command within \author for corresponding author footnotes;
%% use the cortext command for theassociated footnote;
%% use the ead command for the email address,
%% and the form \ead[url] for the home page:
%% \title{Title\tnoteref{label1}}
%% \tnotetext[label1]{}
%% \author{Name\corref{cor1}\fnref{label2}}
%% \ead{email address}
%% \ead[url]{home page}
%% \fntext[label2]{}
%% \cortext[cor1]{}
%% \affiliation{organization={},
%%             addressline={},
%%             city={},
%%             postcode={},
%%             state={},
%%             country={}}
%% \fntext[label3]{}

\title{}

%% use optional labels to link authors explicitly to addresses:
%% \author[label1,label2]{}
%% \affiliation[label1]{organization={},
%%             addressline={},
%%             city={},
%%             postcode={},
%%             state={},
%%             country={}}
%%
%% \affiliation[label2]{organization={},
%%             addressline={},
%%             city={},
%%             postcode={},
%%             state={},
%%             country={}}

\title{Energy-Conserving Neural Network Closure Model for Long-Time Accurate and Stable \revtwo{2D} LES} % Vul in.
% \date{\today}
% \author{T. van Gastelen, W. Edeling, B. Sanderse\\
% Centrum Wiskunde \& Informatica (CWI), Amsterdam} % vul in
\author[1,eindhoven]{T. van Gastelen}
\author[1]{W. Edeling}
\author[1,eindhoven]{B. Sanderse}

%\address[1]{organization={Centrum Wiskunde \& Informatica},%Department and Organization
%            addressline={Science Park 123}, 
%            city={Amsterdam},
%            country={The Netherlands}}

\address[1]{Centrum Wiskunde \& Informatica, %Department and Organization
            Science Park 123, 
            Amsterdam,
            The Netherlands}
\address[Eindhoven]{Centre for Analysis, Scientific Computing and Applications, Eindhoven University of Technology,
    PO Box 513,Eindhoven,
    5600 MB,The Netherlands}

%% Abstract
\begin{abstract}
%% Text of abstract
Machine learning-based closure models for \gls{LES} have shown promise in capturing complex turbulence dynamics but often suffer from instabilities and physical inconsistencies. In this work, we develop a novel skew-symmetric neural architecture as closure model that enforces stability while preserving key physical conservation laws. Our approach leverages a discretization that ensures mass, momentum, and energy conservation, along with a face-averaging filter to maintain mass conservation in coarse-grained velocity fields. We compare our model against several conventional data-driven closures (including unconstrained convolutional neural networks),  and the physics-based Smagorinsky model. Performance is evaluated on decaying turbulence and Kolmogorov flow for multiple coarse-graining factors. In these test cases, we observe that unconstrained machine learning models suffer from numerical instabilities. In contrast, our skew-symmetric model remains stable across all tests, though at the cost of increased dissipation. Despite this trade-off, we demonstrate that our model still outperforms the Smagorinsky model in unseen scenarios. These findings highlight the potential of structure-preserving machine learning closures for reliable long-time \gls{LES}.

\end{abstract}

%%Graphical abstract
%\begin{graphicalabstract}
%\includegraphics{grabs}
%\end{graphicalabstract}

%%Research highlights
\begin{highlights}

\item Introduce energy-conserving neural network closure for turbulence.

\item Skew-symmetric term redistributes energy; negative definite term dissipates energy.

\item Outperforms standard machine learning models; delivers accurate long-time LES.

\item Neural network training procedure consistently yields accurate and stable LES.

\begin{comment}
\item We present an energy-conserving neural network architecture suitable for turbulence closure modeling. The main novelty is that this architecture is skew-symmetric by design. This means it is designed to disperse energy throughout the computational domain. The architecture is enhanced with a dissipative term, to allow for dissipation. The inclusion of the skew-symmetric term extends the closure model's span beyond an eddy-viscosity basis. As our closure model can only redistribute or dissipate energy it provides stability guarantees for the resulting \gls{LES}. 
\item  We compare our architecture against standard machine learning approaches which are plagued by stability issues. Our architecture does not suffer from these. We find that, even though our closure model is trained on reproducing solution trajectories for very short time intervals, it still produces accurate long-time statistics. This means the introduced architecture opens the door to accurate long-time machine learning-based \gls{LES}.
\item We also evaluate the consistency of the training procedure with respect to closure model performance. This is done by training multiple instances of each machine learned-based closure model. We find that performance is `hit-or-miss' for the standard machine learning approaches, meaning good offline performance does not equate to a stable/accurate \gls{LES}. Our architecture does not suffer from this, producing stable and accurate \glspl{LES} for all the trained instances.
\end{comment}

\end{highlights}

%% Keywords
\begin{keyword}
large eddy simulation \sep structure preservation \sep closure modeling \sep machine learning \sep Navier-Stokes equations
%% keywords here, in the form: keyword \sep keyword

%% PACS codes here, in the form: \PACS code \sep code

%% MSC codes here, in the form: \MSC code \sep code
%% or \MSC[2008] code \sep code (2000 is the default)

\end{keyword}

\end{frontmatter}

%% Add \usepackage{lineno} before \begin{document} and uncomment 
%% following line to enable line numbers
%% \linenumbers
\section{Introduction}\label{sec:introduction}

The incompressible Navier-Stokes equations are a set of \glspl{PDE} that describe conservation of mass and momentum of fluid flows. %, and energy conservation (in absence of viscosity). 
They are used to model a multitude of flow phenomena, such as for the design of aircraft and ships, weather modeling, and even the formation of galaxies \cite{sasaki2002navier,NS_galaxy}.
To simulate these phenomena, we solve the Navier-Stokes equations on a computational grid. This requires discretizing the differential operators present in the \gls{PDE}. This can be done with various techniques, such as finite difference, finite volume, and finite element methods
 \cite{benjamin_thesis,girault2012finite,FD_NS}.
In this work we employ a structure-preserving finite difference scheme, introduced in \cite{harlow1965numerical}. The advantage of this scheme is that it not only satisfies mass and momentum conservation, but also conserves the global kinetic energy (in absence of viscosity and boundary contributions). We refer to conservation of mass, momentum and kinetic energy collectively as the physical `structure’ of the system, and we refer to such conservative discretization schemes as `structure-preserving’ or `symmetry-preserving' \cite{coppola_global_2023,verstappen_2003}. Such schemes have the advantage of being unconditionally stable without relying on artificial diffusion, such as upwind schemes \cite{chu2024newlowdissipationcentralupwindschemes}. This makes them more suitable for long-time simulations, where correct physical energy behavior is crucial. However, problems arise when considering high Reynolds number flows \cite{sagaut2006large}.
For such flows we require very fine computational grids to resolve the smallest eddies in the system. This places a large (often insurmountable) burden on the computational resources.

A common approach to reduce computational requirements is \acrfull{LES}. In \gls{LES}, the system is coarse-grained by applying a filter to the velocity field, so that the dimension of the problem is effectively reduced. In this work, we take the `discretize first, filter next’ approach \cite{Syver,Melchers2022,VANGASTELEN2024113003}. This means that coarse-graining is done on the discrete level by applying a discrete filter to a fine-grid discretization. In particular, we use a face-averaging filter, which has the advantage that the filtered velocity field still satisfies mass conservation \cite{Agdestein_2025}. The latter is necessary to preserve the energy-conserving properties of the convection operator. This provides substantial stability benefits, see \cite{Agdestein_2025}.
From the coarse-graining procedure, a commutator error arises. In the `discretize first, filter next' framework, this commutator error also includes the discretization error. Modeling the commutator error is referred to as closure modeling, and the corresponding models are closure models. Closure modeling is a main subject in \gls{LES} research \cite{sagaut2006large}.

The most commonly used closure models are eddy-viscosity (functional) models \cite{shankar2024differentiableturbulenceclosurepartial,smagorinsky1963general}. Their primary job is to remove (dissipate) energy from the system to account for energy transfer from the resolved scales to the unresolved scales. These models approximate the subgrid stress solely from the resolved scales by assuming proportionality between the subgrid-scale stress tensor and the rate-of-strain tensor \cite{sagaut2006large,Pope_2000}. The classical Smagorinsky model is the best-known example, but it is strictly dissipative \cite{smagorinsky1963general,sagaut2006large,Pope_2000} and therefore cannot represent backscatter, i.e., energy transfer from unresolved to resolved scales. Backscatter, however, plays an important role in many flows, including geophysical turbulence relevant for weather and climate modeling \cite{Domaradzki_1987,Carati_1995,grooms_2015}.

\R{rev22}\revtwo{To address this limitation, structural models, such as scale-similarity closures, aim to reproduce the actual structure of the subgrid-scale stress tensor rather than parameterize its dissipation. While these models can, in principle, capture backscatter, they lack inherent dissipation. Consequently, they are often numerically unstable, requiring stabilization procedures such as explicit filtering or averaging. This need for both physical fidelity and numerical robustness motivated the development of mixed models, which combine a structural component with a dissipative functional component to achieve the accuracy of the former and the stability of the latter \cite{meneveau2000}. The dynamic Smagorinsky model \cite{germano_1992} represents a related effort to increase flexibility within functional closures by adapting the eddy-viscosity coefficient based on resolved-scale information. Although it can produce limited backscatter, it too can suffer from numerical instabilities \cite{lilly1992}.
Several physics-based strategies have since been proposed to improve backscatter representation while maintaining stability. These include enforcing a consistent subgrid-scale energy budget \cite{energy_budget_jansen}, introducing filter-based artificial dissipation \cite{edoh2017}, and imposing explicit dissipation constraints \cite{iyer2024,shi2008}. Nevertheless, state-of-the-art subgrid-scale closures used in global climate models still exclude backscatter \cite{Hewitt2020}, primarily due to concerns over robustness.}
\R{rev_3_M_3_a}\revthree{In response, the machine-learned closure models that we will propose aim, in a somewhat similar spirit as mixed models, to combine the best of functional and structural approaches: we keep the structure of the subgrid stress tensor, while at the same time ensuring that the models are energy dissipative (but not as much as existing functional models).}

Shifting our attention to such machine learning algorithms, neural networks have shown great potential in accurately modeling the closure term, while accounting for backscatter \cite{beck2,beck4,List,shankar2024differentiableturbulenceclosurepartial,Maulik_2018,Kochkov,park,NEURIPS2023_dabaded6,clipping_and_instability_and_dissipation_CNN_Guan_2022}. Offline testing often shows good agreement of the neural network outputs with the true closure term. However, when using the closure term in an actual simulation, problems arise and instabilities occur \cite{beck1, beck2, List, park,clipping_and_instability_and_dissipation_CNN_Guan_2022,Maulik_2018}. 
One approach to handle this issue is by clipping the neural network such that the output becomes strictly dissipative, for example by projecting onto an eddy-viscosity basis \cite{beck1, park,clipping_and_instability_and_dissipation_CNN_Guan_2022}. However, this 
results in closure models which are generally too dissipative \cite{clipping_and_instability_and_dissipation_CNN_Guan_2022}, which will be confirmed by our results. Another way to deal with instabilities is to add artificial noise to the training \cite{beck3}.
However, in \cite{beck3} it was shown that this only delays the instability and does not prevent it entirely.
Stochastic approaches have also been suggested, e.g. 
\cite{NEURIPS2023_dabaded6} combines idealized \gls{LES} with neural stochastic differential equations 
and show increased stability.

Stability issues are often combatted by introducing some form of \textit{a posteriori} learning  \cite{Frezat,List,MacArt_embedded_learning,Syver,Melchers2022,beck4,Bae2022,sanderse2024scientificmachinelearningclosure}. In this way, a closure model is trained based on reproducing the solution trajectory rather than reproducing the closure term itself. This aids in the stability of the \gls{LES}. In \cite{List}, it was shown that by increasing the number of solver steps one unrolls during training,  the stability and performance are improved. \cite{Melchers2022} shows that there is an optimum in the number of unrolled steps, related to the chaotic nature of the system. However, in \cite{clipping_and_instability_and_dissipation_CNN_Guan_2022} it is shown that limited training data still causes instabilities for neural network-based closure models. 

In \cite{eigenframe}, a strictly local small neural network approach is suggested for the representation of the subrid-scale stress tensor. Here, a small set of carefully chosen Galilean invariant and non-dimensionalized inputs are used such that the resulting closure model is symmetric, Galilean invariant, rotationally and reflectionally invariant, and unit invariant. The authors show that training on a single snapshot is enough to obtain a closure model which generalizes well to their considered test cases, even without clipping. However, in \cite{shankar2024differentiableturbulenceclosurepartial} it is shown that \glspl{CNN}, combined with Fourier neural operators, with non-invariant inputs, is capable of outperforming such approaches. 
For an overview on machine learning-based closure modeling see \cite{sanderse2024scientificmachinelearningclosure}.

\R{rev_2_M_5}\revtwo{However, while the aforementioned references observed improved stability, none of the discussed approaches \textit{guarantees} stability, without applying some form of backscatter clipping, or other ad-hoc measures, such as e.g.\ trial-and-error by changing input features \cite{park} or data augmentation \cite{beck3}. This makes long-time \gls{LES} with a data-driven closure model unreliable. Given that we are especially interested in the statistics of turbulent flows, e.g.\ the average energy spectrum obtained over long time horizons, it is of crucial importance that the combination of discretization and closure model yields stable simulations. Hence, the main aim and novelty of this article is to derive a machine-learned LES closure model, where stability is guaranteed by design, independent of the network weights or amount of training data. To achieve this}, we propose to build upon our earlier work presented in \cite{VANGASTELEN2024113003}. In this work, the closure model was represented by an energy-conserving skew-symmetric term and a dissipative symmetric negative-definite term. This was accomplished by using a set of compressed subgrid-scale variables that allow the transfer of energy from unresolved scales to resolved scales. \revtwo {In this work, we extend this approach from one to two spatial dimensions. While it is certainly true that turbulence is fundamentally a 3D phenomenon, we reiterate that the main aim of our article is focused on the stability issue, which has largely been studied in the context of 2D turbulence, see \cite{shankar2024differentiableturbulenceclosurepartial, beck2, List, Maulik_2018, Kochkov, clipping_and_instability_and_dissipation_CNN_Guan_2022, Frezat} among others.} 

As a first step, we exclude the compressed subgrid variables, as their precise meaning and definition in multiple dimensions is still an open question. Instead, we show that the skew-symmetric framework without subgrid variables already offers significant stability advantages over existing approaches. 
%However, it is not quite clear how to extend this to multiple spatial dimensions, beyond the 1D applications that were considered. It is therefore that we introduce this framework without these subgrid-scale variables. 
Although this means we do not explicitly account for backscatter, the skew-symmetric term is still capable of redistributing energy throughout the domain. This extends the predictive capability of the closure model beyond an eddy-viscosity basis. The work presented in \cite{Hernandez_2022} discusses similar ideas regarding structure-preserving neural networks, although outside the realm of \gls{LES}.

The outline of this paper is as follows: In section \ref{sec:preliminaries}, we start with introducing the incompressible Navier-Stokes equations, the physical structure present in the system, and the structure-preserving discretization. In section \ref{sec:prelimenaries_2}, we discuss coarse-graining of the discrete system of equations by applying a discrete spatial filter, we derive the exact closure term, and discuss closure modeling approaches. In section \ref{sec:methodology}, we introduce the machine learning-based closure models: using a \gls{CNN} to model the closure term, using a \gls{CNN} to model the subgrid-scale stress tensor, and finally our skew-symmetric neural network architecture. An extended motivation for this architecture can be found in \ref{app:motivation}. Next, we discuss the test cases and the results in section \ref{sec:results}, regarding closure model performance and stability. We conclude our work in section \ref{sec:conclusion}.

\section{Preliminaries}\label{sec:preliminaries}

\subsection{Navier-Stokes equations}

In convervative form, the incompressible Navier-Stokes equations read as follows:
\begin{subequations}\label{eq:NS}
\begin{equation}
    \frac{\partial \mau }{\partial t } + \nabla \cdot (\mau \mau ^T) = -\nabla p + \nu \nabla^2 \mau + \mathbf{f},
\end{equation}
\begin{equation}
    \nabla \cdot \mau   = 0.
\end{equation}
\end{subequations}
This \gls{PDE} describes the evolution of a fluid velocity field $\mau(\mathbf{x},t) \in \mathbb{R}^d$ in $d$-dimensional space $\mathbf{x} \in \mathbb{R}^{d}$ and time $t$. Here we restrict ourselves to 2D such that $\mathbf{u}(\mathbf{x},t) = \begin{bmatrix}
    u(\mathbf{x},t) & v(\mathbf{x},t)
\end{bmatrix}^T$. The different forces acting on the velocity field are due to convection, the gradient of the pressure $p(\mathbf{x},t)$, and friction (for a nonzero viscosity $\nu \geq 0$), appearing from left to right in the equation. In addition, there is  $\mathbf{f}(\mathbf{x},t)\in \mathbb{R}^2$ which represents body-forces, such as gravity. \R{rev_3_m_2}\revthree{In this text we refrain from discussing boundary conditions, as we consider a periodic spatial domain $\Omega$.}

\subsection{Physical structure}

This set of equations represent a set of fundamental physical laws, namely: conservation of mass, momentum $\mathbf{P} = \int_\Omega \mathbf{u}\text{d}\Omega$, and (kinetic) energy $E := \frac{1}{2}\int_\Omega \mathbf{u} \cdot \mathbf{u} \text{d}\Omega$ (for $\nu = 0$). These are collectively referred to as the system's physical structure.
In equation form, these laws read
\begin{align}
    \nabla \cdot \mathbf{u} &= 0, \\
    \frac{\text{d}\mathbf{P}}{\text{d}t} =& \int_\Omega \frac{\partial\mau}{\partial t}\text{d}\Omega = \int_\Omega \mathbf{f}(x,t)\text{d}\Omega, \\
    \frac{\text{d}E}{\text{d}t} =&\int_\Omega \mau \cdot \frac{\partial\mau}{\partial t}\text{d}\Omega= -\int_\Omega \nu  ||\nabla \mau||_2^2  \text{d}\Omega + \int_\Omega \mau \cdot \mathbf{f}(\mathbf{x},t)  \text{d}\Omega,
\end{align}
Derivations of these conservation laws are presented in \ref{sec:structure_appendix}. From these laws, we can see that momentum and energy (for zero dissipation) are conserved in the absence of forcing.
   
\subsection{Discretization}\label{sec:sd-NS}

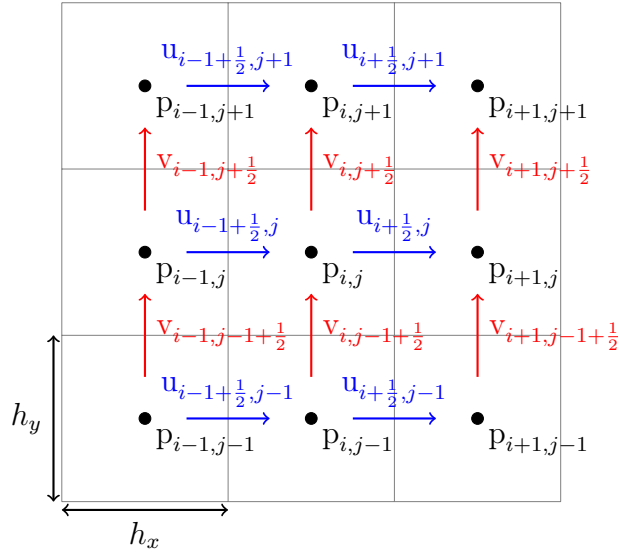
\begin{figure}
    \centering
\begin{tikzpicture}[scale=1.1]

    % Grid
    \draw[step=2cm, gray, very thin] (0,0) grid (6,6);

    % Pressure Points (P) with Indices
    \foreach \x/\i in {1/i-1,3/i,5/i+1}
        \foreach \y/\j in {1/j-1,3/j,5/j+1}
            \filldraw[black] (\x,\y) circle (2pt) node[below right] {$\text{p}_{\i,\j}$};

    % U-Velocity (u) at vertical faces with indices
    \foreach \x/\i in {2/i-1,4/i}
        \foreach \y/\j in {1/j-1,3/j,5/j+1}
            \draw[->,blue,thick] (\x-0.5,\y) -- (\x+0.5,\y) 
            node[midway, above] {$\text{u}_{\i+\frac{1}{2},\j}$};

    % V-Velocity (v) at horizontal faces with indices
    \foreach \x/\i in {1/i-1,3/i,5/i+1}
        \foreach \y/\j in {2/j-1,4/j}
            \draw[->,red,thick] (\x,\y-0.5) -- (\x,\y+0.5) 
            node[midway, right] {$\text{v}_{\i,\j+\frac{1}{2}}$};
    
    % Grid spacing arrows in bottom-left corner
    \draw[<->,thick] (0,-0.1) -- (2,-0.1) node[midway, below] {$h_x$};
    \draw[<->,thick] (-0.1,0) -- (-0.1,2) node[midway, left] {$h_y$};

\end{tikzpicture}
\caption{Staggered grid discretization of Navier-Stokes equations. The pressure points are indexed with whole numbers, while for the velocity components we have an offset of $\frac{1}{2}$ in the appropriate direction.}\label{fig:staggered_grid}
\end{figure}

To simulate practical use cases, we require discretizing the set of equations \eqref{eq:NS} on a computational grid. Here we employ a structure-preserving second-order accurate finite \R{rev_3_m_1}\revthree{difference} discretization on a staggered grid \cite{harlow1965numerical,benjamin_thesis}. This discretization is chosen as it satisfies the energy-conserving properties of the convective term. The employed discretization consists of in total $N_x \times N_y = N$ cells $\Omega_{ij}$ with $i=1,\ldots N_x$ and $i=1,\ldots N_y$ for a 2D flow case. The semi-discrete set of equations are written as
\begin{subequations}\label{eq:sd-NS}
\begin{equation}\label{eq:sd-NS_1}
\boldsymbol{\Omega}_h\frac{\text{d}\mauh}{\text{d}t} + \mathbf{C}_h(\mauh)\mauh = - \mathbf{G}_h \mathbf{p}_h + \nu \mathbf D \mauh + \boldsymbol{\Omega}_h\mathbf{f}_h,
\end{equation}
\begin{equation}\label{eq:div_free}
    \mathbf{M}_h\mauh = \mathbf{0},
\end{equation}
\end{subequations}
where $\mauh \in \mathbb{R}^{2N}$ contains the approximations of $u$ and $v$ on the cell faces and $\mathbf{p}_h \in \mathbb{R}^{N}$ the approximation of $p$ in the cell centers. A schematic representation of the employed uniform staggered grid is displayed in Figure \ref{fig:staggered_grid}. The grid-spacing in each direction is indicated by $h_x$ and $h_y$. 
Furthermore, $\boldsymbol{\Omega}_h$ contains the cell volumes on the diagonal. The operators in \eqref{eq:NS} are now represented by matrices. $\mathbf{C}_h(\mauh) \in \mathbb{R}^{2N \times 2N}$ represents the convection operator, $\mathbf{G}_h \in \mathbb{R}^{2N \times N}$ the gradient operator, $\mathbf{D}_h \in \mathbb{R}^{2N \times 2N}$ the diffusion operator,  $\mathbf{M}_h \in \mathbb{R}^{N \times 2N}$ the divergence operator, and $\mathbf{f}_h  \in \mathbb{R}^{2N}$ the forcing at the cell faces. 

\subsection{Structure of the discretization}

This discretization preserves the physical structure in a discrete sense. Discretely, the total momentum and energy are approximated as
\begin{align}
    \mathbf{P}_h &=  \underbrace{\begin{bmatrix} \mathbf{1}_h & \mathbf{0}_h \\ \mathbf{0}_h & \mathbf{1}_h 
    \end{bmatrix}^T}_{=: \boldsymbol{\mathbbm{1}}_h} \boldsymbol{\Omega}_h\mauh, \\
    E_h &= \frac{1}{2} \mauh^T\boldsymbol{\Omega}_h\mauh,
\end{align}
where $\mathbf{0}_h,\mathbf{1}_h \in \mathbb{R}^N$ are column vectors of zeros and ones, respectively.
The change of these quantities for this discretization are given by
\begin{align}
    \frac{\text{d}\mathbf{P}_h}{\text{d}t} &= \boldsymbol{\mathbbm{1}}_h  \frac{\text{d}\mauh}{\text{d}t} =\boldsymbol{\mathbbm{1}}_h\boldsymbol{\Omega}_h\mathbf{f}_h, \\
   \frac{\text{d} E_h}{\text{d}t} &= \mauh^T\frac{\text{d}\mauh}{\text{d}t}  =- \nu||\mathbf{Q}_h \mauh||_2^2 + \mauh^T\boldsymbol{\Omega}_h\mathbf{f}_h,
\end{align}
where we used the fact that the diffusion operator $\mathbf{D}_h$ can be Cholesky decomposed as $-\mathbf{Q}_h^T\mathbf{Q}_h$ \cite{podbenjamin}. As the discrete energy is solely decreasing, in the absence of forcing, this discretization provides stability.
The convective contribution disappears due to the skew-symmetry of the discrete operator, however this requires the discrete solution $\mathbf{u}_h$ to be divergence free, i.e., $\mathbf{M}_h\mauh = \mathbf{0}_h$. A more elaborate discussion is presented in \ref{sec:discretization_appendix}.

\subsection{Pressure projection}

The divergence freeness can also be written as a projection of the \gls{RHS} of the \gls{PDE} discretization \eqref{eq:sd-NS} on a divergence-free basis. We introduce this formulation of the discrete system because it is more convenient for closure modeling \cite{Agdestein_2025}, as it combines \eqref{eq:sd-NS_1} and \eqref{eq:div_free} into a single equation. The projection looks as follows:
\begin{equation}
    \boldsymbol{\Omega}_h\frac{\text{d}\mauh}{\text{d}t} =\mathcal{P}_h\mathbf{m}_h(\mauh),
\end{equation}
where $\mathbf{m}_h(\mauh) \in \mathbb{R}^{2N}$ contains the operators on the \gls{RHS} of \eqref{eq:sd-NS_1}, i.e.,
\begin{equation}
    \mathbf{m}_h(\mauh) = - \mathbf{C}_h(\mauh)\mauh + \nu \mathbf{D}_h \mauh + \boldsymbol{\Omega}_h\mathbf{f}_h,
\end{equation}
and $\mathcal{P}_h \in \mathbb{R}^{2N \times 2N}$ projects the \gls{RHS} on a divergence free basis. $\mathcal{P}_h$ is defined as
\begin{equation}
    \mathcal{P}_h := (\mathbf{I} - \mathbf{G}_h (\mathbf{M}_h\boldsymbol{\Omega}_h^{-1}\mathbf{G}_h)^{-1}\mathbf{M}_h \boldsymbol{\Omega}_h^{-1}).
\end{equation}
A derivation of this operator is presented in \ref{sec:pressure_projection_appendix}.

\section{Closure modeling}\label{sec:prelimenaries_2}

\subsection{Filtering}

\begin{figure}
    \centering

\begin{tikzpicture}[scale = 1.5]

    % Thick border
    \draw[very thick] (0,0) -- (3,0);
    \draw[very thick] (0,3) -- (3,3);
    \draw[very thick] (0,0) -- (0,3);
    \draw[very thick] (3,0) -- (3,3);

    % Grid lines (thin)
    \foreach \x in {1,2} {
        \draw[gray] (\x,0) -- (\x,3);
    }
    \foreach \y in {1,2} {
        \draw[gray] (0,\y) -- (3,\y);
    }

    % Small velocity arrows (staggered locations)
    \foreach \x in {0.5,2.5} {
        \foreach \y in {0,1,2,3} {
            \draw[->, red] (\x,\y-0.2) -- (\x,\y+0.2);
        }
    }

    \foreach \x in {1.5} {
        \foreach \y in {1,2} {
            \draw[->, red] (\x,\y-0.2) -- (\x,\y+0.2);
        }
    }

    \foreach \x in {0,1,2,3} {
    \foreach \y in {0.5,2.5} {
        {\draw[->, blue] (\x-0.2,\y) -- (\x+0.2,\y);}
    }
}

\foreach \x in {1,2} {
    \foreach \y in {1.5} {
        {\draw[->, blue] (\x-0.2,\y) -- (\x+0.2,\y);}
    }
}

    % Big arrows on the grid boundary
    \draw[->, very thick, black] (1.5,-0.5) -- (1.5,0.5);
    \draw[->, very thick, black] (1.5,2.5) -- (1.5,3.5);
    \draw[->, very thick, black] (-0.5,1.5) -- (0.5,1.5);
    \draw[->, very thick, black] (2.5,1.5) -- (3.5,1.5);

    % Labels
    %\node[below] at (1.5,-0.5) {\large $\bar{\text{v}}_H$};
    %\node[above] at (1.5,3.5) {\large $\bar{\text{v}}_H$};
    \node[left] at (-0.5,1.5) {\large $\bar{\text{u}}_H$};
    %\node[right] at (3.5,1.5) {\large $\bar{\text{u}}_H$};
    %\node[right] at (0.2,-0.5) {\large $\text{v}_h$};
    \node[right] at (-0.7,0.5) {\large $\text{u}_h$};

\end{tikzpicture}
    \caption{Schematic representation of the face-averaging filter. In this example, a single coarse-grained cell contains nine fine-grid cells. Three fine-grid velocity components in $\mauh$ on each of the coarse cell faces are averaged to obtain the filtered velocity field $\bmauh$. }
    \label{fig:FA_filter}
\end{figure}
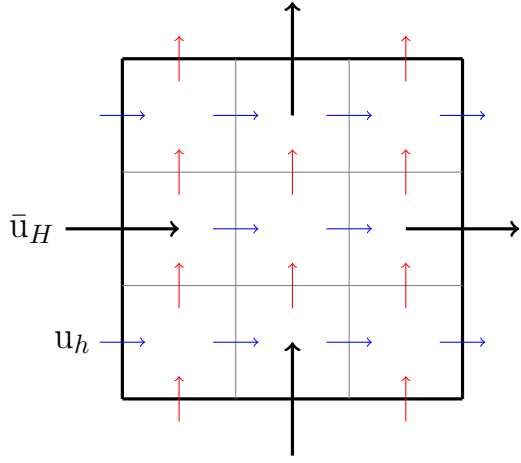

As stated earlier, carrying out a \gls{DNS} is often infeasible for practical use cases. This is why we aim to solve a coarse-grained set of equations. Coarse-graining is done by applying a filter to the velocity field. In our case, we take the `discretize first, filter next' approach \cite{Syver,Melchers2022}. This means we start by discretizing our velocity field on an adequately fine grid, such that we resolve the relevant scales of the flow. Next, we apply a linear filter $\mathbf{W}^{2\bar{N}\times 2N}$ to obtain the filtered velocity field:
\begin{equation}
    \bmauh = \mathbf{W}\mauh,
\end{equation}
where the filtered velocity $\bmauh \in \mathbb{R}^{2\bar{N}}$ lives on a coarse grid consisting of $\bar{N} = \bar{N}_x \times \bar{N}_y$ grid cells. In our case we employ a face-averaging filter, as suggested in \cite{Agdestein_2025}. A schematic representation of the filter is shown in \ref{fig:FA_filter}. 
The advantage of this filter, as opposed to, for example, a volume-averaging filter, is that the filtered velocity satisfies divergence freeness on the coarse grid. This ensures skew-symmetry of the coarse-grid convection operator, which aids in stability of the coarse-grained system of equations \cite{Agdestein_2025}. 

\subsection{System of equations}

To model the behavior of the filtered velocity field, we take the following ansatz:
\begin{equation}\label{eq:ansatz_1}
    \boldsymbol{\Omega}_H\frac{\text{d}\bmauh}{\text{d}t} \approx \mathcal{P}_H\mathbf{m}_H (\bmauh) + \tilde{\mathbf{c}}(\bmauh,\theta),
\end{equation}
where the subscript $H$ indicates a coarse-grid equivalent to the operators discussed in section \ref{sec:sd-NS} and $\tilde{\mathbf{c}}(\bmauh,\theta)$ is a (neural network-based) closure model with parameters $\theta$. The projection on a divergence-free basis is achieved through $\mathcal{P}_H$, which is justified due to the face-averaging filter. The closure model is required as the coarse discretization does not resolve all the relevant scales of the flow. In addition, the coarse grid results in a discretization error. Both of these are captured in the commutator error with respect to the fine-grid discretization. The true evolution of $\bmauh$ is given by
\begin{equation}
    \boldsymbol{\Omega}_H\frac{\text{d}\bmauh}{\text{d}t} = \mathcal{P}_H\mathbf{m}_H (\bmauh) + \underbrace{(\mathbf{W} \mathcal{P}_h\mathbf{m}_h(\mauh)-\mathcal{P}_H\mathbf{m}_H(\bmauh) )}_{=: \mathbf{c}(\mauh)},
\end{equation}
where the commutator error $\mathbf{c}_h(\mauh)$ includes both sources of error. As the true filtered velocity field is divergence-free on the coarse grid, we include the closure model in the projection to preserve divergence freeness, as suggested by \cite{Agdestein_2025}. This changes ansatz \eqref{eq:ansatz_1} to 
\begin{equation}\label{eq:ansatz_2}
    \boldsymbol{\Omega}_H\frac{\text{d}\bmauh}{\text{d}t} \approx \mathcal{P}_H(\mathbf{m}_H (\bmauh) + \tilde{\mathbf{c}}(\bmauh,\theta)).
\end{equation}
The energy contribution of the closure model is computed as
\begin{equation}\label{eq:energy_contribution}
\text{energy contribution} = \bmauh^T\tilde{\mathbf{c}}(\bmauh,\theta),
\end{equation}
where the resolved energy is given by
\begin{equation}
    \bar{E}_H = \frac{1}{2}\bmauh^T \boldsymbol{\Omega}_H \bmauh.
\end{equation}
In addition, the total momentum $\bar{\mathbf{P}}_H = \boldsymbol{\mathbbm{1}}_H \boldsymbol{\Omega}_H  \mathbf{u}_H$ is conserved if 
\begin{equation}\label{eq:mom_cons}
    \boldsymbol{\mathbbm{1}}_H\tilde{\mathbf{c}}(\mathbf{u}_H,\theta) = \begin{bmatrix}
        0 \\ 0 
    \end{bmatrix}
\end{equation} 
holds for the closure model.

\subsection{Energy analysis of the closure term}

\begin{figure}
    \centering
\includegraphics[width = 0.48\textwidth]{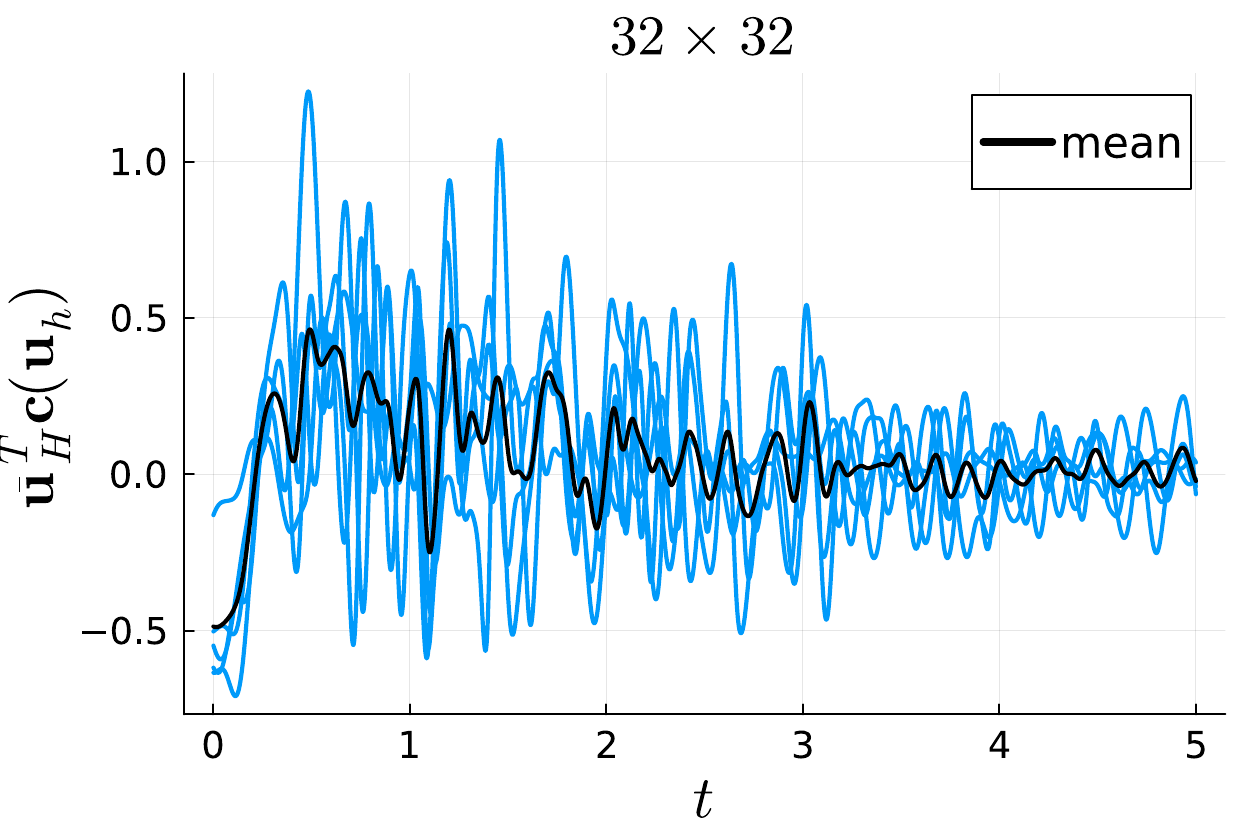}
\includegraphics[width = 0.48\textwidth]{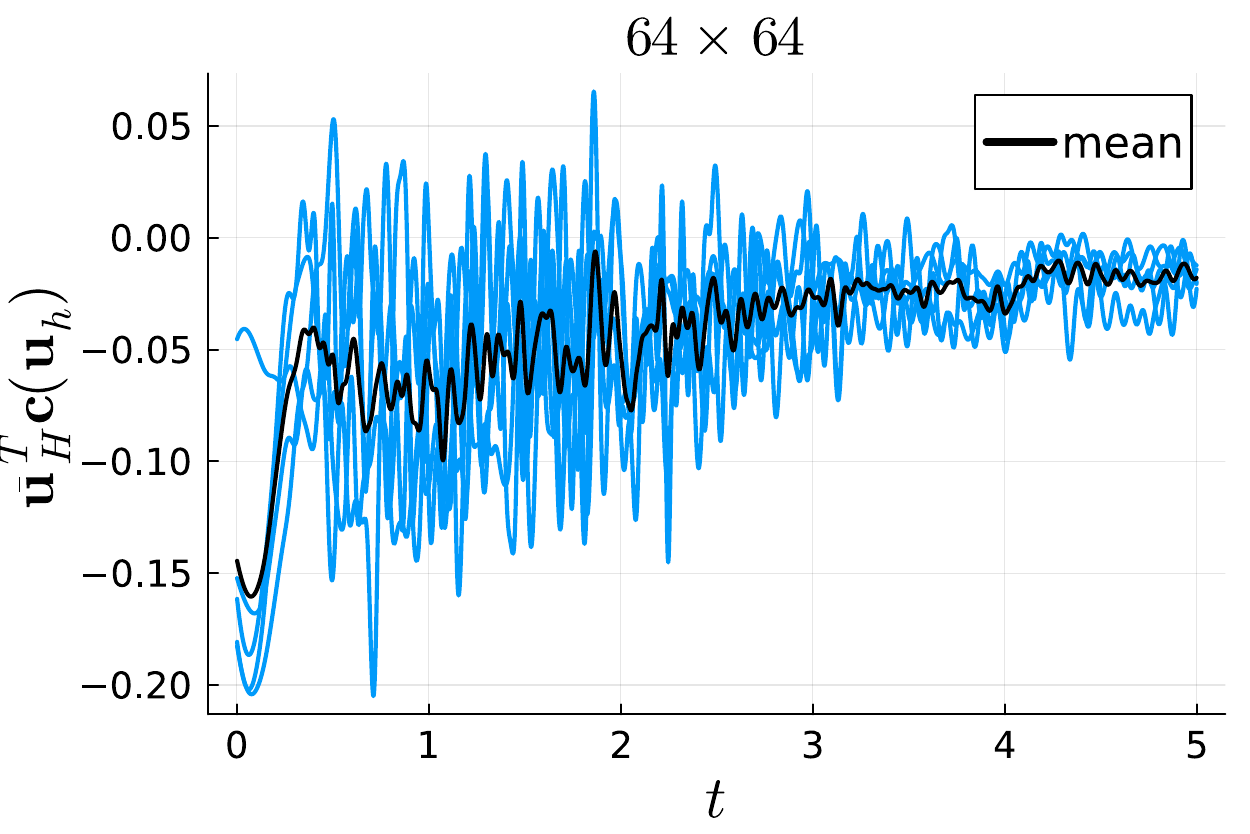}
\includegraphics[width = 0.48\textwidth]{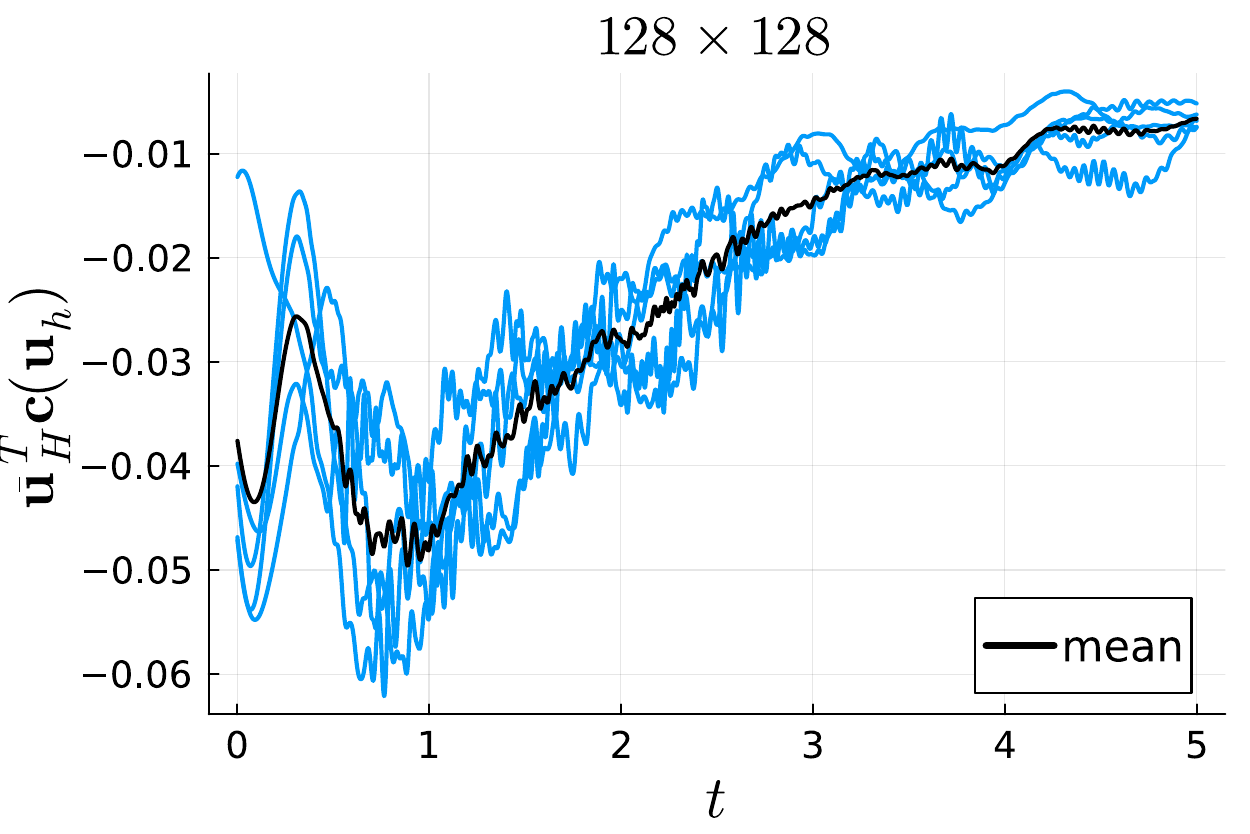}
\includegraphics[width = 0.48\textwidth]{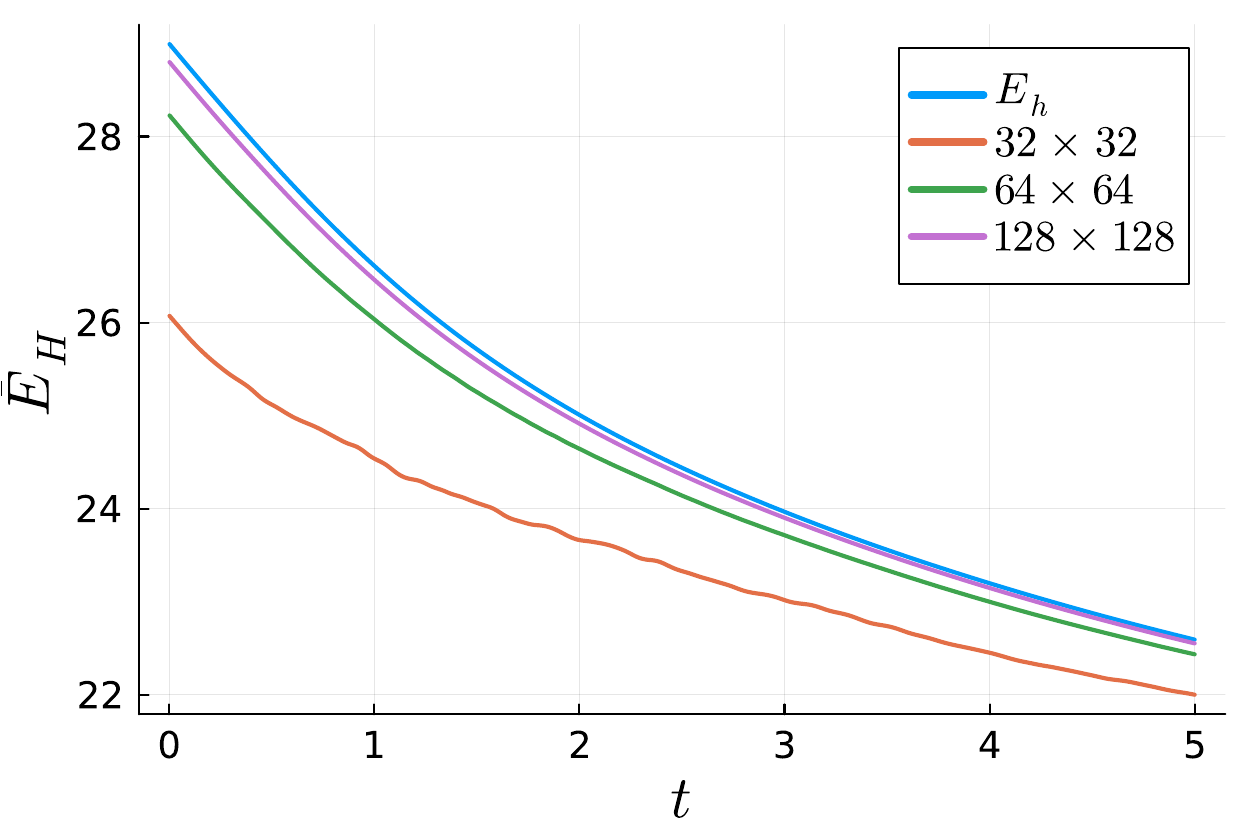}
\caption{Resolved energy contributions for the true closure term, computed for five decaying turbulence simulations which constitute the training data for the machine learning closure models. The trajectories are presented for different levels of coarse-graining. The reference grid has a resolution of $2048 \times 2048$. See section \ref{sec:experimental_setup} for the exact simulation conditions. (Bottom-right) Resolved energy trajectories for one of the simulations. }\label{fig:offline}
\end{figure}

Based on training data, we analyze the energy contribution of the true closure term, see \eqref{eq:energy_contribution}, for different levels of coarse-graining. In this case, the reference simulation was carried out on a $2048 \times 2048$ grid. Exact simulation conditions are discussed in section \ref{sec:experimental_setup}. The resulting energy contributions, in addition to the resolved energy trajectories, are presented in Figure \ref{fig:offline}. 
We observe that for a resolution of $32 \times 32$ the closure term produces a significant amount of backscatter, as the energy contribution is mostly positive. Also, in the resolved energy trajectory we observe that the decay is less smooth, as opposed to larger resolutions. This means dissipative models will likely perform poorly for this resolution. For a resolution of $64 \times 64$, we find the energy contribution to be mostly dissipative, whereas for $128 \times 128$ it is strictly dissipative for this test case. This motivates the use of dissipative closure models, such as the skew-symmetric neural network architecture we will introduce in Section \ref{sec:architecture}

\subsection{Fitting of the model parameters}

To find the optimal set of parameters $\theta$, one can take different approaches. The most straightforward approach would be to match the true commutator error as best as possible. However, this often results in inaccurate simulations or even instabilities \cite{Syver,Melchers2022,List,beck1,beck2,beck3,beck4}. This is why we resort to optimizing the parameters in order to accurately reproduce the \gls{FDNS} solution,  also known as  `trajectory fitting' or `solver in the loop'.
%, enables optimization by computing the gradient of the loss function with respect to \( \theta \). 
The corresponding loss function is:
\begin{equation}\label{eq:traj_loss}
    \mathcal{L}_n(\mathbf{X};\theta) = \sum_{\mauh \in \mathbf{X}} \sum^n_{i=1}|| \bar{\mathcal{S}}^i_\theta (\mathbf{W}\mauh)- \mathbf{W}\mathcal{S}^{i(\overline{\Delta t}/\Delta t)} (\mauh) ||_2^2,
\end{equation}
where $\mathbf{X}$ is a snapshot matrix consisting of samples of $\mauh$ as columns, representing the training data set.
The notation $\bar{\mathcal{S}}_{\theta}^i(\mathbf{W}\mauh)$ represents the predicted coarsened velocity field after applying an explicit time integration scheme for \( i \) steps, each with a step size of \( \overline{\Delta t} \), starting from the initial condition \( \mathbf{W}\mauh \), incorporating the closure model. The corresponding \gls{DNS} solution is denoted by \( \mathcal{S}^{i(\overline{\Delta t}/\Delta t)}(\mauh) \), using a smaller step size \( \Delta t \) and initialized with \( \mauh \). The ratio \( \overline{\Delta t}/\Delta t \) arises because the coarse grid permits larger time steps \cite{de2013courant_CFL}. Note that $\bar{\mathcal{S}}_{\theta}$ is the solver that works on the coarse grid and incorporates the closure model, while $\mathcal{S}$ is the \gls{DNS} solver that works on the fine grid. The Adam optimization algorithm \cite{kingma_adam} will be used to minimize the loss. Further details on the training procedure are discussed in section \ref{sec:experimental_setup}.

\section{Methodology}\label{sec:methodology}
As explained in section \ref{sec:prelimenaries_2}, a main challenge in closure modeling is deriving an expression for $\tilde{\mathbf{c}}(\bmauh,\theta)$. In this section, we propose a skew-symmetric framework that results in a new expression for $\tilde{\mathbf{c}}(\bmauh,\theta)$, providing stability. In sections \ref{sec:smag} and \ref{sec:CNN} we first introduce the Smagorinsky model and \glspl{CNN}, respectively. These not only serve as something to compare our framework to, but also as necessary preliminaries. In section \ref{sec:architecture} the new framework is derived.

\subsection{Smagorinsky model}\label{sec:smag}

We start off with the Smagorinsky model, which is an eddy-viscosity model. As stated earlier, the main assumption in eddy-viscosity models is that this subgrid-scale stress tensor is proportional to the rate-of-strain tensor $\bar{\text{S}}_{ij} = \frac{1}{2} (\frac{ \partial \bar{\text{u}}_i}{\partial \text{x}_j}  + \frac{\partial \bar{\text{u}}_j}{\partial \text{x}_i} )$, where $\bar{\mathbf{u}} \in \mathbb{R}^2$ is a continuous representation of the resolved velocity field. Eddy-viscosity type closure models have the following form:
\begin{equation}
    \tilde{c}(\bar{\mathbf{u}}) = \nabla \cdot (\nu_t\bar{\mathbf{S}}),
\end{equation}
where $\nu_t \geq 0$ is the eddy-viscosity. The Smagorinsky model assumes the following form for $\nu_t$: 
\begin{equation}\label{eq:vt}
    \nu_t = (C_s\Delta)^2 \sqrt{2 \text{tr}(\bar{\mathbf{S}}^2)},
\end{equation}
where $\Delta$ is often chosen as the grid-spacing, i.e.\ $\Delta = \sqrt{h_x h_y}$. The model contains only a single parameter $C_s$, which can be tuned. In practice, we use a discretization of the Smagorinsky model. This yields a discrete closure model:
\begin{equation}
    \tilde{\mathbf{c}}^{\text{SMAG}}(\bmauh,C_s) =   (C_s\Delta)^2 \sqrt{2\text{tr}(\bar{\mathbf{S}}_H^2)}\nabla_H \cdot \bar{\mathbf{S}}_H,
\end{equation}
where the subscript $H$ indicates a discretization of the derivative operators. We employ a central difference scheme for these derivatives. The global energy contribution, see \eqref{eq:energy_contribution}, of the Smagorinsky model is always negative, i.e., it is strictly dissipative. In addition, the momentum conservation constraint \eqref{eq:mom_cons} is satisfied for the Smagorinsky model.

\subsection{Neural network closure}\label{sec:CNN}

A straightforward machine learning approach is to use a \gls{CNN} as a closure model, i.e.
\begin{equation}
    \tilde{\mathbf{c}}^\text{CNN}(\bmauh,\theta) = \text{CNN}(\bmauh,\theta).
\end{equation}
These models are well-suited to Cartesian grids \cite{Agdestein_2025,List,Kochkov}, such as the one we employ here. In addition, they are translation equivariant.
Here a \gls{CNN} is used to map the filtered solution $\bmauh \in \mathbb{R}^{2\bar{N}}$ to $\tilde{\mathbf{c}}\in \mathbb{R}^{2\bar{N}}$ by chaining a series of convolutions with non-linear activation functions $\sigma^n$, where $n$ indicates which layer of the \gls{CNN} is considered. A single layer of such a network is represented as
\begin{equation}
 \mathbf{z}^{n+1} = \sigma^n(\mathcal{A}^{n}\mathbf{z}^n + \mathbf{b}^n),
\end{equation}
where each vector $\mathbf{z}^n$ contains a set of fields, typically referred to as `channels' in \gls{CNN} literature. The matrix $\mathcal{A}$ contains $s_{n+1} \times s_n$ submatrices encoding convolutional stencils, where $s_n$ is the number of channels represented in $\mathbf{z}_n$. By choosing $s_{n+1}>s_{n}$, the data is effectively lifted to a higher-dimensional space. The vector $\mathbf{b}_n$ contains $s_{n+1}$ bias channels, which are constant fields each determined by a single parameter. For example, a single convolution $\mathbf{A}$ (parameterized by weights $\alpha_{jk}$) of a single channel $\mathbf{z}$, with bias vector $\mathbf{b}$ (parameterized by bias $\beta$), is represented as
\begin{equation}
    (\mathbf{A}\mathbf{z} + \mathbf{b})_{ij} = \sum_{k,l=-r}^{r}  (\alpha_{kl} \text{z}_{i+k,j+l}) + \beta
\end{equation}
The double index notation is used to indicate the location on the 2D grid (one index for each spatial dimension), see Figure \ref{fig:staggered_grid}. Moreover, $r$ represents the radius of the convolution in both spatial directions. On the edge of the domain, one uses padding of $\mathbf{z}$ to keep the size of the channels constant. In our case, we use circular padding to represent periodic \glspl{BC}. For intermediate layers, we use a ReLU activation function $\sigma^n$, and for the final layer, we take $\sigma^n$ to be the identity \cite{RELU}. Using a \gls{CNN} as a closure model does allow for modeling backscatter as its energy contribution, see \eqref{eq:energy_contribution}, can be both negative and positive. However, it violates momentum conservation, see \eqref{eq:mom_cons}. 

A straightforward way to resolve the latter is to use a \gls{CNN} to predict a stress tensor $\boldsymbol{\tau}^\text{CNN}$. The closure model is then obtained by taking the divergence of this tensor \cite{park,Melchers2022}:
\begin{equation}
    \tilde{\mathbf{c}}^\text{DIV}(\bmauh,\theta) = \nabla_H \cdot \boldsymbol{\tau}^\text{CNN}(\bmauh,\theta).
\end{equation}
This ensures that momentum conservation, see \eqref{eq:mom_cons}, is satisfied. This formulation will be referred to as \acrshort{DIV}.
However, neither of these formulations is guaranteed to be stable and can therefore result in poor simulation results.

\subsection{Skew-symmetric framework}\label{sec:architecture}

For our novel closure modeling approach we build on earlier work which we presented in \cite{VANGASTELEN2024113003}. In this work a skew-symmetric neural network architecture was introduced and applied to 1D equations. Moreover, a coarse-grid representation of the subgrid-scale energy was included in the coarse-grained system. However, as stated earlier in section \ref{sec:introduction}, it is unclear how to extend this representation to multiple dimensions. On the other hand, applying the proposed neural network architecture, without this subgrid-scale representation, to 2D problem does not require any modifications to the architecture.

In this skew-symmetric framework, the closure model is represented as the sum of a skew-symmetric and negative-definite term:
\begin{equation}
    \tilde{\mathbf{c}}^\text{SKEW}(\bmauh,\theta) = (\mathcal{K}- \mathcal{K}^T)\bmauh  - \mathcal{Q}^T\mathcal{Q}\bmauh, 
\end{equation}
where $\mathcal{K}(\bmauh,\theta),\mathcal{Q}(\bmauh,\theta)\in \mathbb{R}^{2\bar{N} \times 2\bar{N}}$ are build from \gls{CNN} outputs. The different terms in this closure model are represented by matrix multiplications of the velocity vector. Model flexibility stems from the fact that these matrices depend nonlinearly on the solution vector via neural network outputs. The closure model is always dissipative:
\begin{equation}
   \bmauh^T \tilde{\mathbf{c}}^\text{SKEW}(\bmauh,\theta) = \bmauh^T (\mathcal{K}- \mathcal{K}^T)\bmauh  - \bmauh^T\mathcal{Q}^T\mathcal{Q}\bmauh = - ||\mathcal{Q}\bmauh||_2^2 \leq 0,
\end{equation}
as the skew-symmetric contribution cancels.
This means the closure model is guaranteed to be stable, unlike the previously presented machine learning approaches.
\R{rev_3_M_1_a}\revthree{Note that stability is only guaranteed up to a time discretization error. Analyzing stability while accounting for the time discretization error would require knowledge of the Jacobian of the system. This makes formal analysis rather cumbersome, due to non-linearity of the neural network \cite{butcher2016num_integration}. Therefore, we consider this outside the scope of this research.}

\R{rev_3_M_2_a}\revthree{To ensure stability, ignoring the time discretization, the closure model does not allow for backscatter, but does increase modeling freedom beyond an eddy-viscosity basis through the skew-symmetric term.
To see this, we note any real square matrix $\mathbf{A}$ can be decomposed uniquely into the sum of a symmetric and a skew-symmetric matrix \cite{horn2013matrix, bauschke2009borwein}:
\begin{equation}
\mathbf{A} = \mathbf{S} + \mathbf{R}, \qquad
\mathbf{S} = \tfrac{1}{2}(\mathbf{A} + \mathbf{A}^T), \quad
\mathbf{R} = \tfrac{1}{2}(\mathbf{A} - \mathbf{A}^T).
\end{equation}
By constraining the symmetric part
$\mathbf{S}$ to be negative definite, we restrict ourselves to the space of stable linear operators, since for any nonzero
$\mathbf{x}$,
\begin{equation}
\mathbf{x}^T \mathbf{S} \mathbf{x} < 0 \quad \Rightarrow \quad \mathbf{x}^T \mathbf{A} \mathbf{x} = \mathbf{x}^T \mathbf{S} \mathbf{x} < 0,
\end{equation}
implying strictly decreasing energy in all directions. Diagonalizing the symmetric operator $\mathbf{S}$ yields real, strictly negative eigenvalues and an orthogonal eigenbasis. Hence, the system’s energy decays independently along these orthogonal modes. This negative-definite term can thus be interpreted as a generalized eddy-viscosity model that captures dissipative effects.
However, a purely negative-definite formulation cannot represent conservative or rotational behaviors, since those correspond to energy-preserving dynamics for which $\mathbf{x}^T \mathbf{A}\mathbf{x}=0$. The skew-symmetric component $\mathbf{R}$ addresses this limitation: it represents interactions that conserve energy, such as rotations, couplings, or advective terms in fluid dynamics \cite{gallavotti2002foundations}. By combining both components, we obtain the representation
\begin{equation}
\mathbf{A} = (\mathbf{K} - \mathbf{K}^T) - \mathbf{Q}^T \mathbf{Q},
\end{equation}
which spans the entire space of stable linear operators while remaining expressive enough to capture both dissipative and conservative effects. This serves as motivation for the proposed decomposition of the neural network architecture. This decomposition is not unique, since $\mathbf{A}$ remains the same if $\mathbf{K}$ is perturbed by an arbitrary symmetric matrix.}

During our testing, we observed the effect of the skew-symmetric term to be much larger than that of the negative-definite term, see Figure \ref{fig:skew_decomposition}. This further motivates its presence. From the computed energy contributions of the true closure term (see Figure \ref{fig:offline}), we believe backscatter is limited for reasonable coarse-graining factors. This means such a constrained closure model should be able to capture the energy behavior of the true closure term. 
In the upcoming sections, we will explain how the operators in this skew-symmetric architecture are built.

\subsubsection{Skew-symmetric term}

As described in \cite{VANGASTELEN2024113003}, $\mathcal{K}$ is constructed as follows:
\begin{equation}
    \mathcal{K}(\bmauh,\theta) = \mathcal{B}_1^T \mathbf{k}(\bmauh,\theta) \mathcal{B}_2,
\end{equation}
where $\mathbf{k}(\bmauh,\theta) = \text{diag}(\mathbf{k}_1,\mathbf{k}_2)\in \mathbb{R}^{2 \bar{N} \times 2 \bar{N}} $ is a matrix containing the neural network outputs $\mathbf{k}_1, \mathbf{k}_2 \in \mathbb{R}^{2 \bar{N}}$ on the diagonal. \R{rev_1_M_1}\revone{Here we choose to represent each matrix in the decomposition to be square. However, one is free to choose the dimensions of the matrices, as long as the $\mathcal{K}$ is square. Also, the sparsity of the matrices can be varied.
In our case, the underlying neural network architecture is a \gls{CNN}, as we employ a uniform grid. This means a \gls{CNN} is used to map $\bmauh$ to output channels $\mathbf{k}_1$ and $\mathbf{k}_2$. The matrices $\mathcal{B}$ are linear convolutional layers mapping from two input channels to two output channels, similarly to the convolutional layers introduced in section \ref{sec:CNN}. The $\mathcal{B}$ matrices thus contain four submatrices encoding convolutions.
A clear motivation for this specific decomposition of $\mathcal{K}$, besides sparsity and computational efficiency, is outlined in \ref{app:motivation}. The main idea is that the proposed decomposition resembles a simple advection operator for specific choices of $\mathcal{B}$ matrices, where the matrix $\mathbf{k}$ supplies enough degrees of freedom to freely traverse the energy-conserving solution space, by locally controlling the advection in each direction. 
Another option would be to use more output channels, i.e., $\mathbf{k}_1,\ldots,\mathbf{k}_n$ and construct a larger matrix $\mathbf{k}= \text{diag}(\mathbf{k}_1,\ldots,\mathbf{k}_n)$, where $n$ would be a hyperparameter. This would require the $\mathcal{B}$ matrices to be non-square and contain more submatrices encoding different convolutions. 
Such an extended architecture could possibly lead to faster convergence of the training procedure and more expressive power. In addition, a smaller $\mathbf{k}$ matrix, i.e. $\mathbf{k} \in \mathbb{R}^{c\times c}$ with $c << \bar{N}$, could also be considered. This would effectively apply the non-linear neural network operations in a reduced latent space, encoded by the $\mathcal{B}$ matrices \cite{bank2023autoencoders}. We consider both options outside the scope of the current work, but view them as interesting future research directions.} 

To satisfy momentum conservation, see \eqref{eq:mom_cons}, we require the sum of the convolution weights in the submatrices to be zero such that both $\mathcal{B}$ and $\mathcal{B}^T$ are in the nullspace of $\boldsymbol{\mathbbm{1}}_H$. To achieve this, \R{rev_3_comment_5}\revthree{we let the weights $\bar{b}_{kl}$ of such a convolution depend on a set of parameters} $b_{kl}$:
\begin{equation}
    \bar{b}_{kl} = b_{kl} - \frac{1}{(2r + 1)^2} \sum_{k,l=-r}^{r}  b_{kl},
\end{equation}
such that $ \sum_{k,l=-r}^{r} \bar{b}_{kl} = 0$ holds.
To extend the architecture to unstructured grids, the convolutions in the \gls{CNN} and in the $\mathcal{B}$ matrices can simply be replaced by graph convolutions.

\subsubsection{Negative-definite term}\label{sec:neg_def_term}

As described in \cite{VANGASTELEN2024113003}, $\mathcal{Q}$ is constructed as follows  
\begin{equation}
    \mathcal{Q}(\bmauh,\theta) = \mathbf{q}(\bmauh,\theta) \mathcal{B}_3,
\end{equation}
where  $\mathbf{q}(\bmauh,\theta) = \text{diag}(\mathbf{q}_1,\mathbf{q}_2)$ is also constructed from outputs of the \gls{CNN} $\mathbf{q}_1,\mathbf{q}_2 \in \mathbb{R}^{2 \bar{N}}$. This means the \gls{CNN} has in total four output channels to build the fields $\mathbf{k}_1,\mathbf{k}_2, \mathbf{q}_1, \mathbf{q}_2$. The parameters of the model include both the \gls{CNN} weights and the parameters of the $\mathcal{B}$ matrices. As these are all sparse operations, the model remains computationally efficient. Because the model only contains convolutions, it is translation equivariant. Furthermore, it can be safely applied to larger grids because it uses only local information. \R{rev_1_M_2}\revone{Similarly to the skew-symmetric term, this decomposition could be adjusted by extending the matrix containing the neural network outputs to multiple channels, i.e. $\mathbf{q} = \text{diag}(\mathbf{q}_1,\ldots,\mathbf{q}_n)$ A more in-depth motivation behind the proposed neural network architecture is presented in \ref{app:motivation}, where we relate the choice of $n=d$, where $d$ is the number of spatial dimensions, to a diffusion operator. In this way, the diagonal elements of $\mathbf{q}$ locally control the diffusivity.}

\subsection{Overview of the closure models}

In this section, we introduced a set of four closure models, namely the standard \gls{SMAG}, a \gls{CNN}, using a \gls{CNN} to predict a stress tensor and then taking the divergence (\acrshort{DIV}), and finally our \gls{SKEW}. In addition, we also compare to \gls{NC}, i.e.\  $\tilde{\mathbf{c}} = \mathbf{0}_H$.
An overview of the closure models and their properties is depicted in \R{rev_3_m_6}\revthree{Table \ref{tab:models}.}
    \begin{table}[]
\centering
\caption{Overview of the different closure models and their properties when combined with the coarse discretization. These closure models consist of the \acrfull{SMAG}, a \gls{CNN}, the divergence of a \gls{CNN} (\acrshort{DIV}), our \acrfull{SKEW}, and \acrfull{NC}.}\label{tab:models}
\begin{tabular}{lccccc}
% \hline
\toprule
       & \textbf{SMAG} & \textbf{CNN} & \textbf{DIV} & \textbf{SKEW} (ours) & \textbf{NC}  \\ 
       \midrule %\hline
Mass conservation & \cmark       & \cmark      & \cmark       & \cmark      & \cmark          \\ %\hline

Momentum conservation  & \cmark        & \xmark      & \cmark       & \cmark       & \cmark         \\ %\hline
Dissipative   & \cmark        & \xmark      & \xmark      & \cmark        & \cmark        \\ %\hline
\bottomrule
\end{tabular}
\end{table}

\section{Results}\label{sec:results}

\subsection{Experimental setup}\label{sec:experimental_setup}

To evaluate the closure models,, we consider two test cases: 2D decaying turbulence and Kolmogorov flow. The test cases are inspired by those considered in \cite{shankar2024differentiableturbulenceclosurepartial}. For both test cases we consider a periodic domain of $\Omega = [-\pi,\pi] \times [-\pi,\pi]$ and a viscosity of $\nu = \frac{1}{1000}$. Each velocity component of the flow is initialized by a random initial condition only containing energy in the low wavenumbers (below $\kappa_\text{max} = 10$):
\begin{align}\label{eq:init_cond}
    \mathbf{u}(\mathbf{x}, 0) &=\text{Re} \left( \begin{bmatrix}
         \sum_{ \{ \boldsymbol{\kappa} \in \mathbb{Z}^2 | 0 < ||\boldsymbol{\kappa}||_2 < \kappa_{\text{max}}\}} c^u_{\boldsymbol{\kappa}} e^{i {\boldsymbol{\kappa}} \cdot \mathbf{x}}  \\ \sum_{ \{ \boldsymbol{\kappa} \in \mathbb{Z}^2 | 0 < ||\boldsymbol{\kappa}||_2 < \kappa_{\text{max}}\}} c^v_{\boldsymbol{\kappa}} e^{i {\boldsymbol{\kappa}} \cdot \mathbf{x}} 
    \end{bmatrix} \right),
\end{align}
where the real and complex components of the coefficients $c_{\boldsymbol{\kappa}}^u,c_{\boldsymbol{\kappa}}^v  \in \mathbb{C}$ are sampled uniformly from the interval $(-1,1)$. 
\R{rev_3_m_9}\revthree{Finally, the coefficients are scaled such that the normalized kinetic energy at \( t = 0 \), namely 
\[
    \frac{1}{2|\Omega|}\int_\Omega ||\mathbf{u}(\mathbf{x},0)||_2^2 \, \mathrm{d}\Omega,
\]
equals \(1.2\). Increasing this initial energy amplifies the characteristic velocity scales of the flow, thereby increasing the effective Reynolds number \( \mathrm{Re}_\text{eff} = U L / \nu \), where \(U\) is a representative velocity magnitude and \(L\) a characteristic length scale of the largest eddies. A higher \(\mathrm{Re}_\text{eff}\) leads to a broader separation of scales and stronger non-linear interactions, which make closure modeling more challenging. The chosen value of \(1.2\) was found, through preliminary numerical tests, to yield sufficiently rich turbulent dynamics while remaining numerically stable and allowing meaningful comparison between closure models.
Before the simulation starts, the sampled initial condition is projected onto a divergence-free basis.}

The training data set consists of five simulations starting from an initial condition sampled in this manner. For the \gls{DNS} we use a computational grid of resolution $2048 \times 2048$ and a time step size $\Delta t = 2 \times 10^{-4}$, as in \cite{shankar2024differentiableturbulenceclosurepartial}. For the time integration of both the \gls{DNS} and the closure model-based simulations, we use a 4\textsuperscript{th}-order Runge-Kutta integration scheme \cite{RK4_Butcher:2007,SANDERSE_RK4}. For training purposes, we save a snapshot every $10$ time steps until $t = 5$. Regarding coarse-graining, we consider three different resolutions, namely $32 \times 32$, $64\times 64$, and $128 \times 128$, and a time step size of $\overline{\Delta t} = 10 \Delta t$, as coarser grids allow for larger time steps \cite{de2013courant_CFL,CFL}. For each coarse-graining factor, the machine learning models are trained to reproduce the true solution, i.e., minimize \eqref{eq:traj_loss}, for $n = 5$ time steps. To optimize the closure model parameters, we use the ADAM optimization algorithm \cite{kingma_adam} with a learning rate of $10^{-3}$, decay rates for the first and second momentum estimates at $0.9$ and $0.999$, respectively, and a mini-batch size of $20$. We optimize for in total $500$ epochs. These hyperparameter settings resulted in smooth convergence, see \ref{sec:convergence_appendix}. We consider further hyperparameter studies to be outside the scope of this research, as this research focuses on the neural network architecture. We implemented the closure models in the Julia programming language \cite{bezanson2017julia} using the Flux.jl package \cite{Flux.jl-2018,flux_innes:2018}. For each of the \gls{CNN}-based closure we use four input channels, namely $\bmauh$ and $\mathbf{m}_H(\bmauh)$, each containing two channels. The intermediate layers of the \glspl{CNN} each contain 32 channels, with a total of four intermediate layers (same amount as in \cite{Agdestein_2025}). The convolutions in each layer have a radius of $r=2$.
\R{rev_1_m_3}\revone{The number of parameters is dominated by the number of hidden layers and channels, which is the same for each of the neural network-based closure models. In this way, the comparison is carried out as fairly as possible.}
In between the hidden layers, we employ a ReLU activation function and a linear activation function at the final layer \cite{RELU}. The purely \gls{CNN}-based closure simply has two output channels to model the closure term in each spatial direction. For \acrshort{DIV} we have three output channels, two for the diagonal elements of the stress tensor and one for the off-diagonal ones, due to the symmetry of the true stress tensor \cite{sagaut2006large,Pope_2000}. For \gls{SKEW} the underlying \gls{CNN} has four output channels corresponding to $\mathbf{k}_1,\mathbf{k}_2,\mathbf{q}_1,\mathbf{q}_2$. The convolutions in the $\mathcal{B}$ matrices are chosen to have a convolution radius of $r=2$. Training of a single neural network takes roughly two hours on an A100 GPU of the Dutch National supercomputer Snellius \cite{surf_snellius}. Regarding the optimization of the Smagorinsky constant, we choose the constant value that minimizes the $L_2$-norm of the energy spectra for the training data at $t=2$ in $\log_{10}$ space. The energy spectra are determined by computing a \gls{FFT} of the velocity field. The wavenumbers are then divided into dyadic bins, and the energy belonging to the wavenumbers in the bin is summed to produce the spectrum. This procedure is described in \cite{Agdestein_2025,Dyadic_bins}. The considered values for the Smagorinsky constant are $0.0$ to $0.30$ in intervals of $0.01$. The obtained optimal values for $C_s$ at each resolution are $0.23$, $0.22$, and $0.18$ for $32 \times 32$, $64 \times 64$, and $128 \times 128$, respectively. These values are within the range of what is typically used for 2D turbulence, namely around 0.18 \cite{smag_constant_value_1,smag_constant_value_2}.

\subsection{Decaying turbulence}

\begin{figure}
    \centering
\includegraphics[width = 0.48\textwidth]{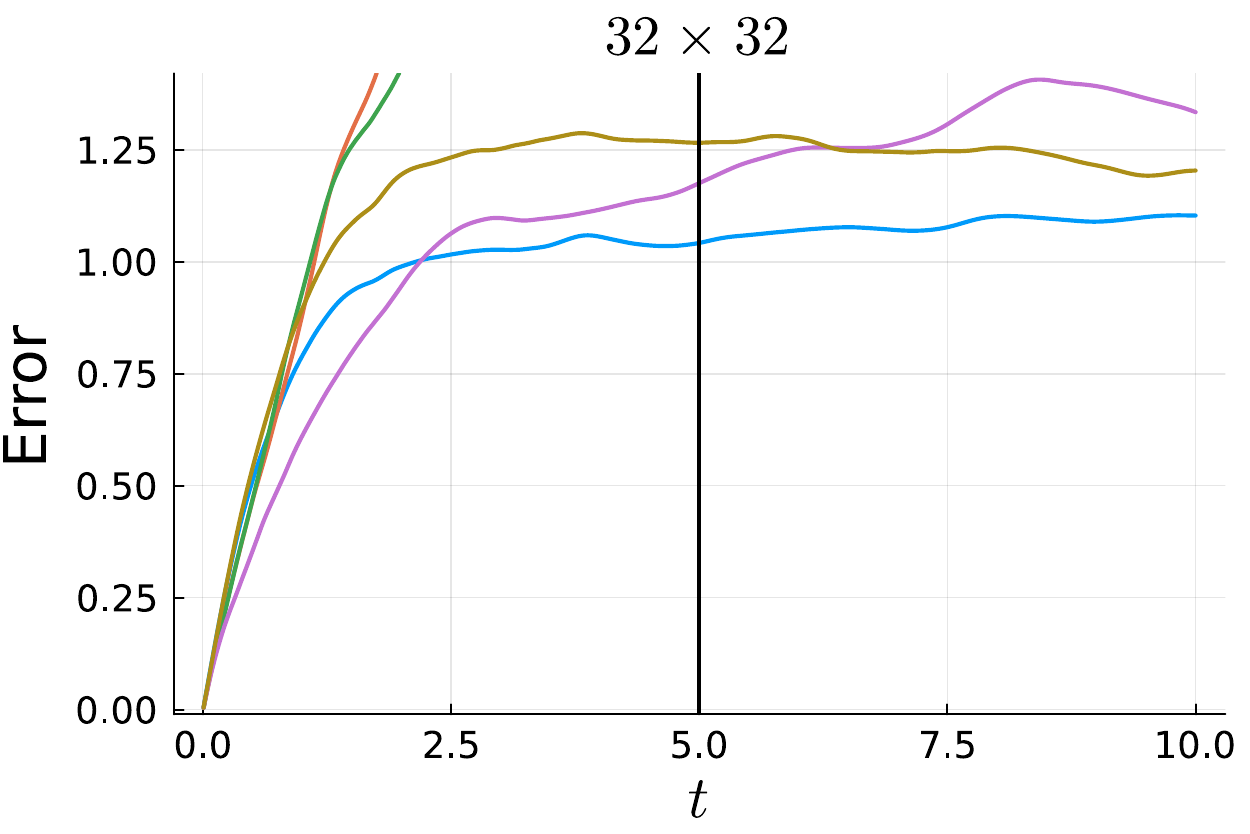}
\includegraphics[width = 0.48\textwidth]{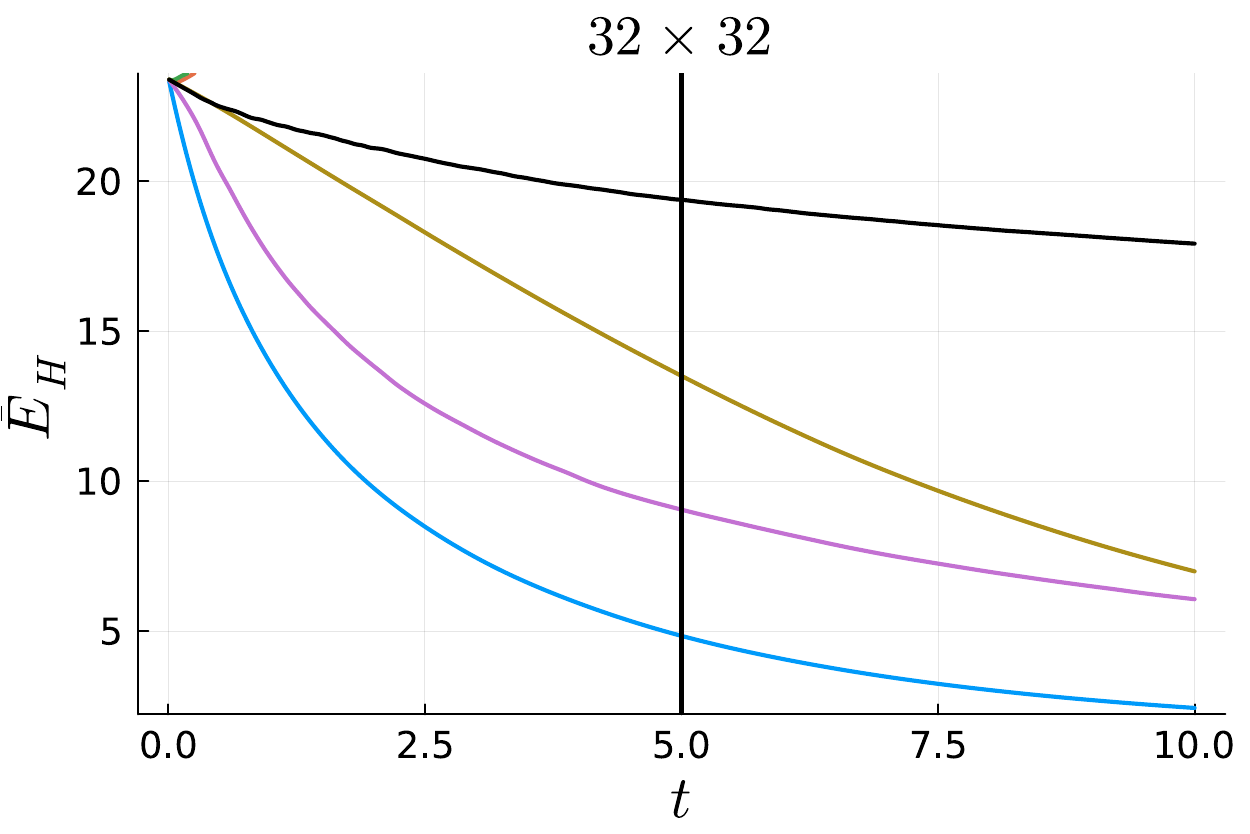}
\includegraphics[width = 0.48\textwidth]{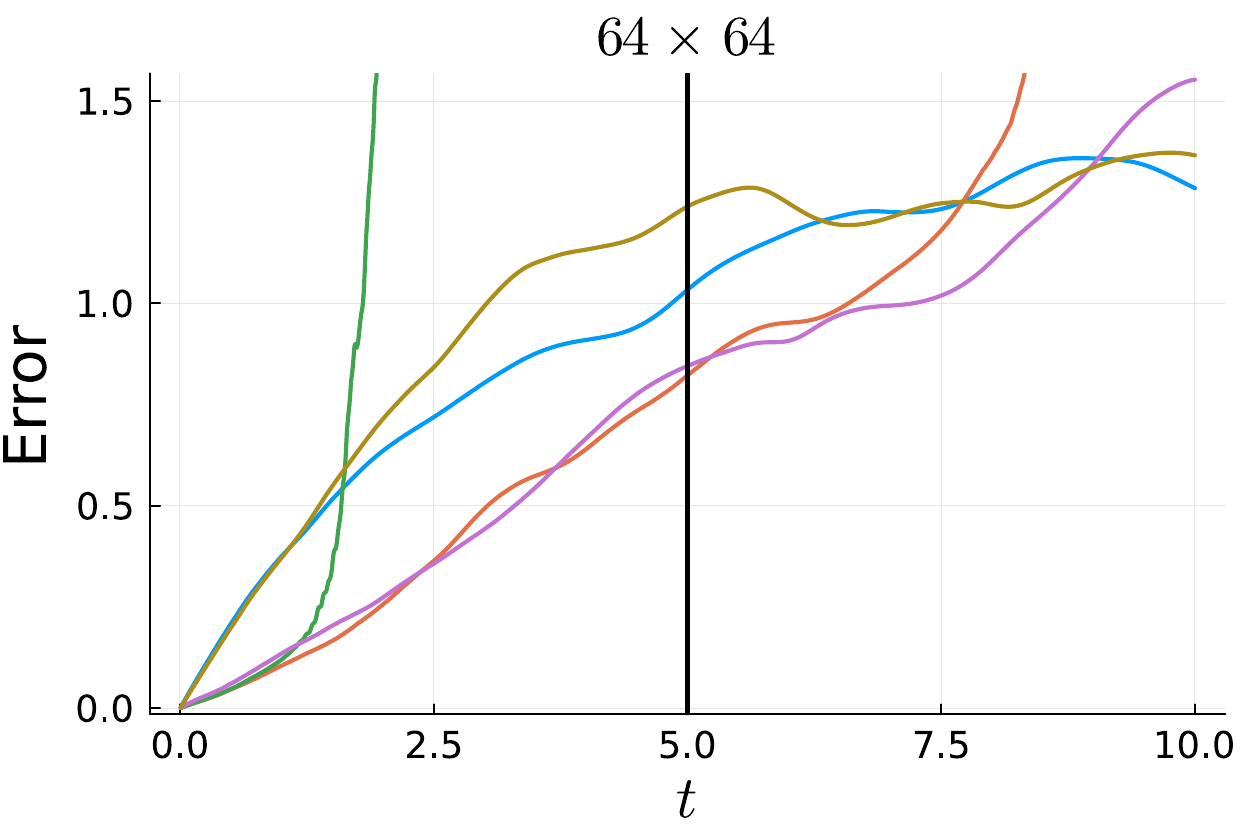}
\includegraphics[width = 0.48\textwidth]{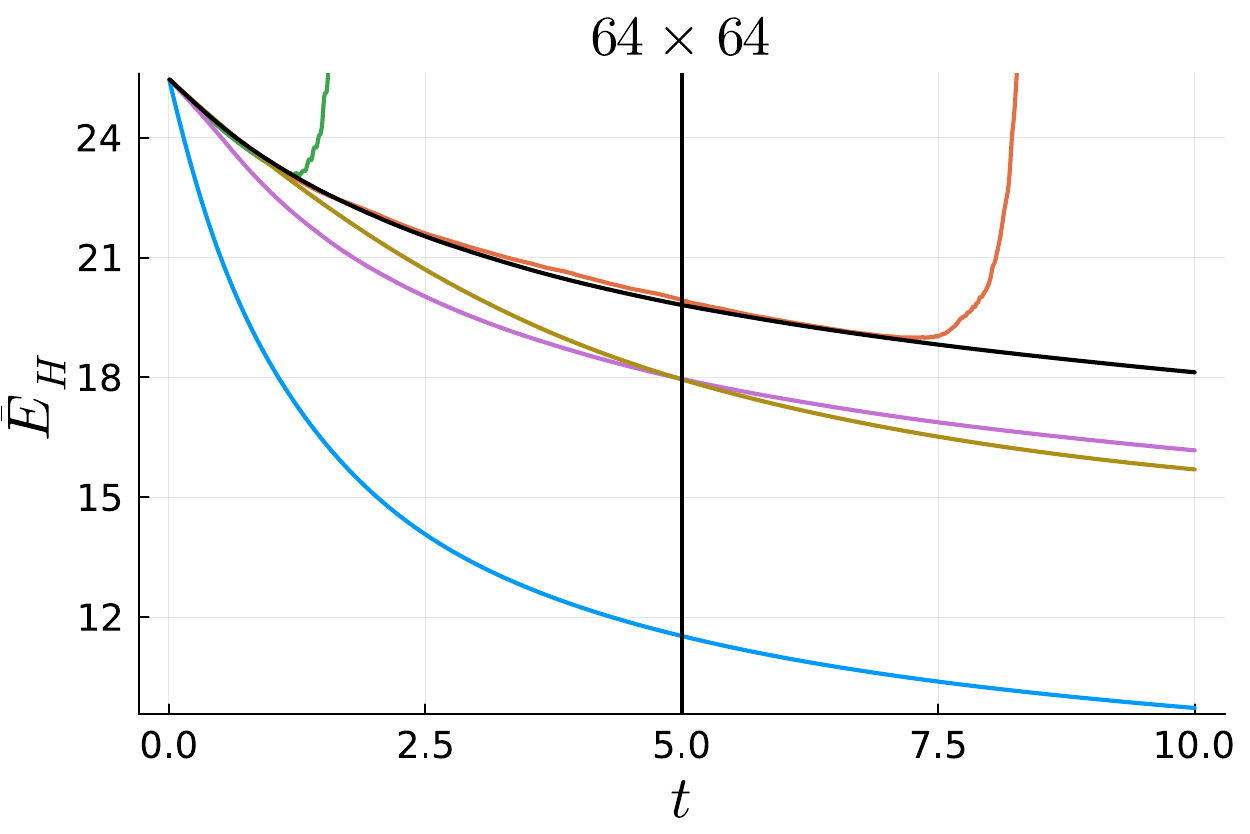}
\includegraphics[width = 0.48\textwidth]{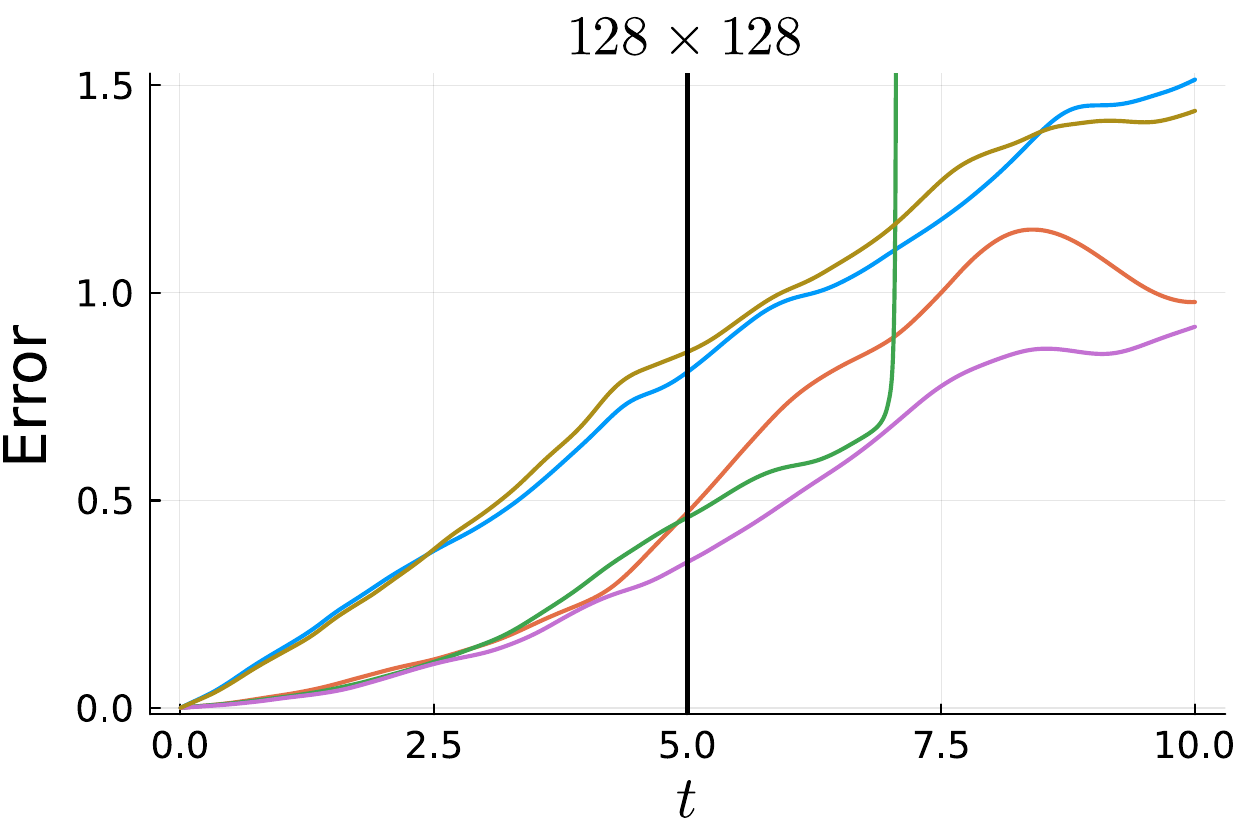}
\includegraphics[width = 0.48\textwidth]{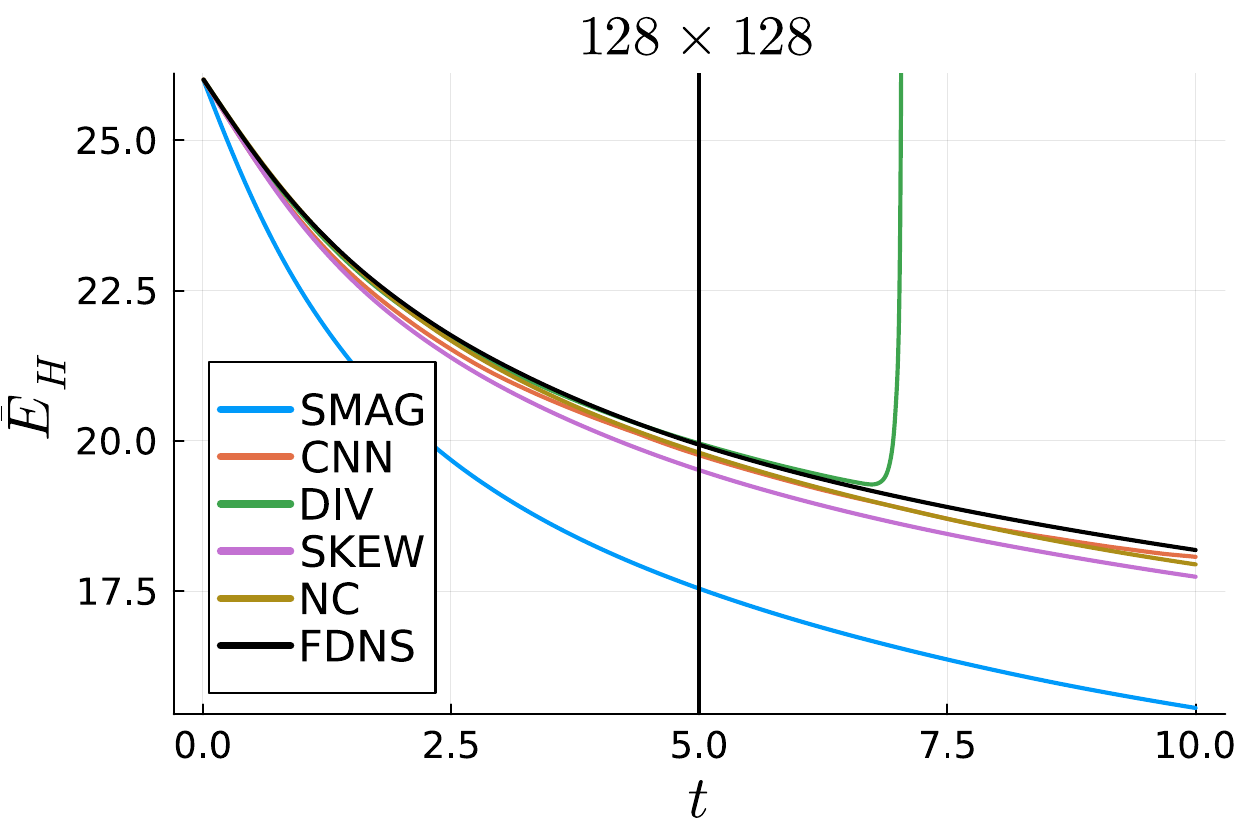}

\caption{(Left) Error for each coarse-graining factor as a function of time for the decaying turbulence test case. (Right) Resolved energy trajectories for each coarse-graining factor. For an overview of the methods, see Table \ref{tab:models}. The black vertical line indicates $t=5$. Everything to the right of this line corresponds to extrapolation in time.}\label{fig:error_and_energy}
\end{figure}

\begin{figure}
    \centering
\includegraphics[width = 1\textwidth]{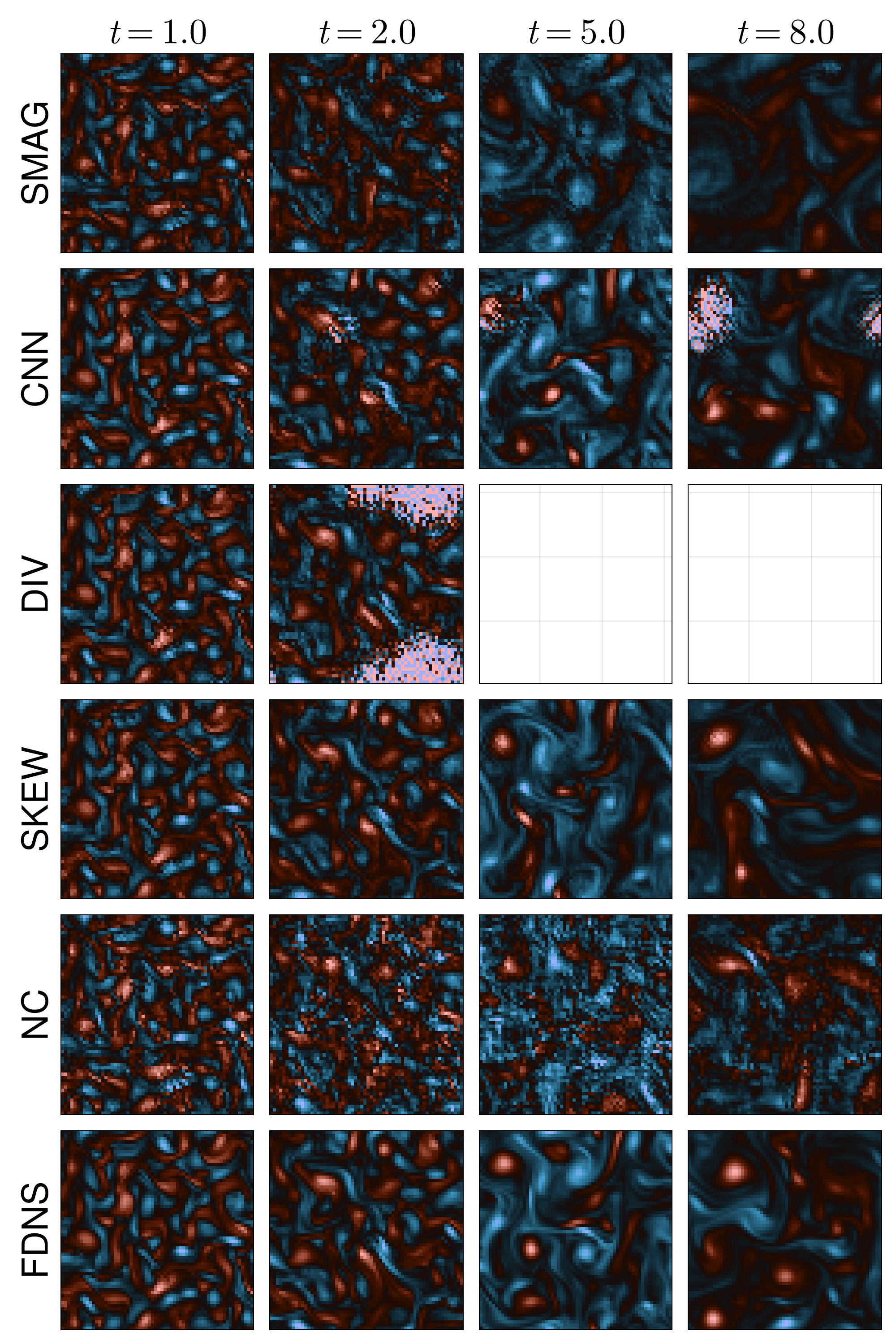}

\caption{Vorticity fields at each point in time for each of the closure models on a $64 \times 64$ grid. Simulations correspond to the decaying turbulence test case. Blank boxes indicate an unstable simulation.}\label{fig:64_64}
\end{figure}

The first test case we consider is decaying turbulence, i.e., there is no forcing. Here we consider a single simulation starting from a randomly generated initial condition, generated according to \eqref{eq:init_cond}. The simulations are carried out up to the final time $t=10$. Note this is twice as long as present in the training data. Therefore, the second half of the simulation corresponds to extrapolation in time. For each coarse-graining factor, the resolved energy and error trajectories are presented in Figure \ref{fig:error_and_energy}.
The error is defined as
\begin{equation}\label{eq:error}
    \text{error} := \sqrt{\frac{||\bmauh - \bmauh^\text{model}||^2_2}{||\bmauh||_2^2}},
\end{equation}
where $\bmauh$ is the \gls{FDNS} result and $\bmauh^\text{model}$ is the model prediction.
We observe that initially the unconstrained machine learning approaches (\gls{CNN} and \acrshort{DIV}) produce the best energy trajectories for resolutions $64 \times 64$ and $128 \times 128$, whereas \gls{SKEW} is too dissipative and the Smagorinsky model is even more dissipative. However, as the simulation progresses, the unconstrained machine learning methods often suddenly become unstable. For a resolution of $32 \times 32$, these approaches tend to diverge rather quickly, whereas finer resolutions can stay stable over a more extended time period. That said, this seems merely a postponement of the inevitable, as unconstrained machine learning acting on a finer resolution remains highly prone to a sudden and unpredictable catastrophic failure. One can hope for a one-off stable simulation, as shown by the $128\times 128$ resolution \gls{CNN}. This particular simulation remained stable during the entire considered time period, producing a good energy trajectory and an improved error with respect to \gls{NC} and \gls{SMAG}. 
However, in the vorticity fields presented in Figure \ref{fig:64_64}, we already notice a build-up of numerical noise for the \gls{CNN} at the end of the simulation, which may well result in instabilities in the future. In fact, in Section \ref{sec:consistency} we show that this is indeed the expected outcome. As such, the purely data-driven, physics-blind machine learning methods considered here are unsuitable as a viable closure modeling approach for long-term predictions, incapable of outperforming existing physics-based closure models.  %This build-up of numerical noise likely corresponds to the increased error observed around $t=7.5$.

\begin{figure}
    \centering

\includegraphics[width = 0.42\textwidth]{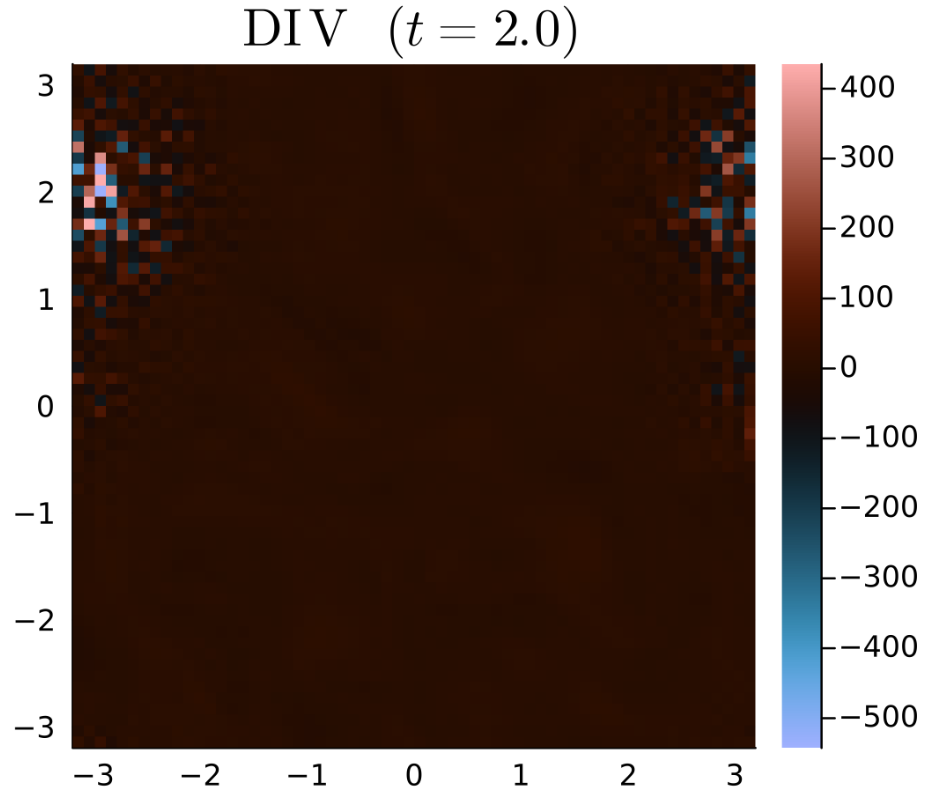}
\includegraphics[width = 0.48\textwidth]{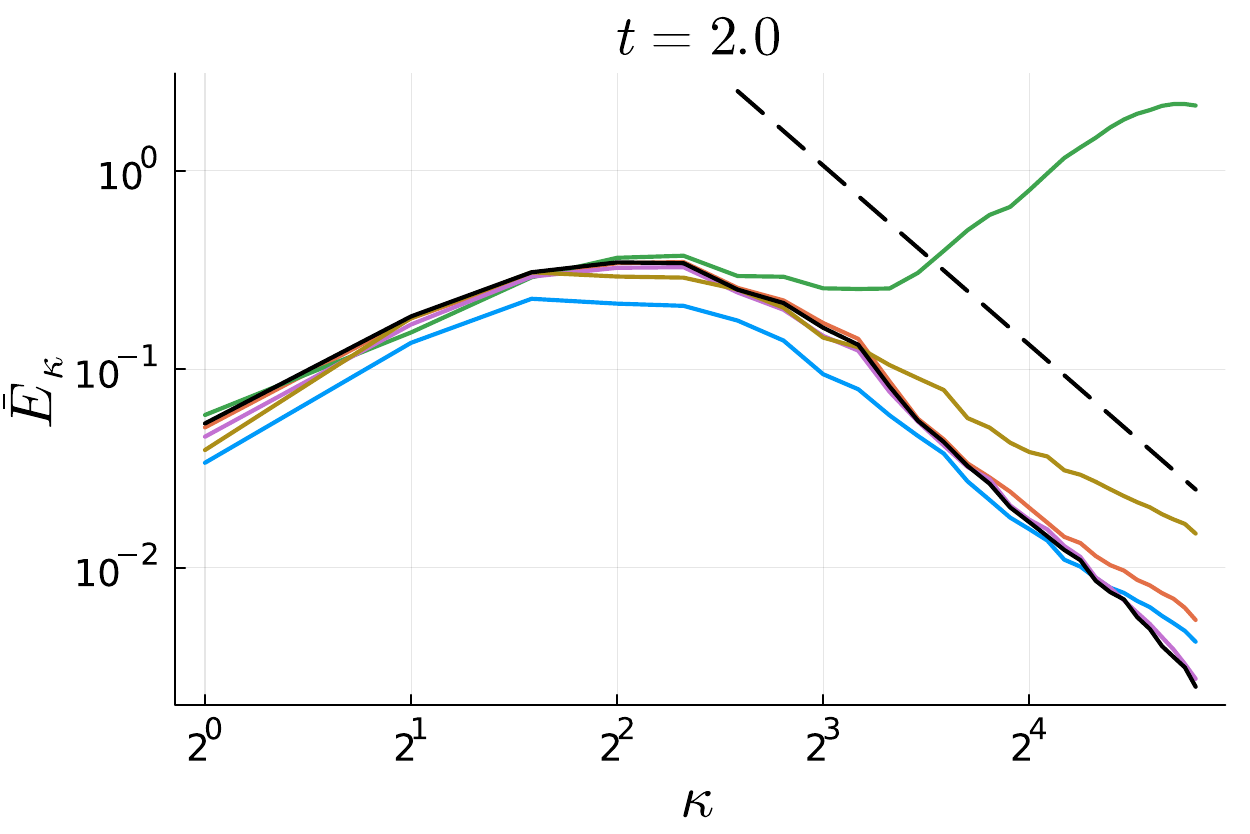}
\includegraphics[width = 0.42\textwidth]{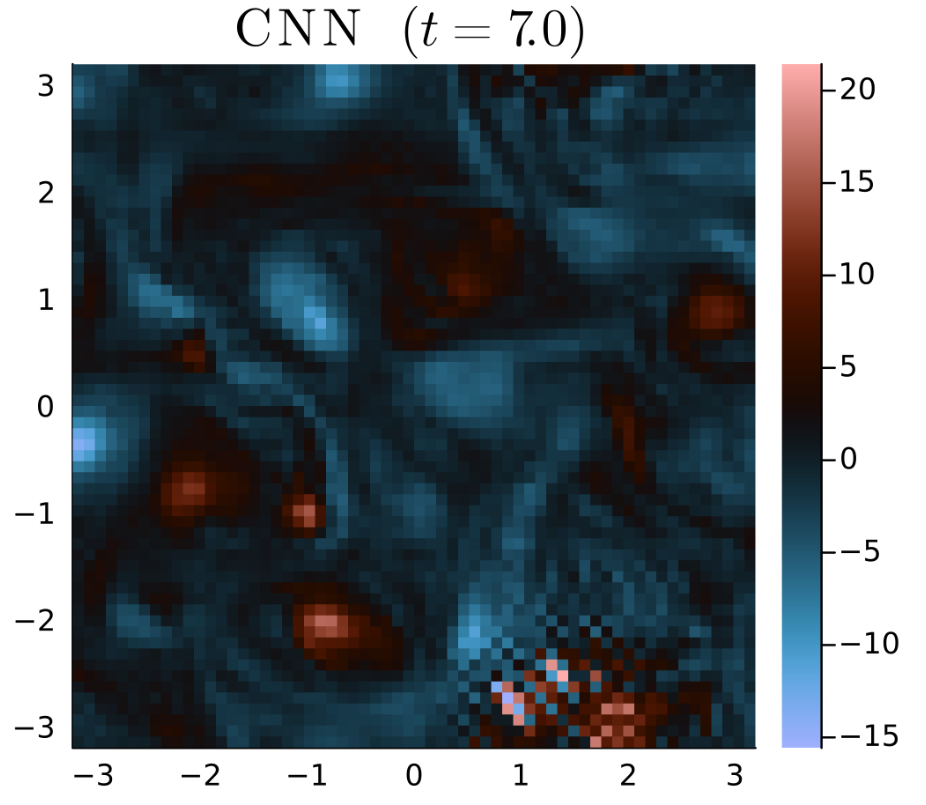}
\includegraphics[width = 0.48\textwidth]{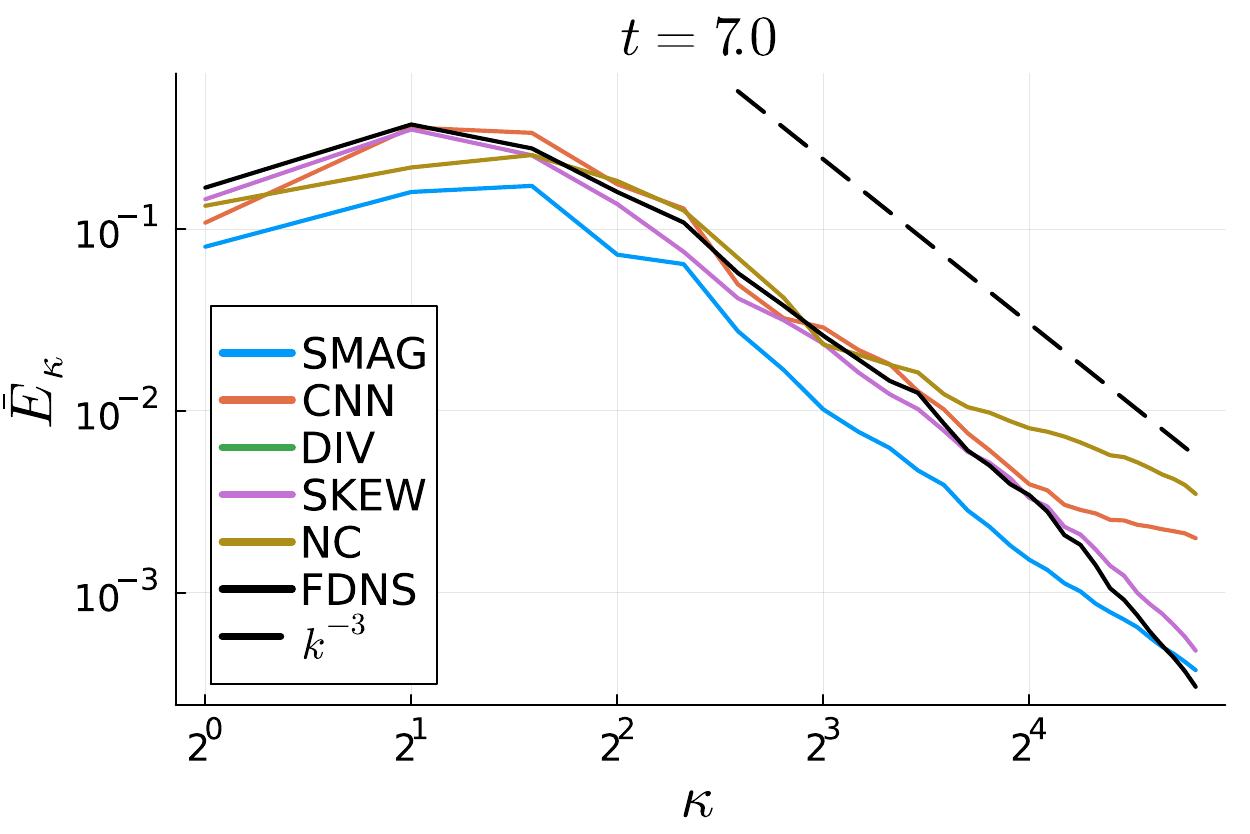}

\caption{(Left) Two snapshots where we see numerical oscillations occurring for \acrshort{DIV} and \gls{CNN}. The oscillations eventually result in instabilities. (Right) Energy spectra at the time of these snapshots. The numerical oscillations cause a clear increase in energy in the high wavenumbers.}\label{fig:examples}
\end{figure}

\begin{figure}
    \centering

\includegraphics[width = 0.48\textwidth]{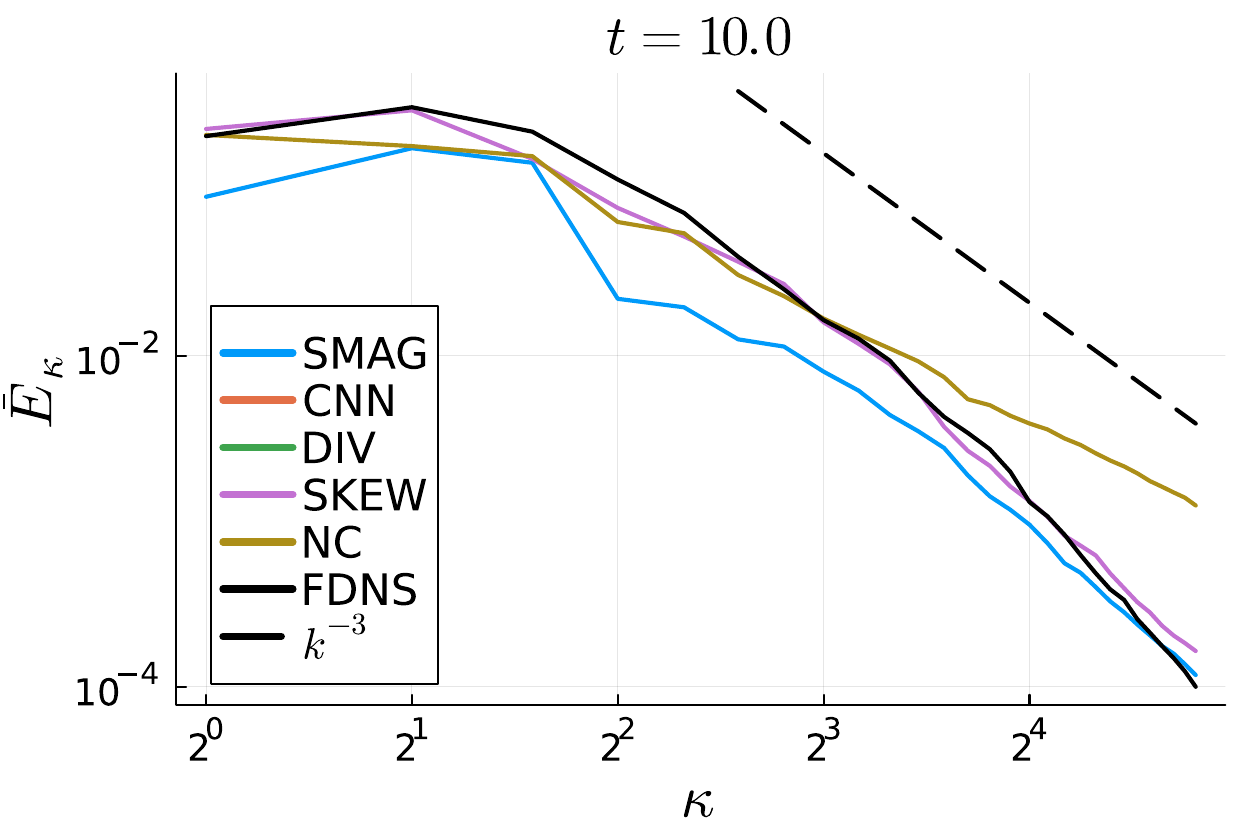}

\caption{Energy spectra at the end of the decaying turbulence simulation at $t=10$. Both the \gls{CNN} and \acrshort{DIV} have become unstable at this point and are therefore omitted from the figure.}\label{fig:spectrum_final}
\end{figure}

On the other hand, for our constrained approach \gls{SKEW}, we find it consistently provides improved error trajectories with respect to \gls{NC}, except at a resolution of $32 \times 32$. However, it is too dissipative, as stated earlier. This likely comes from the fact that it is constrained to not create energy. This means energy is solely redistributed through the skew-symmetric term and dissipated through the negative-definite term. Especially at large coarse-graining factors, this becomes more problematic, see resolution of $32 \times 32$. This is likely caused by the fact that backscatter is more prevalent, see Figure \ref{fig:offline}, as more information is being discarded to the subgrid scales. At a resolution of $64 \times 64$ \gls{SKEW} is also too dissipative, however, it does produce an improved error trajectory with respect to the other closures, even in the extrapolation region $t \in (5,10]$. However, near the end of the simulation, the error becomes larger than for \gls{NC} and \gls{SMAG}. Later in this section, we consider the energy spectra to compare the velocity fields in a more statistical fashion, see Figure \ref{fig:examples} and Figure \ref{fig:spectrum_final}. 

Carrying out the \gls{DNS} simulation took roughly 45 minutes on an A100 GPU, whereas the closure model-based simulations took between 3 and 4 minutes (with a training time of roughly two hours), for all closure models, see Table \ref{tab:computation_times}. This amounts to a computational speed-up of more than $10\times$ with respect to the \gls{DNS}. For these coarse resolutions, the evaluation time seems to be dominated by computational overhead, as we did not observe a big difference in computation time between the different closure models and resolutions. All closure model-based simulations require more computation time than \gls{NC}, where \gls{SKEW} consistently takes the most time. This is likely caused by the additional convolutions in the $\mathcal{B}$ matrices.
\begin{table}[]
\centering
\caption{Computation time in seconds for the different closure models for the decaying turbulence test case. For an overview of the methods, see Table \ref{tab:models}.}\label{tab:computation_times}
\begin{tabular}{ccccccc}
% \hline
\toprule
$\bar{N}$        & \textbf{SMAG} & \textbf{CNN} & \textbf{DIV} & \textbf{SKEW} & \textbf{NC} & \textbf{DNS} \\ %\hline
\midrule
$32\times 32$   & 182.69       & 186.64      & 189.99      & 211.39       & 170.42     & 2559.88     \\ %\hline

$64 \times 64$   & 210.46       & 216.26      & 203.76      & 225.79       & 165.27    & 2559.88     \\ %\hline
$128 \times 128$ & 195.29       & 200.06     & 217.75      & 243.32      & 187.82     & 2559.88     \\ %\hline
\bottomrule
\end{tabular}
\end{table}
%\end{comment}
Next, we look at the scalar vorticity field $\omega = (\nabla \times \mau)_3$, at different points in time, produced by the closure models. For a resolution of $64 \times 64$, these are depicted in Figure \ref{fig:64_64}. 
The other resolutions are depicted in \ref{sec:plots_appendix}. \R{rev_3_M_3_b}\revthree{For the remainder of this text, we will stick to a resolution of $64 \times 64$, as we believe this is a nice middle ground between the poor results obtained for $32 \times 32$ and almost perfect reproduction for $128 \times 128$. This coincides with the coarse-graining factor applied in \cite{shankar2024differentiableturbulenceclosurepartial}.}
Regarding the vorticity fields, we find that initially all three machine learning closures nicely match the \gls{FDNS} result. However, as the simulation progresses, we observe a build-up of numerical noise for the unconstrained machine learning closures. This eventually leads to unstable simulations. For \gls{SKEW} this does not happen. Even after diverging from the \gls{FDNS}, it still produces smooth results which seem to match the \gls{FDNS} on a qualitative level. The instabilities in the unconstrained machine learning approaches can possibly be alleviated by adding more training data or adding noise to the training \cite{clipping_and_instability_and_dissipation_CNN_Guan_2022,beck3}. However, we argue that given the same amount of training data, using \gls{SKEW} has clear stability benefits.

To assess the physical consistency of the produced trajectories, we compare the energy spectra at different points during the simulation. \R{rev_3_M_4_a}\revthree{This is done, as a pointwise error, such as presented in Figure \ref{fig:error_and_energy}, no longer provides relevant information after the solutions diverge from the \gls{FDNS} \cite{shankar2024differentiableturbulenceclosurepartial}.}. The spectra are depicted in Figure \ref{fig:examples}.
Here we observe that \gls{SMAG} is too dissipative at the large scales, but nicely reproduces the $\kappa^{-3}$ decay of the energy spectrum expected from 2D turbulence \cite{k-3slope}. Regarding \gls{SKEW} we find it has a better overall fit with the \gls{FDNS}, as compared to \gls{SMAG}, for both the high and low wavenumbers. Regarding the unconstrained machine learning approaches, we observe a clear buildup in energy in the large wavenumbers. From the depicted vorticity fields, we can clearly see the numerical noise responsible for this.

In Figure \ref{fig:error_and_energy}, we observed that \gls{SKEW} reaches a larger error than \gls{NC} and \gls{SMAG} at the end of the simulations. To assess whether or not the simulation produced by \gls{SKEW} is still physically consistent, we consider the energy spectrum at this point. This is depicted in Figure \ref{fig:spectrum_final}.
Here we observe that even though the error, see \eqref{eq:error}, is larger, \gls{SKEW} still produces an energy spectrum that more closely matches the \gls{FDNS} spectrum, as compared to \gls{SMAG} and \gls{NC}. It also produces the expected $\kappa^{-3}$ slope. \gls{SMAG} also achieves this, but underestimates the energy in all wavenumbers, whereas \gls{NC} suffers from a build-up of energy in the large wavenumbers. The latter likely corresponds to numerical noise, due to the coarseness of the grid.  
\R{rev_3_m_13}\revthree{After a while \gls{SKEW} diverges from the \gls{FDNS}. This behavior is expected given the characteristic sensitivity on initial conditions of the two-dimensional Navier–Stokes equations in the turbulent regime \cite{feng2017unpredictability}. Nevertheless, the \gls{SKEW} model continues to produce physically consistent statistics and spectra, even when extrapolating in time.}

\begin{figure}
    \centering
\includegraphics[width = 0.48\textwidth]{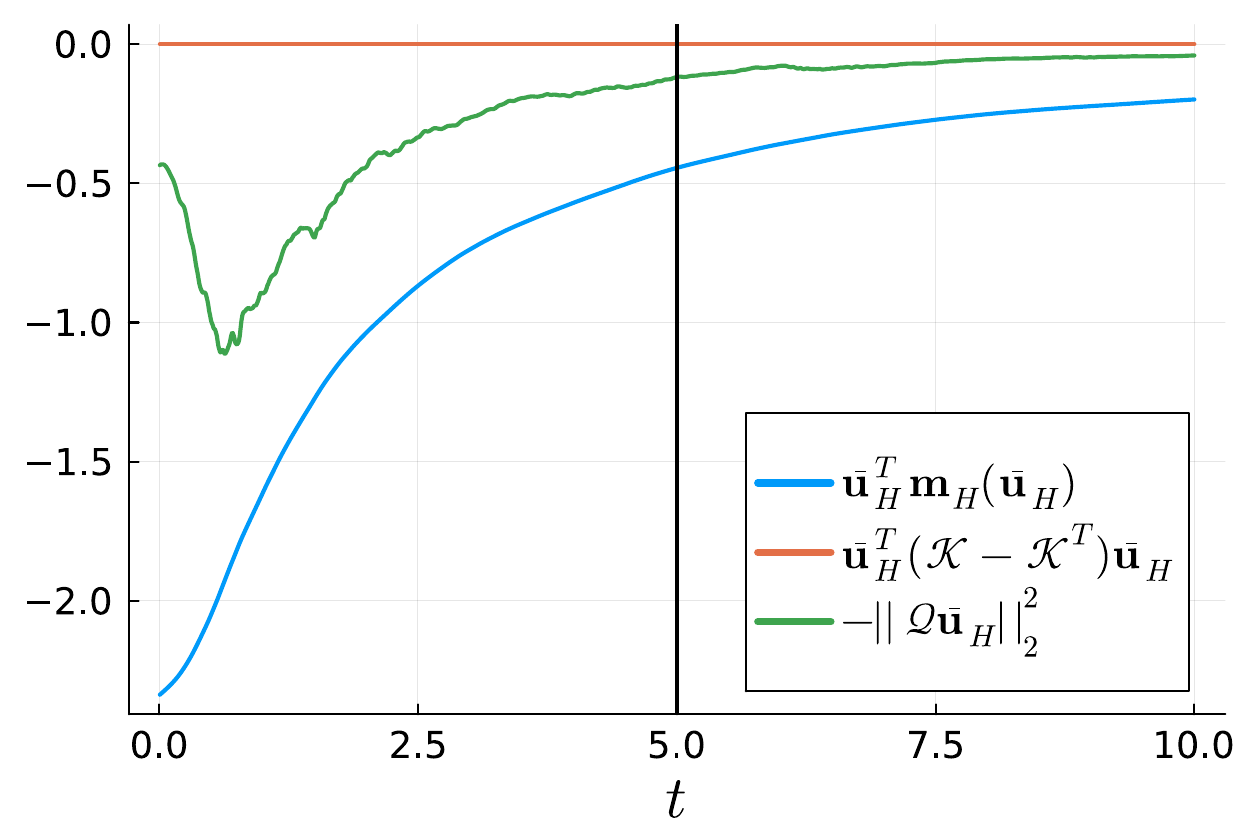}
\includegraphics[width = 0.48\textwidth]{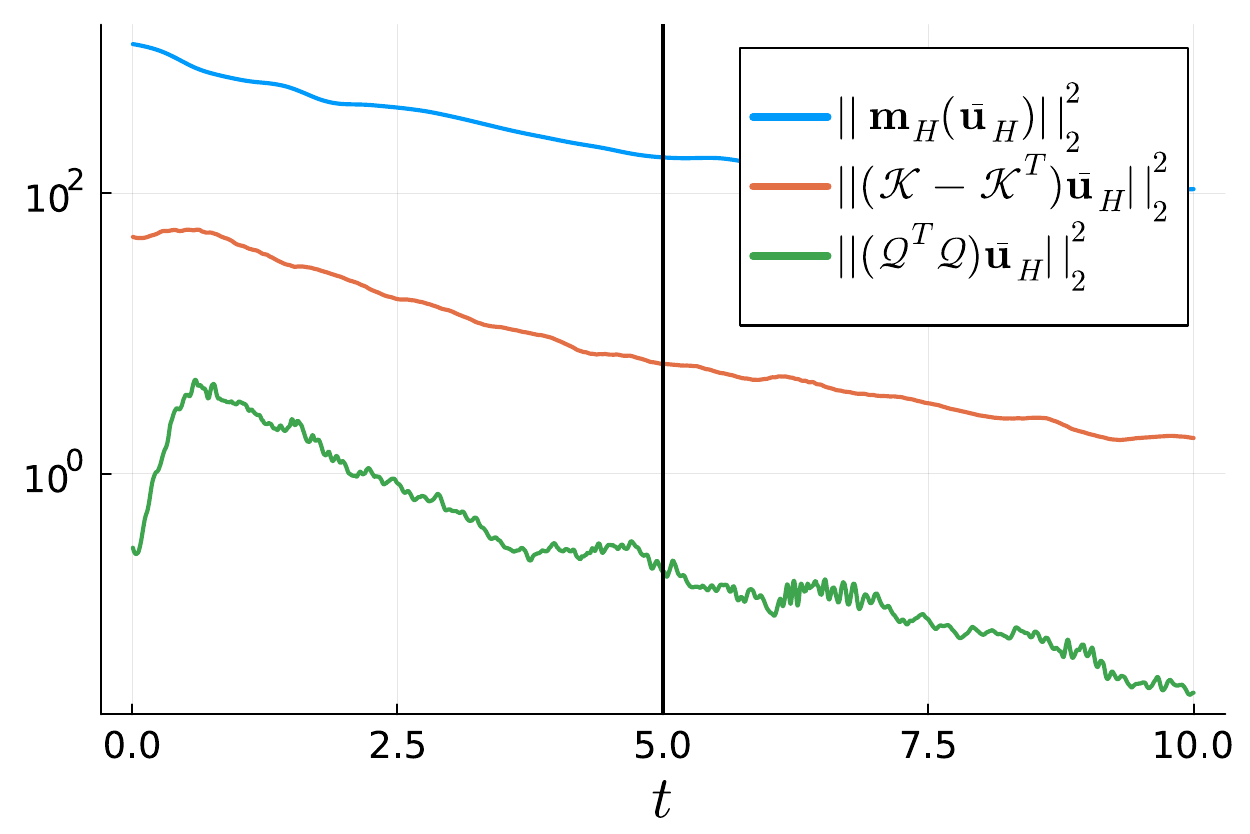}
\caption{(Left) Energy contribution for each of the terms in the \gls{SKEW} architecture along with the contribution of the coarse discretization. (Right) Magnitude of each of the terms. The black vertical line indicates $t=5$. Everything to the right of this line corresponds to extrapolation in time.}\label{fig:skew_decomposition}
\end{figure}

Finally, we consider the contributions of both the skew-symmetric and negative-definite term in the \gls{SKEW} architecture. We consider both the energy contribution and the magnitude of the terms. This is depicted in Figure \ref{fig:skew_decomposition}.
The first observation is that the energy contribution of the skew-symmetric term is indeed zero, as it should be. Furthermore, we find that the negative-definite term is slightly less dissipative than the coarse discretization. \R{rev_3_M_5}\revthree{The trajectory also has a different shape; the dissipation coming from the negative-definite term peaks around $t = 1$, whereas the dissipation from the coarse discretization decreases smoothly over time. The shape of the dissipation trajectory of the negative-definite term might arise from the transition of the flow from one physical regime to another. More specifically, the initial condition contains energy only at low wavenumbers. This means the flow undergoes a transient period before the formation of its characteristic slope, see Figure \ref{fig:examples}.}
 Next, we look at the magnitude of the different terms. Here we find that the skew-symmetric term has a larger magnitude than the negative-definite term. This means it has a more significant impact on the simulation. This supports the use of a skew-symmetric term in the closure model.

\begin{figure}
    \centering
    \includegraphics[width=0.48\textwidth]{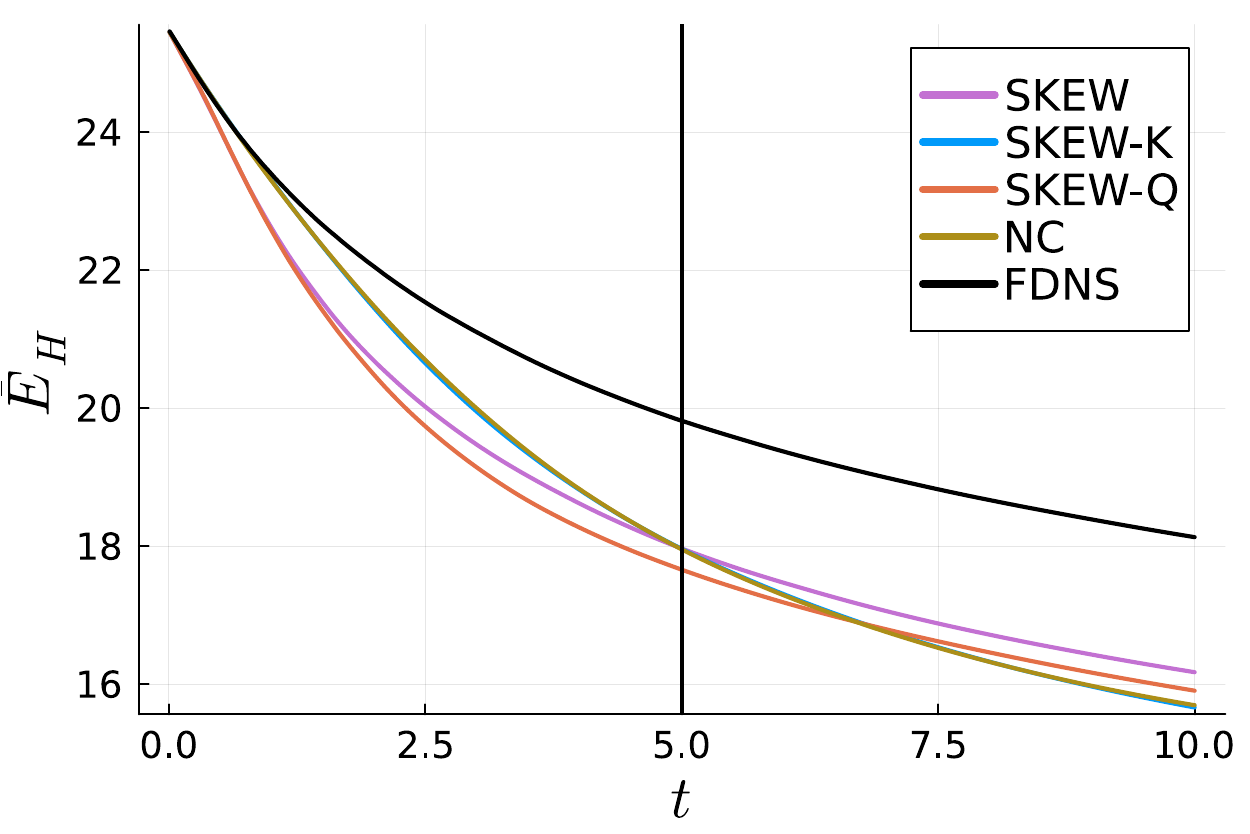}
    \includegraphics[width=0.48\textwidth]{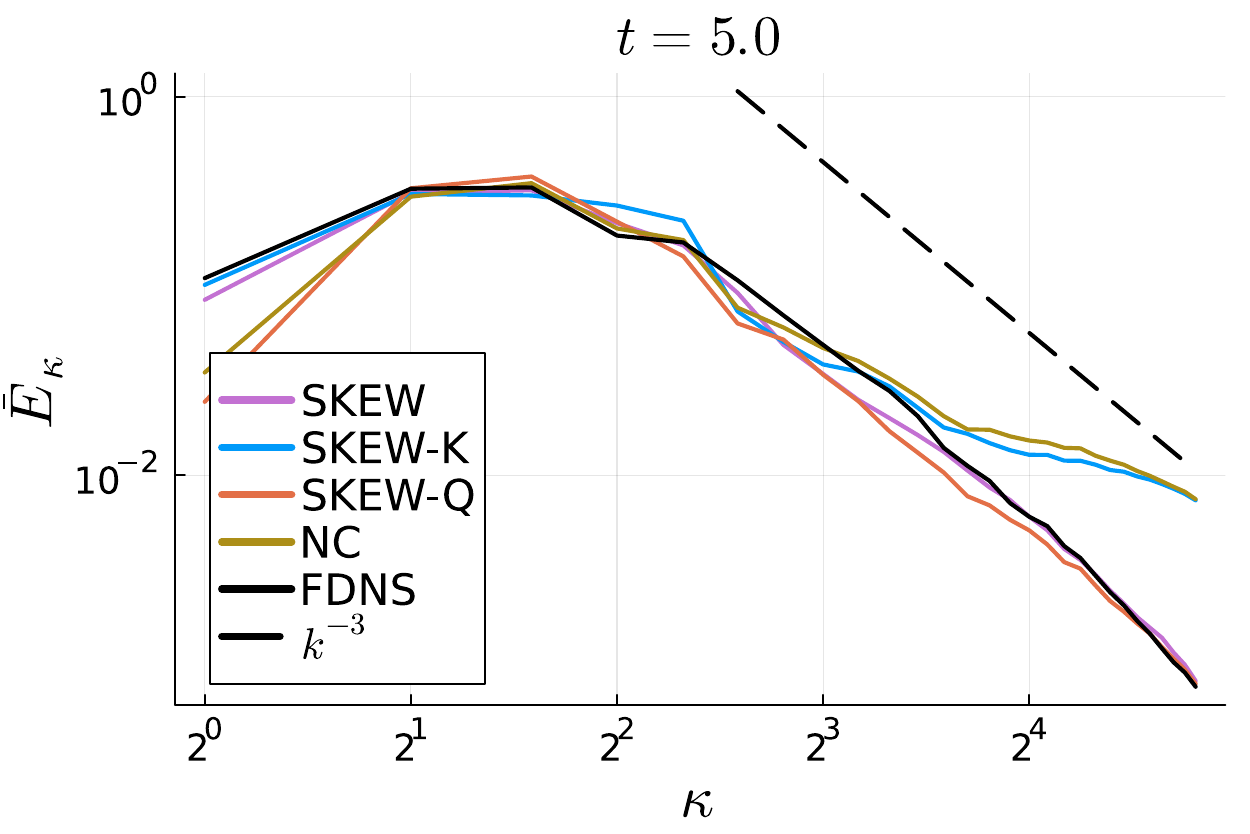}
    \caption{(Left) Resolved energy trajectories for the full \gls{SKEW} closure model, 
    SKEW-K (skew-symmetric term only), and SKEW-Q (dissipative term only). The black vertical line indicates $t=5$. Everything to the right of this line corresponds to extrapolation in time. (Right) Energy spectra halfway through the decaying turbulence simulation at $t=5$.}
    \label{fig:terms}
\end{figure}

\R{rev_3_M_2_e}\revthree{To conclude this section, we also examine the effect of omitting either the skew-symmetric term or the negative-definite term from the \gls{SKEW} architecture. To this end, we perform the decaying turbulence simulation with one of the two terms set to zero. The resulting resolved energy trajectories and energy spectra are shown in Figure \ref{fig:terms}.}
\revthree{We observe that omitting the negative-definite term leads to markedly less dissipative behavior, as expected. The resulting energy trajectory is nearly identical to that of \gls{NC}. The corresponding energy spectrum matches the \gls{FDNS} result well at low wavenumbers; however, at high wavenumbers, a clear build-up of energy appears, resembling the \gls{NC} spectrum and indicating numerical noise. Conversely, omitting the skew-symmetric term produces more dissipative behavior and yields physically consistent energy levels at high wavenumbers, with the characteristic $\kappa^{-3}$ slope accurately recovered. Nonetheless, this variant over-dissipates energy at the low wavenumbers.}

\revthree{From these simulations, we conclude that the skew-symmetric term is essential for accurately capturing the energy at low wavenumbers, whereas the negative-definite term plays a crucial role in suppressing numerical noise and recovering the characteristic $\kappa^{-3}$ slope. This means both terms make significant and complementary contributions to the accuracy and physical consistency of the simulation.}

\subsection{Consistency of closure model performance \label{sec:consistency}}

\begin{figure}
    \centering
\includegraphics[width = 0.48\textwidth]{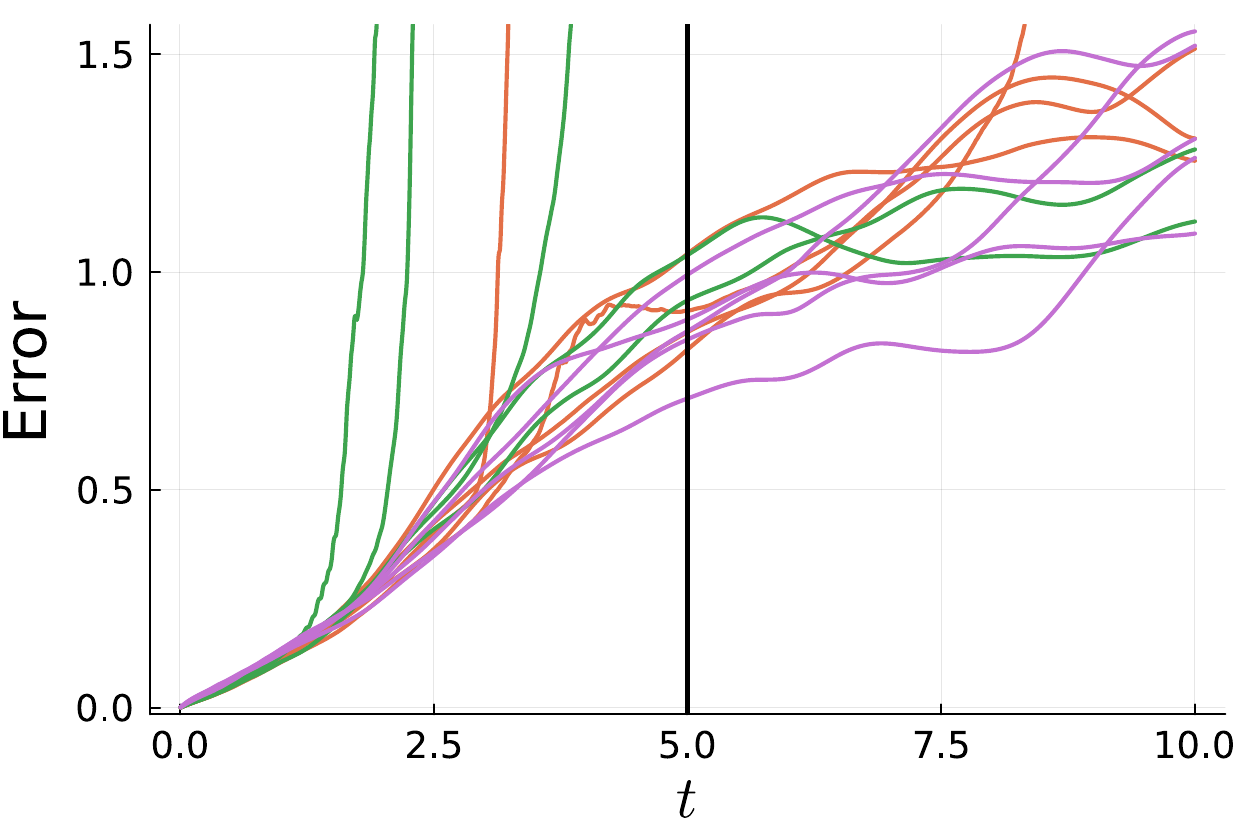}
\includegraphics[width = 0.48\textwidth]{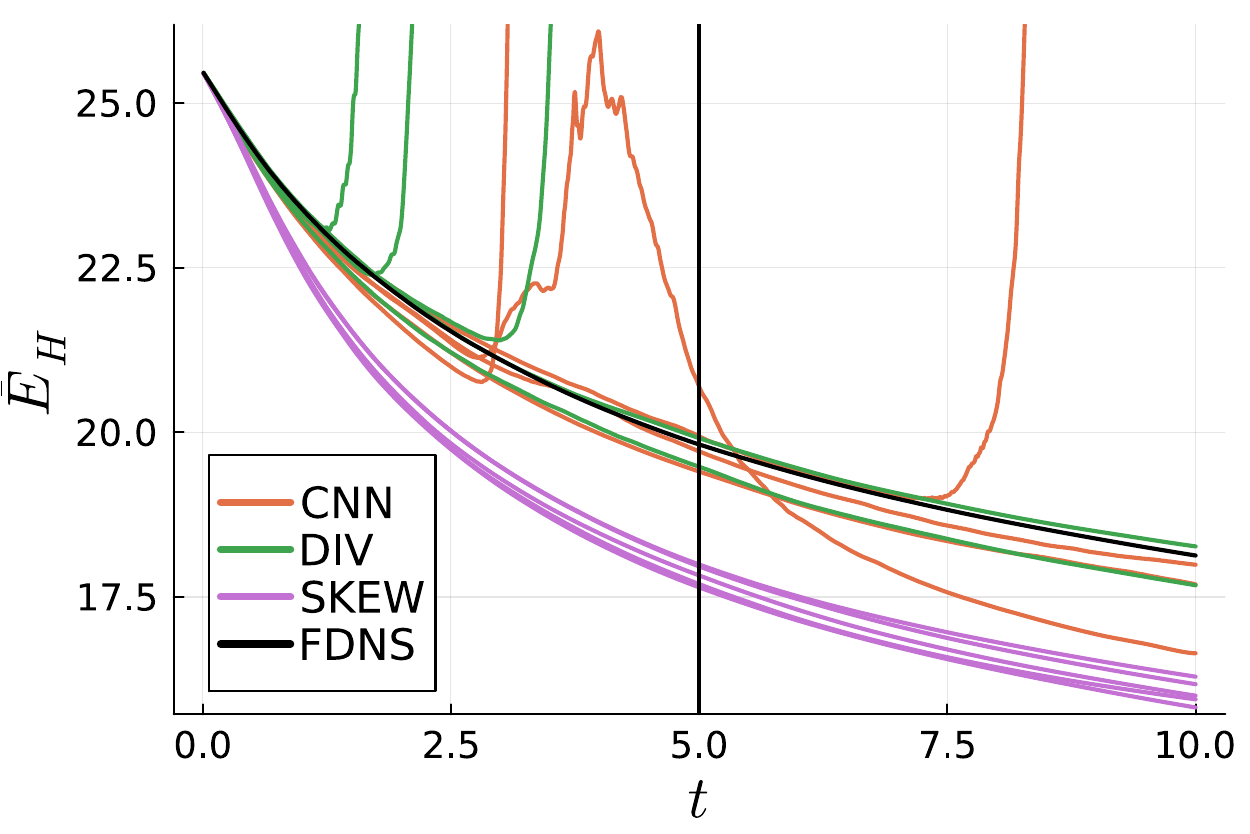}
\includegraphics[width = 0.48\textwidth]{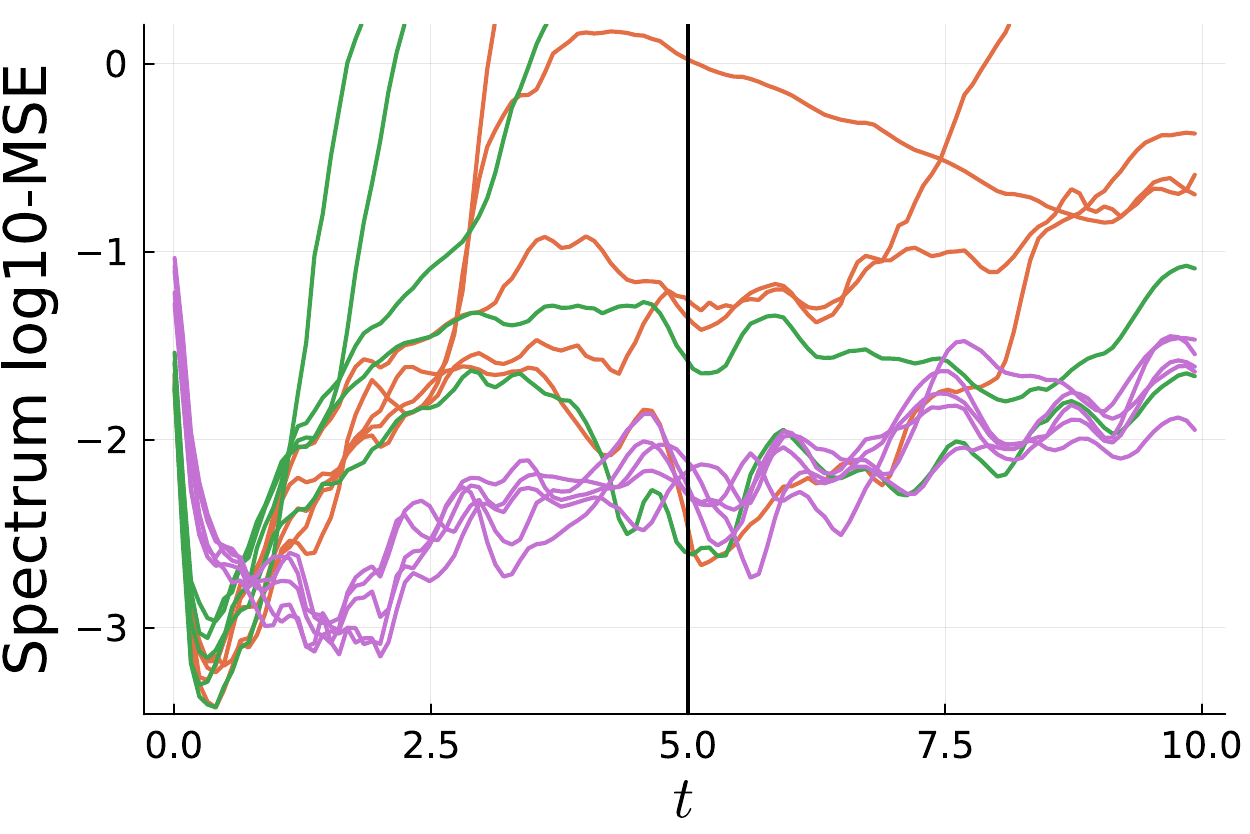}
\caption{(Top-left) Error over time for each closure model in the ensemble of five. (Top-right) Resolved energy trajectories for each closure model in the ensemble of five. (Bottom) Error of the energy spectrum over time calculated by computing the spectrum in $\log_{10}$ space, then computing the \gls{MSE} with respect to the \gls{FDNS} spectrum, and finally reporting the $\log_{10}$ of this value. These trajectories are also depicted for each closure model in the ensemble of five. The black vertical line indicates $t=5$. Everything to the right of this line corresponds to extrapolation in time. }\label{fig:ensemble}
\end{figure}

Training neural networks is inherently random, due to the selection of mini-batches, initialization of the weights, etc. This is why we instantiate multiple replicas of each network to fully assess their potential as a closure model. For this purpose, we train an ensemble of five replicas for each neural network architecture. This allows us to evaluate the consistency of the training procedure in terms of producing good closure models. Before evaluating the networks, we ensured no convergence issues occurred during training. The resulting error and energy trajectories, for each neural network, evaluated on the decaying turbulence test case, are depicted in Figure \ref{fig:ensemble}.
Regarding the plain \gls{CNN} architecture, we observe that two out of five networks result in unstable simulations, and for \acrshort{DIV}, three out of five. Hence, it is worth pointing out that, therefore, simply retraining exactly the same machine-learning architecture can be the difference between a stable simulation and a numerical failure, highlighting the fickle nature of these methods. For \gls{SKEW}, no ensemble members became unstable. 

Regarding the error trajectories, we observe similar performances for all the networks that remained stable, whereas the energy trajectories were best reproduced by some of the unconstrained machine learning closures. \R{rev_3_M_4_b}\revthree{However, to evaluate the build-up of numerical noise and physical consistency, we computed the error in energy spectrum during the simulation. This is also depicted in Figure \ref{fig:ensemble}.} Here we find that the \gls{SKEW} architecture consistently outperforms the other architectures. From this, we conclude that our \gls{SKEW} neural network architecture is not only guaranteed to be stable, but also consistently produces physical results, without a build-up of numerical noise.

\subsection{Kolmogorov flow}

\begin{figure}
    \centering
\includegraphics[width = 1\textwidth]{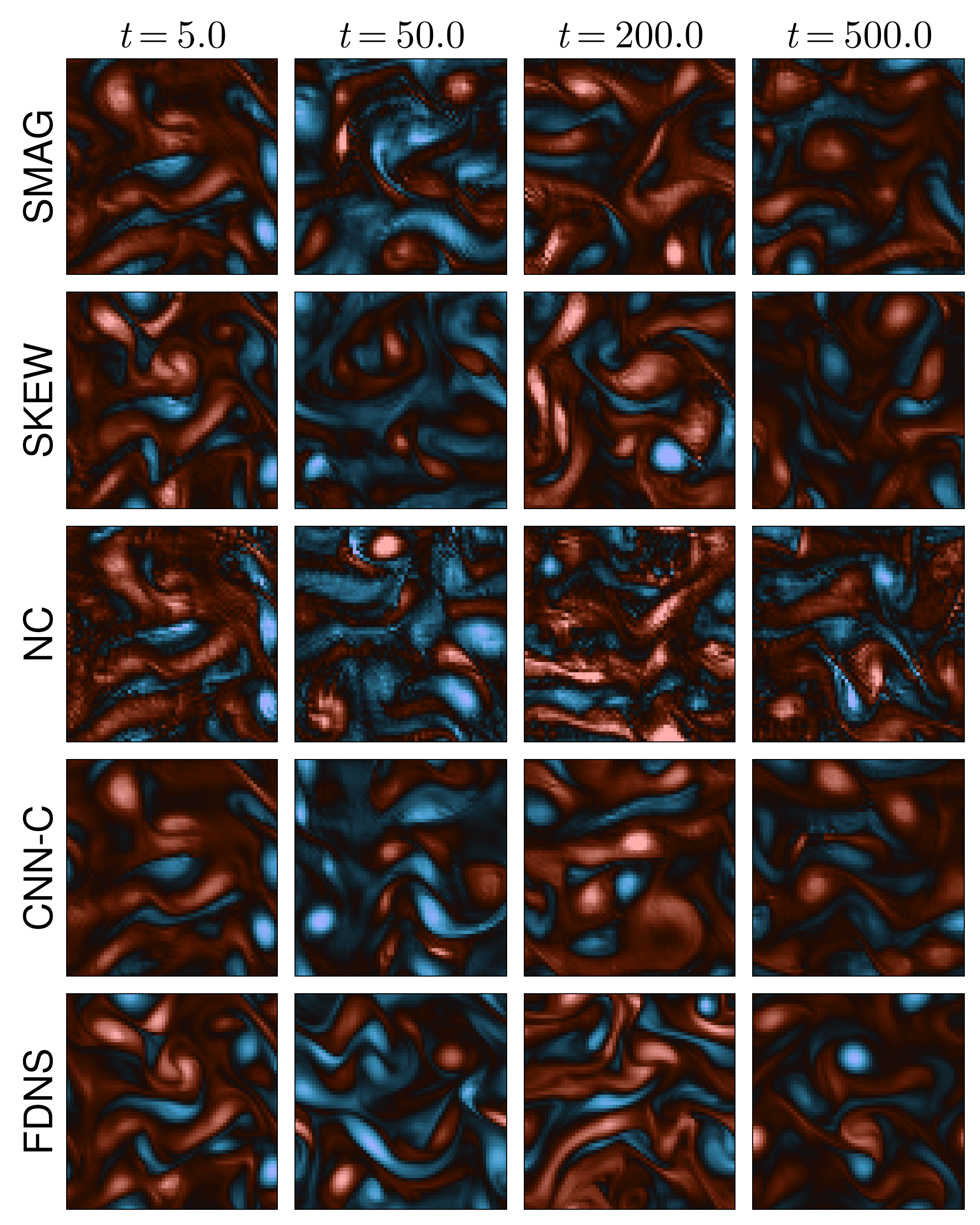}

\caption{Vorticity fields at each point in time for each of the closure models on a $64 \times 64$ grid. Simulations correspond to the Kolmogorov flow test case.}\label{fig:heatmaps_KF}
\end{figure}

\begin{figure}
    \centering

\includegraphics[width = 0.48\textwidth]{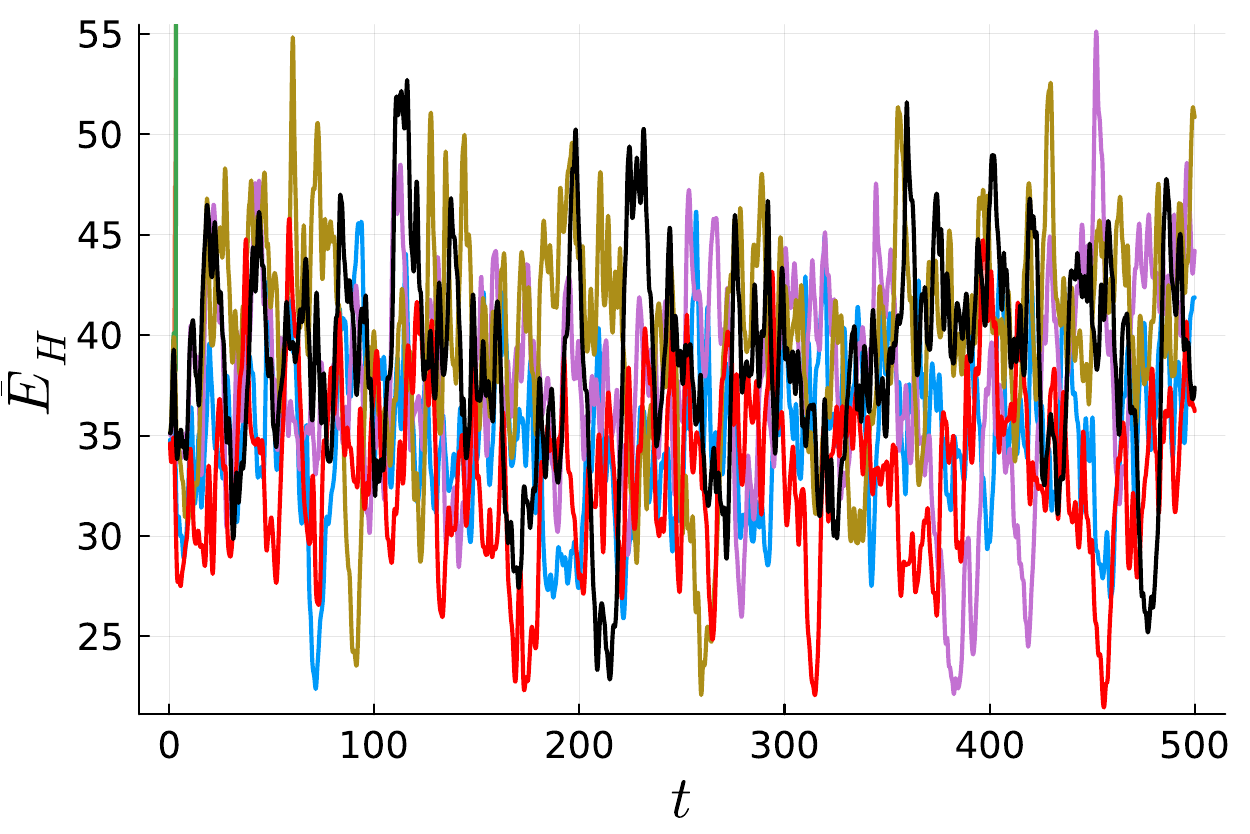}
\includegraphics[width = 0.48\textwidth]{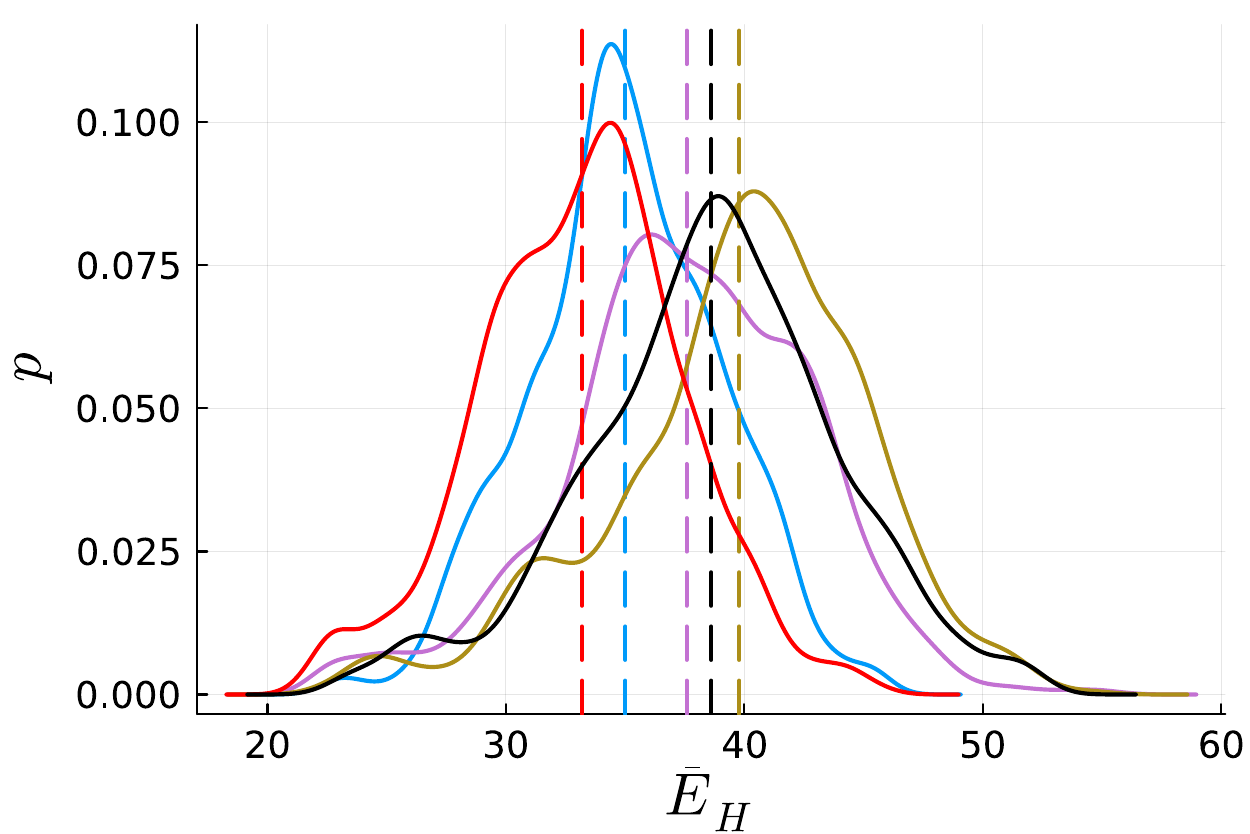}
\includegraphics[width = 0.48\textwidth]{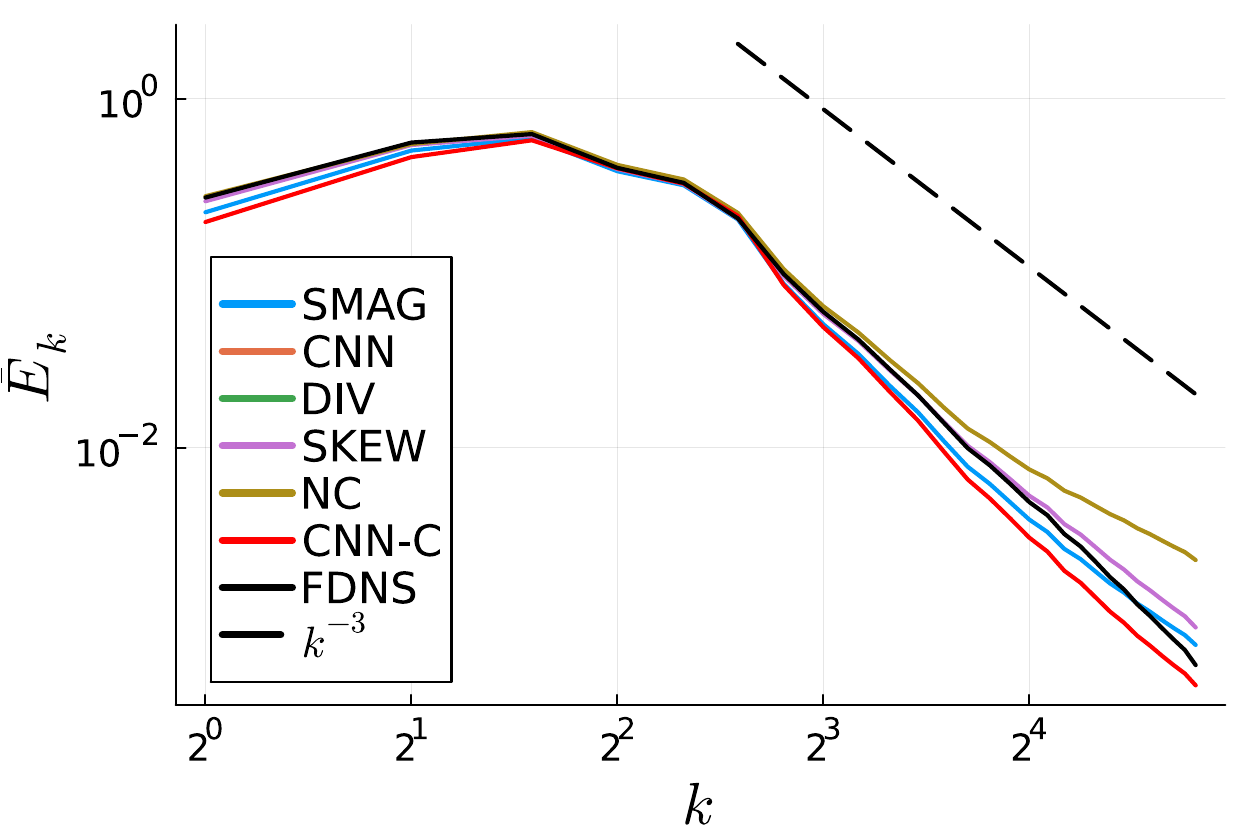}
\caption{(Top-left) Resolved energy trajectories for each of the closure models in the Kolmogorov flow test case. (Top-left) Corresponding distributions of the resolved energy for the entire simulation. Dashed vertical lines correspond to the mean. (Bottom) Energy spectra were obtained by first computing the energy spectrum for each snapshot and then computing the average value for each wavenumber. Both the \gls{CNN} and \acrshort{DIV} closure models resulted in unstable simulations, so their spectrum is omitted from the figures.}\label{fig:trajectories_KF}
\end{figure}

Next, we aim to evaluate the long-term performance of the closure models and their extrapolation capabilities. To do this, we require a different test case, as decaying turbulence from the previous test case eventually decays to zero. We therefore consider Kolmogorov flow, with the same viscosity $\nu = \frac{1}{1000}$ and periodic domain $\Omega = [-\pi,\pi]\times [-\pi,\pi]$. Kolmogorov flow is characterized by the following forcing:
\begin{equation}
    \mathbf{f}(\mathbf{u},\mathbf{x},t) =\begin{bmatrix}
       \sin(4y) \\ 0
    \end{bmatrix} - 0.1\mathbf{u},
\end{equation}
and is often used to evaluate machine learning closure models \cite{Kochkov,Agdestein_2025,shankar2024differentiableturbulenceclosurepartial}. The machine learning models are not retrained for this test case. This means the models will have to extrapolate from the decaying turbulence training data to a test case that includes forcing. To initialize the simulation, we first carry out a \gls{DNS} on a $2048 \times 2048$ grid until $t=25$, starting from initial condition \eqref{eq:init_cond}. This serves as a warm-up of the system, such that it reaches a statistical equilibrium. The final velocity field then serves as an initial condition to evaluate the closure models. We then simulate for 500 model time units starting from this initial condition. This is a significant extrapolation, as the training data was for the interval $t \in [0,5]$ and for a test case without forcing. 
In addition to the previously introduced closure models, we apply backscatter clipping to the trained \gls{CNN} to obtain a stable closure model \R{rev_3_M_3_c}\revthree{by removing backscatter}. This is done by projecting the \gls{CNN} output on an eddy-viscosity model. From this, we obtain an eddy-viscosity value $\nu_t^{\text{clipping}}(\mathbf{x})$ such that the \gls{CNN} output is matched as close as possible in the $L_2$-norm. Negative values for $\nu_t^{\text{clipping}}(\mathbf{x})$ are then set to zero to provide stability. The clipping procedure is described in \cite{beck1}. We will refer to this closure model with the acronym \acrshort{CNN-C}.
Vorticity snapshots from the simulations are depicted in Figure \ref{fig:heatmaps_KF}.
Snapshots for \gls{CNN} and \acrshort{DIV} are not depicted, as these simulations become unstable quickly after their initialization, see Figure \ref{fig:trajectories_KF}. For \gls{NC} we observe a lot of numerical noise, whereas in \gls{SMAG} (to a lesser extent), \gls{SKEW}, and \acrshort{CNN-C} this is smoothed out. Regarding the snapshot at $t=5$, we find the filtered \gls{DNS} results are most closely matched by \gls{SKEW}.

To make a more thorough comparison, we consider the resolved energy trajectories, along with an average energy spectrum for the simulation, see Figure \ref{fig:trajectories_KF}. 
The first thing we observe from the resolved energy trajectories is that the unconstrained machine learning approaches both become unstable at the start of the simulation. 
The remaining closure models remain stable, as they are strictly dissipative. To make a statistical comparison of the resulting flow fields, we consider the distribution of the resolved energy for the entire simulation and the average energy spectra. Here we find that \gls{SMAG} and \acrshort{CNN-C} are too dissipative, as the distributions are shifted to the left with respect to \gls{FDNS}. For 
\acrshort{CNN-C}, this phenomenon is reported in \cite{clipping_and_instability_and_dissipation_CNN_Guan_2022}. \R{rev_3_m_14}\revthree{Both \gls{NC} and \gls{SKEW} seem to give a good prediction of both the mean and shape of the distribution.}

Looking at the energy spectra, we find that \gls{NC} indeed produces numerical noise, looking at the energy in the large wavenumbers. In addition, we find that \gls{SKEW} performs the best in the low and intermediate wavenumbers, while \gls{SMAG} performs the best in the high wavenumbers. Overall, we find that for this extrapolation test case \gls{SKEW} performs, at worst, as well as \gls{SMAG}. From this, we conclude that \gls{SKEW} is both stable and accurate, and is capable of extrapolating to different test cases, without being retrained.

\subsection{\R{rev_2_M_3}\revtwo{Comparison to the dynamic Smagorinsky model}}

\revtwo{To provide a comparison between our \gls{SKEW} architecture and a more robust physics-based closure model, we consider the dynamic Smagorinsky model. The formulation and underlying theory of this model are described in \cite{germano_1992}. The main idea is that the Smagorinsky constant in \eqref{eq:vt} is determined dynamically and depends on space, time, and the resolved velocity field, i.e.\ $C_s = C_s(\mathbf{x},t,\bar{\mathbf{u}})$.}

\revtwo{The dynamic procedure builds on the assumption that the subgrid-scale stresses are self-similar across filter scales and that the same eddy-viscosity ansatz holds at both the grid filter and a larger test filter. By enforcing consistency between these two levels, the model computes $C_s$ from the resolved flow itself rather than prescribing it a priori. This allows the model to reduce dissipation in laminar or well-resolved regions while increasing it where subgrid activity is strong. Its main strength is therefore that it provides dissipation only where it is needed, making it significantly less dissipative than the standard Smagorinsky model.}

\begin{figure}
    \centering
    \includegraphics[width=0.48\textwidth]{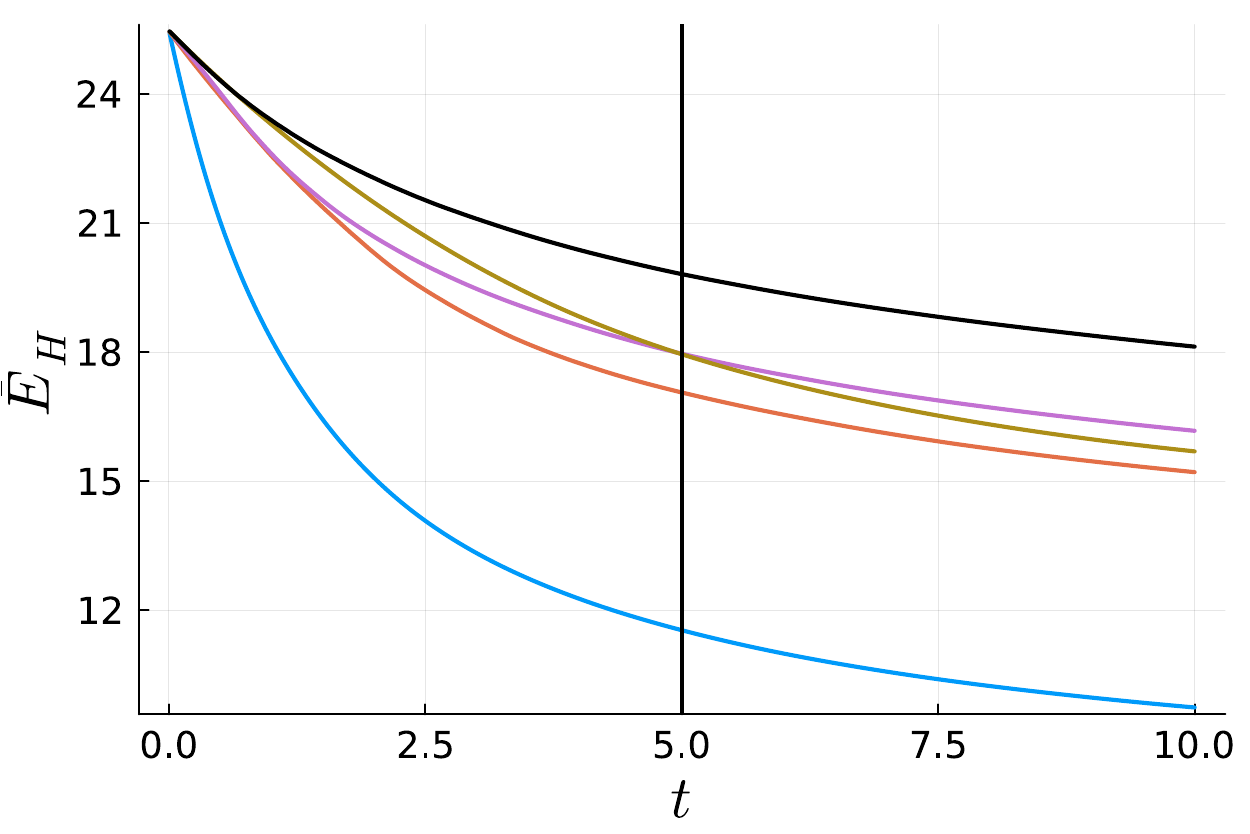}
    \includegraphics[width=0.48\textwidth]{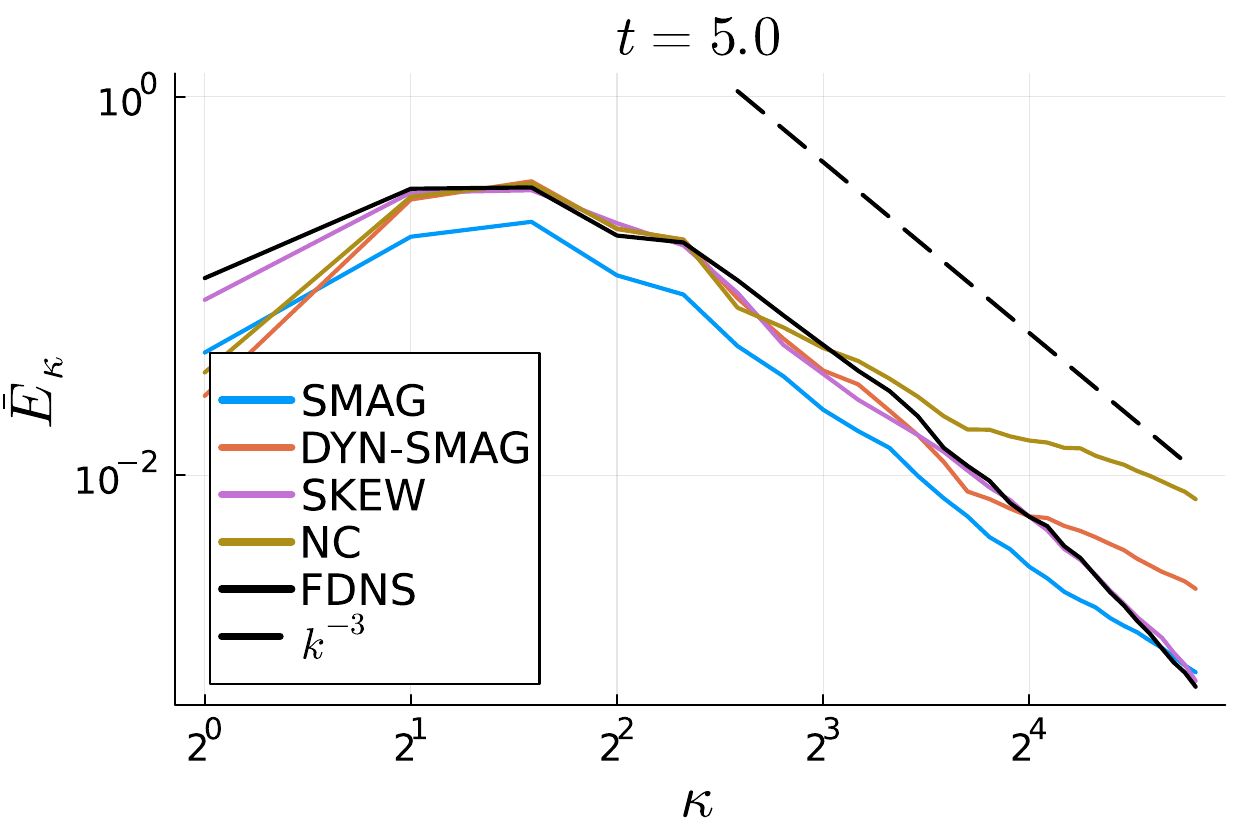}
    \caption{(Left) Resolved energy trajectories for the decaying turbulence test case, where DYN-SMAG corresponds to the dynamic Smagorinsky model. The black vertical line indicates $t=5$. Everything to the right of this line corresponds to extrapolation in time. (Right) Energy spectra halfway through the decaying turbulence simulation at $t=5$.}
    \label{fig:dyn_smag}
\end{figure}

\revtwo{The resulting energy trajectory and energy spectrum for the dynamic Smagorinsky model in the decaying turbulence test case are shown in Figure \ref{fig:dyn_smag}, together with the results from our \gls{SKEW} architecture.}

\revtwo{For the energy trajectory, we find that the dynamic Smagorinsky model is indeed less dissipative than the standard \acrfull{SMAG}. However, it remains slightly more dissipative than our \gls{SKEW} architecture, which matches the \gls{FDNS} result the closest. Regarding the energy spectrum, the dynamic Smagorinsky model provides a substantial improvement over \gls{SMAG} and \gls{NC}. Nonetheless, it still underpredicts energy at low wavenumbers and overpredicts energy at high wavenumbers, whereas our \gls{SKEW} architecture does not exhibit these deficiencies.}

\subsection{\revtwo{Limitations of Machine Learning Closure Models}}

\revtwo{
Note that a neural network is not strictly necessary to enforce a skew-symmetric framework. Traditional calibration or data assimilation approaches could be used, which typically assume a fixed functional form and tune a limited set of coefficients to \gls{DNS} data. In contrast, a neural network provides a flexible functional representation capable of capturing complex, non-linear dependencies of the closure term on local flow features, dependencies that are difficult to express or calibrate explicitly using fixed parametric forms. While the skew-symmetric structure enforces energy conservation, the entries of the skew-symmetric matrix (as well as those of the dissipative term) are learned functions of the flow state, allowing the model to adapt dynamically to local flow conditions rather than relying on globally calibrated coefficients.}

\revtwo{Finally, it should be noted that fundamental challenges exist when employing machine learning within the context of \gls{LES} (or any physics-based simulation) \cite{coveney2016big}. Neural networks easily have millions (or more) tunable parameters with no physical meaning, are data-hungry, and act as black boxes, making them more complex than their physics-based counterparts. They are also prone to instabilities over long integration times, an issue that we have explicitly addressed in the present manuscript. Our proposed framework, therefore, represents a step towards embedding physical constraints (in this case, guaranteed energy conservation) directly into the architecture of a data-driven model. This combination of physical structure and data-driven flexibility embodies a key advantage of physics-informed machine learning: it ensures physical consistency while retaining the expressive power of modern machine-learning methods. That said, many of the aforementioned issues remain and require further study.}

\section{Conclusion}\label{sec:conclusion}

In this work, we started off by exploring the conservation laws inherent in the incompressible Navier-Stokes equations, specifically, mass, momentum, and energy conservation. We employed a discretization that preserves these laws in a discrete sense. To coarse-grain the simulation, we used a face-averaging filter, ensuring that the resulting coarse-grained velocity field continues to satisfy mass conservation. We then examined different approaches to modeling the commutator error introduced by coarse-graining. We first considered the Smagorinsky model, which is strictly dissipative, followed by more advanced machine learning approaches based on convolutional neural networks, capable of modeling backscatter. However, these unconstrained models lack stability guarantees. To address this limitation, we introduced our skew-symmetric neural architecture \cite{VANGASTELEN2024113003}. This architecture enforces stability while increasing model freedom from the negative-definite eddy-viscosity basis by introducing a skew-symmetric term. A change from our previous work \cite{VANGASTELEN2024113003} is that the subgrid-scale energy is no longer explicitly modeled. In addition, we tackled much more challenging 2D turbulence applications, as opposed to simple 1D test cases considered in \cite{VANGASTELEN2024113003}. Based on offline analysis on the training data, we hypothesized that dissipative closure models, such as the skew-symmetric architecture we introduce, are likely to perform well for the considered test cases.

We evaluated our closure models across three coarse-graining factors, from a $2048\times2048$
grid down to resolutions of $128 \times 128$, $64\times 64$, and $32\times 32$. The closure models were tested in a decaying turbulence simulation with an initial condition different from those in the training data and over an extended simulation time. We found that none of the models performed well at the largest coarse-graining factor (from $2048\times 2048$ down to $32 
\times 32$). Initially, the unconstrained machine learning models provided promising results, even outperforming our skew-symmetric model in kinetic energy predictions. However, numerical errors accumulated, leading to instability. Trajectory fitting, i.e., training the closure models to reproduce the filtered \gls{DNS} solution, alone is therefore not enough to guarantee stable closure models. In contrast, our skew-symmetric model remained stable throughout, albeit at the cost of a larger dissipation rate. Despite the increased dissipation, we argue that stability is a worthwhile trade-off. In addition, our skew-symmetric architecture outperformed the Smagorinsky model in this test case and reproduced the expected $k^{-3}$ in the energy spectrum for 2D turbulence. Furthermore, our architecture also outperformed the dynamic Smagorinsky model, a widely used and well-established physics-based closure model, in this test case.

To account for the inherent randomness in training neural networks, we trained five instances of each model with different weight initializations and mini-batch selections. All instances of the skew-symmetric model remained stable and accurate, while conventional machine learning models exhibited significant instabilities. This highlights the improved consistency of our approach in training robust closure models.

We further assessed the models on the Kolmogorov flow test case, which was not represented in the training data. The unconstrained machine learning models again suffered from instabilities, whereas our skew-symmetric closure model successfully captured the correct energy spectrum when averaged over time. In this regard, it performed comparably to the Smagorinsky model, which proved to be well-suited to this test. We also applied backscatter clipping to the trained \gls{CNN}. This resulted in a stable simulation; however, it did not perform better than the skew-symmetric architecture or the Smagorinsky model. 

Overall, our results demonstrate that the skew-symmetric architecture we introduced here significantly enhances the stability of machine-learning-based closure models, albeit at the cost of increased dissipation. We believe this work represents a step toward enabling long-time simulations with machine learning closures.

For future work, several directions merit exploration. The treatment of boundary conditions, particularly the padding used for convolutional neural networks, requires careful attention. Extending our approach to unstructured grids is another promising avenue, where graph neural networks could provide a useful framework \cite{graph_NN_belbute2020combining}. Finally, introducing an additional energy source to counteract the absence of backscatter from the skew-symmetric architecture could be beneficial, though careful clipping mechanisms would be needed to prevent instabilities. Explicitly modeling the subgrid-scale energy, as in \cite{VANGASTELEN2024113003}, is another logical extension. % but requires some methodological changes to be extended beyond 1D.

\section*{CRediT authorship contribution}

\textbf{T. van Gastelen:} Conceptualization, Methodology, Software, Writing - original draft. \textbf{W. Edeling:} Writing - review \& editing. \textbf{B. Sanderse:} Conceptualization, Methodology, Writing - review \& editing, Funding acquisition.

\section*{Data availability}

The code used to generate the training data and the implementation of the neural networks will be made available on request.

\section*{Acknowledgements}

This publication is part of the project ``Unraveling Neural Networks with Structure-Preserving Computing” (with project number OCENW.GROOT.2019.044
of the research programme NWO XL which is financed by the Dutch Research
Council (NWO)). Part of this publication is funded by Eindhoven University of Technology. Finally, we thank the reviewers for their feedback, enhancing the quality of the article.

\section*{Declaration of generative AI and AI-assisted technologies in the writing process}

During the preparation of this work the author(s) used ChatGPT in order to improve language and grammar. After using this tool/service, the author(s) reviewed and edited the content as needed and take(s) full responsibility for the content of the published article.
%% main text
%%

%% The Appendices part is started with the command \appendix;
%% appendix sections are then done as normal sections

%% For citations use: 
%%       \cite{<label>} ==> [1]

%%

%% If you have bib database file and want bibtex to generate the
%% bibitems, please use
%%
%%  \bibliographystyle{elsarticle-num} 
%%  \bibliography{<your bibdatabase>}

%% else use the following coding to input the bibitems directly in the
%% TeX file.

%% Refer following link for more details about bibliography and citations.
%% https://en.wikibooks.org/wiki/LaTeX/Bibliography_Management
\printnoidxglossaries

\bibliographystyle{elsarticle-num}
\bibliography{references}

\appendix
\clearpage

\section{Physical structure of the Navier-Stokes equations}\label{sec:structure_appendix}

The Navier-Stokes equations, see \eqref{eq:NS}, represent a set of fundamental physical laws, namely: conservation of mass, momentum, and energy (for zero dissipation). These are collectively referred to as the physical structure of the system.
Conservation of mass can easily be shown by computing the change of mass in a volume $\mathcal{V}$ due to the flux across the surface $\mathcal{S}$. This reads
\begin{equation}
\oint_\mathcal{S} \mau \cdot \text{d}\mathbf{s} = \int_\mathcal{V}  \nabla \cdot \mau  \text{d}V = 0,
\end{equation}
where we used the divergence theorem to write the surface integral as a volume integral. 

Conservation of momentum is also straightforward to derive: The momentum is defined as
\begin{equation}
    \mathbf{P} = \int_\Omega \mau\text{d}\Omega,
\end{equation}
where $\Omega$ is the spatial domain.
The change in momentum is computed as follows:
\begin{equation}
    \begin{split}
        \frac{\text{d}\mathbf{P}}{\text{d}t} =& \int_\Omega \frac{\partial\mau}{\partial t}\text{d}\Omega \\ =& - \oint_{\partial \Omega}  \mau\mau^T \cdot \text{d}\mathbf{s} + \oint_{\partial \Omega} p \text{d}\mathbf{s}+ \nu \oint_{\partial \Omega}  \nabla \mau \cdot \text{d}\mathbf{s}  +\int_\Omega \mathbf{f}\text{d}\Omega  \\ =&\int_\Omega \mathbf{f}\text{d}\Omega,
    \end{split}
\end{equation}
where we filled in \eqref{eq:NS} and rewrote most of the terms to boundary integrals. These disappear on periodic domains. This means that the momentum in each direction only changes due to the body force. 

Finally, the total (kinetic) energy in the system is defined as
\begin{equation}
    E = \frac{1}{2} \int_\Omega \mau \cdot \mau \text{d}\Omega.
\end{equation}
Using the product rule, we can write the change in energy as
\begin{equation}
    \frac{\text{d}E}{\text{d}t} = \int_\Omega \mau \cdot \frac{\partial\mau}{\partial t} \text{d}\Omega = \int_\Omega \mau \cdot (- \nabla \cdot (\mau \mau ^T)  -\nabla p + \nu \nabla^2 \mau + \mathbf{f}) \text{d}\Omega.
\end{equation}
The pressure and friction contributions can be simplified using integration by parts:
\begin{align}
    \int_\Omega - \mau \cdot  \nabla p  \text{d}\Omega = \int_\Omega p  (\nabla \cdot \mau)      \text{d}\Omega = 0, \\
     \int_\Omega \nu \mau \cdot  \nabla^2 \mau  \text{d}\Omega =  -\int_\Omega \nu  ||\nabla \mau||_2^2  \text{d}\Omega,
\end{align}
where we used the fact $\nabla \cdot \mau =0$ and that the boundary contributions cancel on periodic domains.

To find the energy contribution for the convective term, we start off by rewriting it using the product rule for outer products:
\begin{equation}
    \nabla \cdot (\mau \mau ^T) = (\mau \cdot \nabla) \mau + (\nabla \cdot \mau) \mau = (\mau \cdot \nabla) \mau,
\end{equation}
where the second term vanishes due to divergence freeness.
Next, we rewrite the term using a vector calculus identity:
\begin{equation}
    (\mau \cdot \nabla) \mau = \frac{1}{2}\nabla (\mau \cdot \mau) - \mau \times (\nabla \times \mau).
\end{equation}
We then take the inner product with $\mathbf{u}$ to obtain
\begin{equation}
    \mau \cdot \nabla \cdot (\mau \mau ^T) = \mau \cdot \frac{1}{2}\nabla (\mau \cdot \mau) - \mau \cdot (\mau \times (\nabla \times \mau)) = \mau \cdot \frac{1}{2}\nabla (\mau \cdot \mau),
\end{equation}
where the second term cancels due to the fact that the cross product between two vectors is orthogonal to both of these vectors. Using the product rule starting from $\nabla \cdot (\mau (\mau \cdot \mau))$ this can be rewritten to
\begin{equation}
    \mau \cdot \frac{1}{2}\nabla (\mau \cdot \mau) = \frac{1}{2}\nabla \cdot (\mau (\mau \cdot \mau)) - \frac{1}{2} (\mau \cdot \mau) (\nabla \cdot \mau ) = \frac{1}{2}\nabla \cdot (\mau (\mau \cdot \mau)),
\end{equation}
which is in divergence form. Once again, we used the fact that $\mathbf{u}$ is divergence-free to simplify this expression. This term integrates to zero:
\begin{equation}
    \int_\Omega \mau \cdot \nabla \cdot (\mau \mau ^T)\text{d}\Omega = \int_\Omega \frac{1}{2}\nabla \cdot (\mau (\mau \cdot \mau))\text{d}\Omega =  \oint_{\partial \Omega} \frac{1}{2} (\mau (\mau \cdot \mau)) \cdot \text{d}\mathbf{s} =  0,
\end{equation}
due to the divergence theorem and the fact that $\Omega$ is a periodic domain. The change in energy is finally written as
\begin{equation}
    \frac{\text{d}E}{\text{d}t} = -\int_\Omega \nu  ||\nabla \mau||_2^2  \text{d}\Omega + \int_\Omega \mau \cdot \mathbf{f}  \text{d}\Omega,
\end{equation}
which means the energy is always decreasing in the absence of forcing.

\section{Stucture-preserving finite volume discretization}\label{sec:discretization_appendix}

For the employed finite volume discretization, presented in \cite{harlow1965numerical}, the physical structure of the Navier-Stokes equations is preserved in a discrete sense.
Discretely, the total momentum and energy are approximated as
\begin{align}
    \mathbf{P}_h &=  \boldsymbol{\mathbbm{1}}_h \boldsymbol{\Omega}_h\mauh, \\
    E_h &= \frac{1}{2} \mauh^T\boldsymbol{\Omega}_h\mauh.
\end{align}
The change in momentum for this discretization is given by
\begin{equation}
\begin{split}
    \frac{\text{d}\mathbf{P}_h}{\text{d}t} &=  \boldsymbol{\mathbbm{1}}_h\boldsymbol{\Omega}_h\frac{\text{d}\mauh}{\text{d}t} \\ &=   \boldsymbol{\mathbbm{1}}_h(- \mathbf{C}_h(\mauh)\mauh  - \mathbf{G}_h \mathbf{p}_h + \nu \mathbf{D}_h \mauh + \boldsymbol{\Omega}_h\mathbf{f}_h) \\ &=   \boldsymbol{\mathbbm{1}}_h\boldsymbol{\Omega}_h\mathbf{f}_h,
\end{split}
\end{equation}
as the discrete operators are carefully constructed such that the column vectors sum up to zero.
This means momentum conservation is satisfied by this discretization. Using the product rule, we obtain the change in energy as
\begin{equation}\begin{split}
    \frac{\text{d} E_h}{\text{d}t} &= \mauh^T\boldsymbol{\Omega}_h\frac{\text{d}\mauh}{\text{d}t} \\ &=   \mauh^T(- \mathbf{C}_h(\mauh)\mauh  - \mathbf{G}_h \mathbf{p}_h + \nu \mathbf{D}_h \mauh + \boldsymbol{\Omega}_h\mathbf{f}_h) \\ &=- \mauh^T \mathbf{Q}_h^T\mathbf{Q}_h \mauh + \mauh^T\boldsymbol{\Omega}_h\mathbf{f}_h = - ||\mathbf{Q}_h \mauh||_2^2 + \mauh^T\boldsymbol{\Omega}_h\mathbf{f}_h,
\end{split}    
\end{equation}
where we used the fact that the diffusion operator $\mathbf{D}_h$ can be Cholesky decomposed as $-\mathbf{Q}_h^T\mathbf{Q}_h$ \cite{benjamin_thesis,podbenjamin}. The convective contribution disappears due to the skew-symmetry of the discrete operator:
\begin{equation}
    \mathbf{C}_h(\mauh) = - \mathbf{C}_h^T(\mauh) \quad \rightarrow \quad \mauh^T \mathbf{C}_h(\mauh)\mauh = -\mauh^T \mathbf{C}_h^T(\mauh)\mauh = 0,
\end{equation}
where we used the symmetry of the inner product. Note that the skew-symmetry of the convection operator is only true for a divergence-free $\mathbf{u}_h$ \cite{Agdestein_2025}.
The pressure term disappears because the discretization satisfies $\mathbf{G}_h = -\mathbf{M}_h^T$ \cite{podbenjamin}. Writing the energy contribution we obtain:
\begin{equation}
    -\mauh^T \mathbf{G}_h\mathbf{p}_h = \mauh^T \mathbf{M}_h^T \mathbf{p}_h = \mathbf{p}^T_h\mathbf{M}_h\mauh = 0,
\end{equation}
where we used the divergence-free constraint on the velocity field. 

\section{Pressure projection}\label{sec:pressure_projection_appendix}

The divergence freeness can also be written as a projection of the \gls{PDE} discretization \eqref{eq:sd-NS} on a divergence-free basis \cite{Agdestein_2025}. To see this, we compute the divergence of the non-pressure related terms in \eqref{eq:sd-NS} as
\begin{equation}
    \mathbf{M}_h \boldsymbol{\Omega}_h^{-1} (-\mathbf{C}_h(\mauh)\mauh + \nu \mathbf{D}_h \mauh + \boldsymbol{\Omega}_h\mathbf{f}_h ) = \mathbf{M}_h \boldsymbol{\Omega}_h^{-1} \mathbf{m}_h(\mauh),
\end{equation}
where we grouped the different terms into $\mathbf{m}_h(\mauh)$.
The purpose of the gradient of the pressure is to remove this divergence from the \gls{RHS} of \eqref{eq:sd-NS}. To obtain the pressure that achieves this, we solve the following linear system:
\begin{equation}
\begin{split}
    &\mathbf{M}_h \boldsymbol{\Omega}_h^{-1} \mathbf{m}_h(\mauh) - \mathbf{M}_h\boldsymbol{\Omega}_h^{-1}\mathbf{G}_h\mathbf{p}_h = \mathbf{0} \quad \rightarrow \\ &\mathbf{M}_h \boldsymbol{\Omega}_h^{-1} \mathbf{m}_h(\mauh) = \mathbf{M}_h\boldsymbol{\Omega}_h^{-1}\mathbf{G}_h\mathbf{p}_h \quad  \rightarrow \\  &\mathbf{p}_h =  (\mathbf{M}_h\boldsymbol{\Omega}_h^{-1}\mathbf{G}_h)^{-1}\mathbf{M}_h \boldsymbol{\Omega}_h^{-1} \mathbf{m}_h(\mauh).
\end{split}
\end{equation}
By filling this in to \eqref{eq:sd-NS} we obtain
\begin{equation}
\begin{split}
    &\boldsymbol{\Omega}_h\frac{\text{d}\mauh}{\text{d}t} =  \mathbf{m}_h(\mauh) - \mathbf{G}_h (\mathbf{M}_h\boldsymbol{\Omega}_h^{-1}\mathbf{G}_h)^{-1}\mathbf{M}_h \boldsymbol{\Omega}_h^{-1} \mathbf{m}_h(\mauh) \quad \rightarrow \\
    &\boldsymbol{\Omega}_h\frac{\text{d}\mauh}{\text{d}t} = \underbrace{(\mathbf{I} - \mathbf{G}_h (\mathbf{M}_h\boldsymbol{\Omega}_h^{-1}\mathbf{G}_h)^{-1}\mathbf{M}_h \boldsymbol{\Omega}_h^{-1} )}_{:=\mathcal{P}_h}\mathbf{m}_h(\mauh) \quad \rightarrow \\  &\boldsymbol{\Omega}_h\frac{\text{d}\mauh}{\text{d}t} =\mathcal{P}_h\mathbf{m}_h(\mauh),
\end{split}
\end{equation}
where $\mathcal{P}_h \in \mathbb{R}^{2N \times 2N}$ projects $\mathbf{m}_h(\mauh)$ onto a divergence free basis. This transforms the discretized \gls{PDE} into a single equation. Note that in practice, we typically solve for the pressure rather than performing the projection in this manner. However, this formulation is more convenient for closure modeling. 

\section{Motivation behind skew-symmetric architecture}\label{app:motivation}

To motivate the proposed neural network architecture we consider a simple 1D system for which the equation is unknown. The neural network is tasked with predicting the evolution of this system. The system consists of some quantity $w(x,t)$ which evolves on a periodic domain $x \in \Omega$. We know it satisfies the following conservation laws:
\begin{align}\label{eq:cons_law_w_1}
    \int_\Omega \frac{\text{d}w}{\text{d}t} \text{d}x &= 0, \\ \label{eq:cons_law_w_2}
    \int_\Omega w\frac{\text{d}w}{\text{d}t} \text{d}x &= 0,
\end{align}
which resemble momentum and energy conservation in the Navier-Stokes equations. $w(x,t)$ is discretized on a finite volume grid, such that $w(x_i,t)\approx \text{w}_i(t)$, where $x_i$ is the center of finite volume cell $\Omega_i$. The grid contains $N$ grid cells such that the discrete solution is described by the state vector $\mathbf{w}(t) \in \mathbb{R}^N$. We consider the case in which the evolution equation for $w(x,t)$ is unknown; however, we do have data for $w(x,t)$ at different points in space and time. The challenge is finding the \gls{RHS} for $\frac{\text{d}\mathbf{w}_h}{\text{d}t}$, with \eqref{eq:cons_law_w_1} and \eqref{eq:cons_law_w_2} being satisfied discretely, i.e.
\begin{align}\label{eq:cons_law_w_1_discr}
    \mathbf{1}_h^T\mathbf{\Omega}_h \frac{\text{d}\mathbf{w}_h}{\text{d}t} = 0, \\ \label{eq:cons_law_w_2_discr}
    \mathbf{w}_h^T\mathbf{\Omega}_h \frac{\text{d}\mathbf{w}_h}{\text{d}t} = 0,
\end{align}
are satisfied.
As an ansatz for the \gls{RHS} we use our proposed skew-symmetric neural network architecture
\begin{equation}\label{eq:rhs_w}
    \mathbf{\Omega}_h \frac{\text{d}\mathbf{w}_h}{\text{d}t} \approx  \underbrace{(\boldsymbol{\Delta}_c\text{diag}(\mathbf{k})\boldsymbol{\Delta}_f - \boldsymbol{\Delta}_f^T \text{diag}(\mathbf{k})\boldsymbol{\Delta}_c^T)}_{=:\mathcal{Y}}\mathbf{w}_h,
\end{equation}
where $\boldsymbol{\Delta}_c,\boldsymbol{\Delta}_f \in \mathbb{R}^{N \times N}$ are central and forward difference stencils, respectively, such that $(\boldsymbol{\Delta}_c\mathbf{w}_h)_i = \text{w}_{h,i+1} - \text{w}_{h,i-1}$ and $(\boldsymbol{\Delta}_f\mathbf{w}_h)_i = \text{w}_{h,i+1} - \text{w}_{h,i}$. Note that these are specific choices for $\mathcal{B}_1$ and $\mathcal{B}_2$. During testing, we found that predefining the convolutions in $\mathcal{B}_1$ and $\mathcal{B}_2$ resulted in poor performance of the closure model. \R{rev_3_M_2_b}\revthree{However, we choose to do so here to ease the analysis.}  Note that \eqref{eq:cons_law_w_1_discr} is satisfied due to $\boldsymbol{\Delta}_c$ and $\boldsymbol{\Delta}_f$ being in the nullspace of $\mathbf{1}_h$ and \eqref{eq:cons_law_w_2_discr} is satisfied due to the skew-symmetry of $\mathcal{Y}$. We are free to choose $\mathbf{k} \in \mathbb{R}^N$ without violating the conservation laws. In our case, it is represented by the output of a \gls{CNN}, such that $\mathbf{k} = \mathbf{k}(\mathbf{w}_h,\theta) \in \mathbb{R}^N$. 
Writing out the matrix-vector product in \eqref{eq:rhs_w} gives us
\begin{equation*}
\begin{split}
    \mathcal{Y}\mathbf{w}_h &= \begin{bmatrix}
     & & \ddots & & & & \\
        & & \ddots & \text{k}_{i-1} & & & \\
      \ddots & \ddots & \ddots & -\text{k}_{i-1} - \text{k}_{i}  & \ddots & & \\ &
      - \text{k}_{i-1} & \text{k}_{i} + \text{k}_{i-1}  & 0  & -\text{k}_{i} - \text{k}_{i+1}  & \text{k}_{i+1} & \\
        & & \ddots & \text{k}_{i+1} + \text{k}_{i}  & \ddots & \ddots &  \ddots \\
        & & & - \text{k}_{i+1}  & \ddots & & \\ & & & & \ddots& & \\
    \end{bmatrix} \begin{bmatrix}
    \vdots  \\ \vdots \\ \vdots \\ \text{w}_{h,i} \\ \vdots \\ \vdots
      \\   \vdots  \end{bmatrix} \\ &= 
    \begin{bmatrix}
         & \ddots & & & \\
        \ddots & \ddots & \text{w}_{h,i+1} - \text{w}_{h,i}  & &  \\
        & \text{w}_{h,i-1} - \text{w}_{h,i-2}  & \text{w}_{h,i-1} - \text{w}_{h,i+1} & \text{w}_{h,i+2} - \text{w}_{h,i+1} &  \\
        & & \text{w}_{h,i} - \text{w}_{h,i-1} & \ddots & \ddots  \\
        & & & \ddots &  \\
    \end{bmatrix}
      \begin{bmatrix}
    \vdots \\ \vdots \\ \text{k}_{i} \\ \vdots \\ \vdots
        \end{bmatrix},
\end{split}
\end{equation*}
where it is clear that for both matrices the columns lie in the nullspace of $\mathbf{1}_h$. This gives us two perspectives: the first is that of a highly parameterized skew-symmetric matrix multiplied by the solution vector $\mathbf{w}_h$. The second is that of the parameterized vector $\mathbf{k}(\mathbf{w}_h,\theta)$ being multiplied by a matrix for which each column lies in the nullspace of $\mathbf{w}_h$. Therefore, by training $\mathbf{k}(\mathbf{w}_h,\theta)$ on reference data we effectively learn a highly parameterized\R{rev_3_M_2_c}\revthree{, and non-linear,} momentum and energy conserving discretization. \R{rev_3_M_2_d}\revthree{This formulation allows the neural network to freely explore the energy-conserving vector space. By choosing $k_i = 1/(2h)$ and $k_{i-1} = k_{i+1} = 0$, where $h$ is the grid spacing, we simply obtain a central difference approximation of the first derivative, which corresponds to linear advection. The skew-symmetric term can therefore be viewed as highly parameterized advection.}

For dissipative systems, i.e. \eqref{eq:cons_law_w_2} is negative, we require an additional term in our parameterized discretization, as introduced in section \ref{sec:neg_def_term}. This extends \eqref{eq:rhs_w} to
\begin{equation}
    \mathbf{\Omega}_h \frac{\text{d}\mathbf{w}_h}{\text{d}t} \approx  \mathcal{Y}\mathbf{w}_h \underbrace{- \boldsymbol{\Delta}_f^T\text{diag}(\mathbf{q})^2\boldsymbol{\Delta}_f}_{=:\mathcal{Z}} \mathbf{w}_h,
\end{equation}
where $\mathbf{q} = \mathbf{q}(\mathbf{w}_h,\theta) \in \mathbb{R}^N$ is also an output of the \gls{CNN}. In this case we choose $\mathcal{B}_3 = \boldsymbol{\Delta}_f$. \R{rev_3_M_2_e}\revthree{This specific choice is once again made to ease the analysis, whereas in the actual neural network $\mathcal{B}_3$ is a parameterized operator.}
Writing out the corresponding stencil gives:
\begin{equation*}
\begin{split}
    (\mathcal{Z}\mathbf{w}_h)_i &= \text{q}_i^2(\mathbf{w}_h,\theta)(w_{h,i+1} - w_{h,i}) + \text{q}_{i-1}^2(\mathbf{w}_h,\theta)(w_{h,i-1} - w_{h,i}).
\end{split}
\end{equation*}
By choosing $\text{q}_i = 1/h$, we obtain a second-order approximation of the second derivative. Our dissipative term can therefore be thought of as highly parameterized and non-linear diffusion.

\section{Training of the neural networks}\label{sec:convergence_appendix}

\begin{figure}
    \centering
\includegraphics[width = 0.48\textwidth]{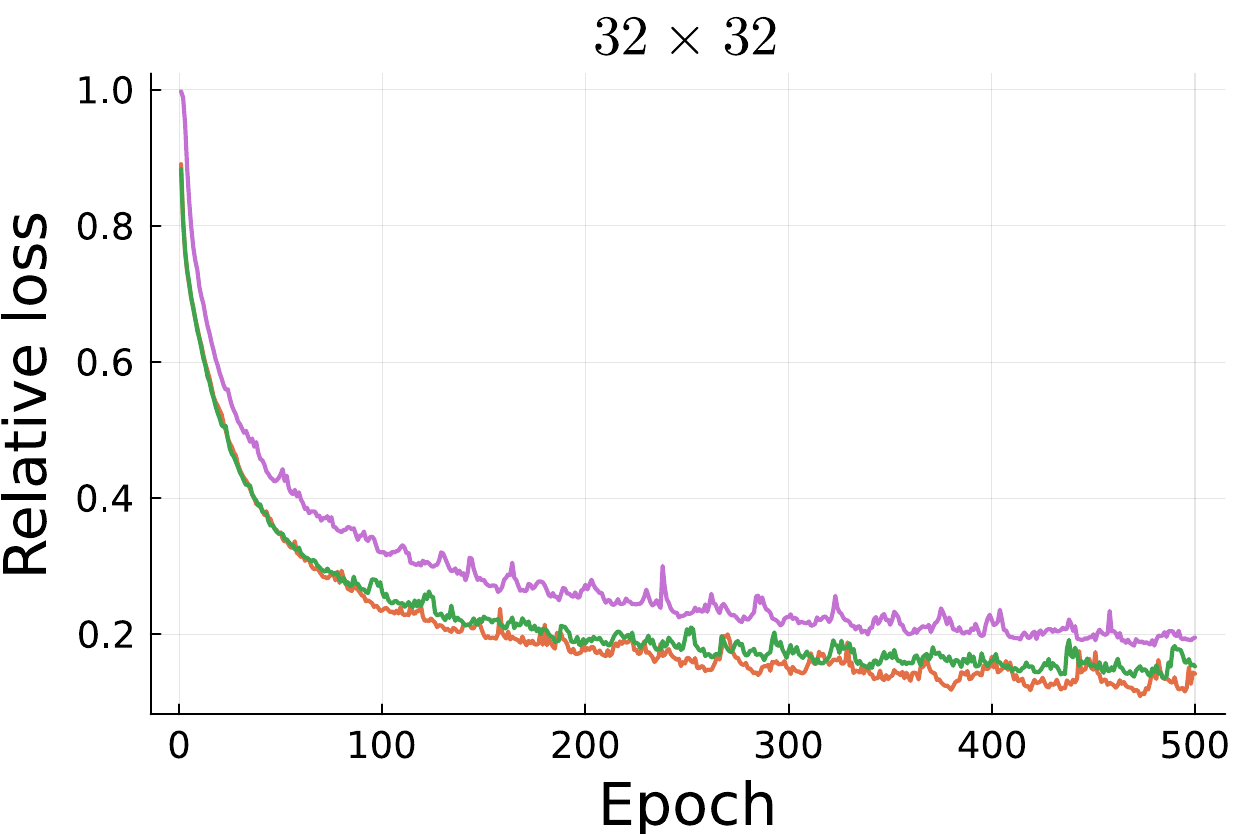}
\includegraphics[width = 0.48\textwidth]{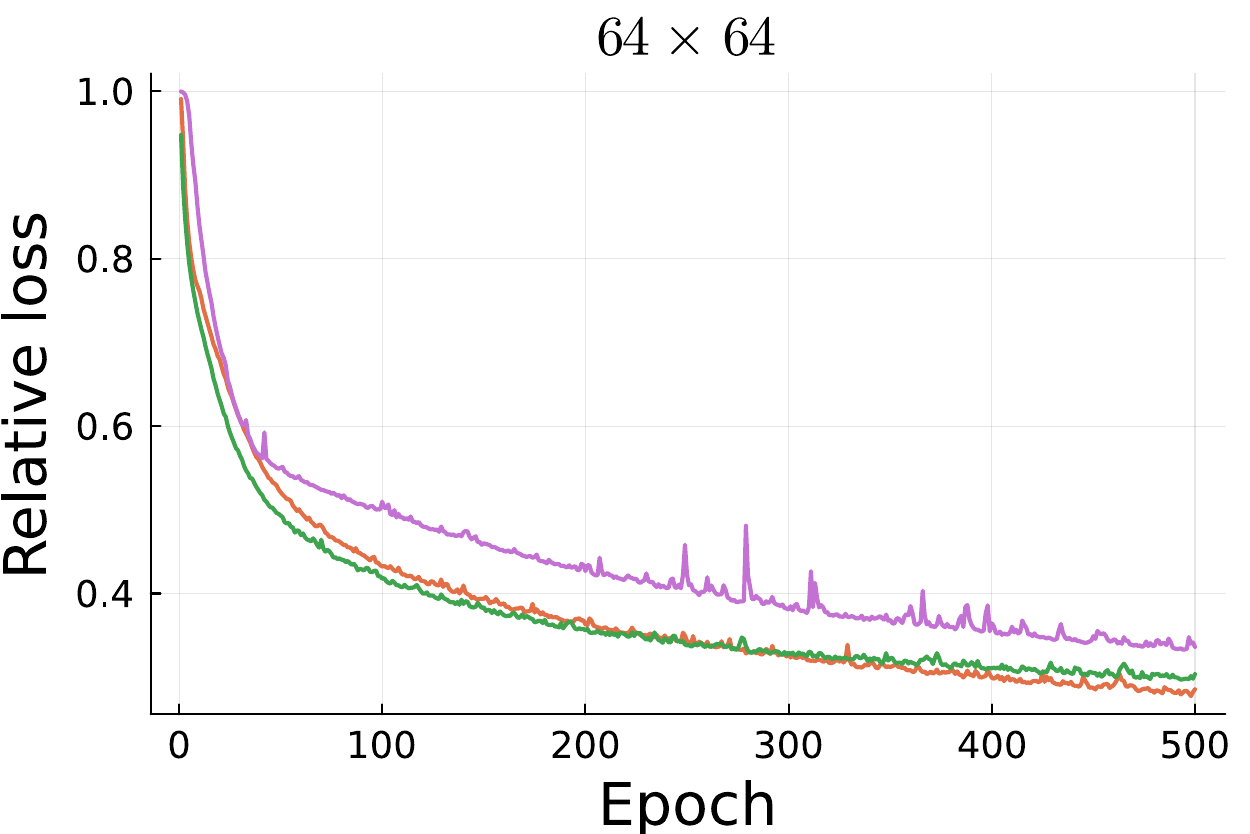}
\includegraphics[width = 0.48\textwidth]{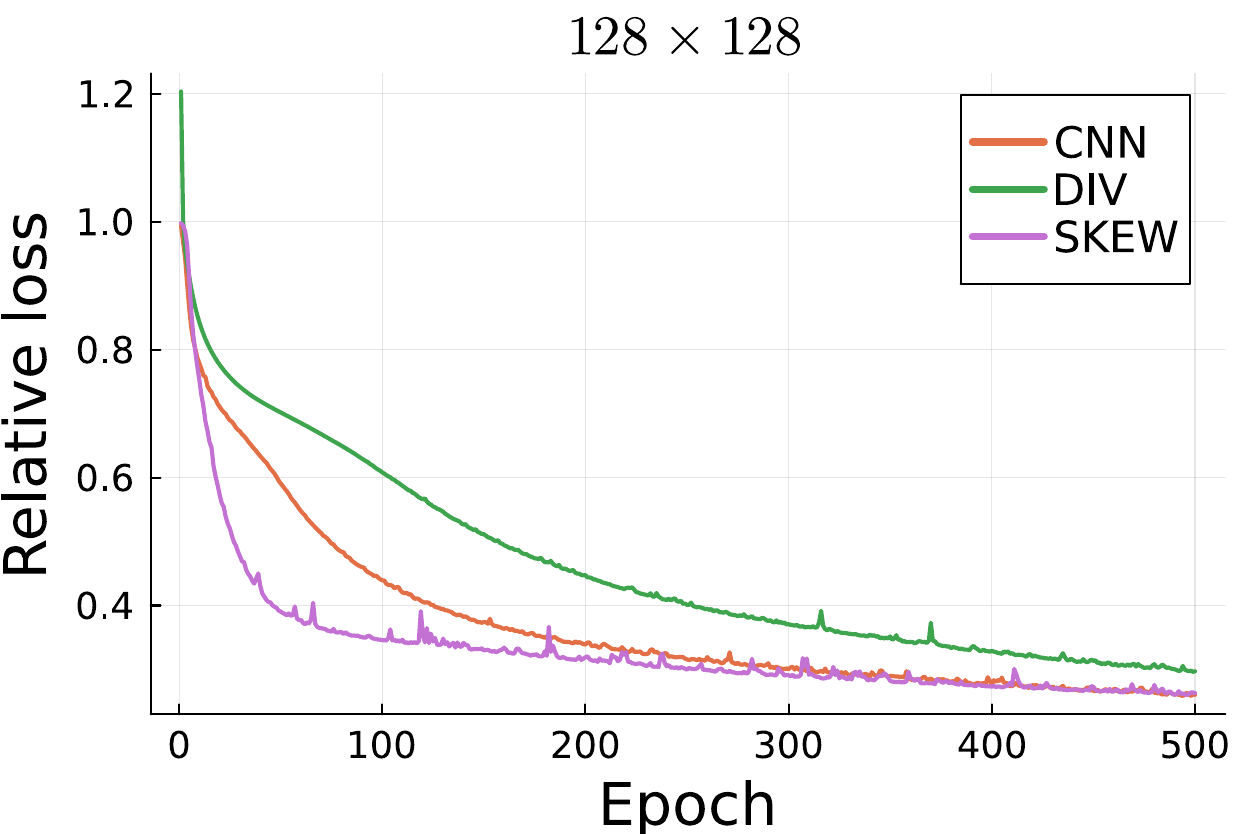}
\caption{Relative loss, see \eqref{eq:traj_loss}, with respect to \gls{NC} for each of the different closure models and coarse-graining factors.}\label{fig:NN_convergence}
\end{figure}

In Figure \ref{fig:NN_convergence} we depict the loss of the different closure model architectures with respect to \gls{NC}.
The choice of hyperparameters is discussed in section \ref{sec:experimental_setup}.

\clearpage

\section{Vorticity fields}\label{sec:plots_appendix}

\begin{figure}
    \centering
\includegraphics[width = 1.0\textwidth]{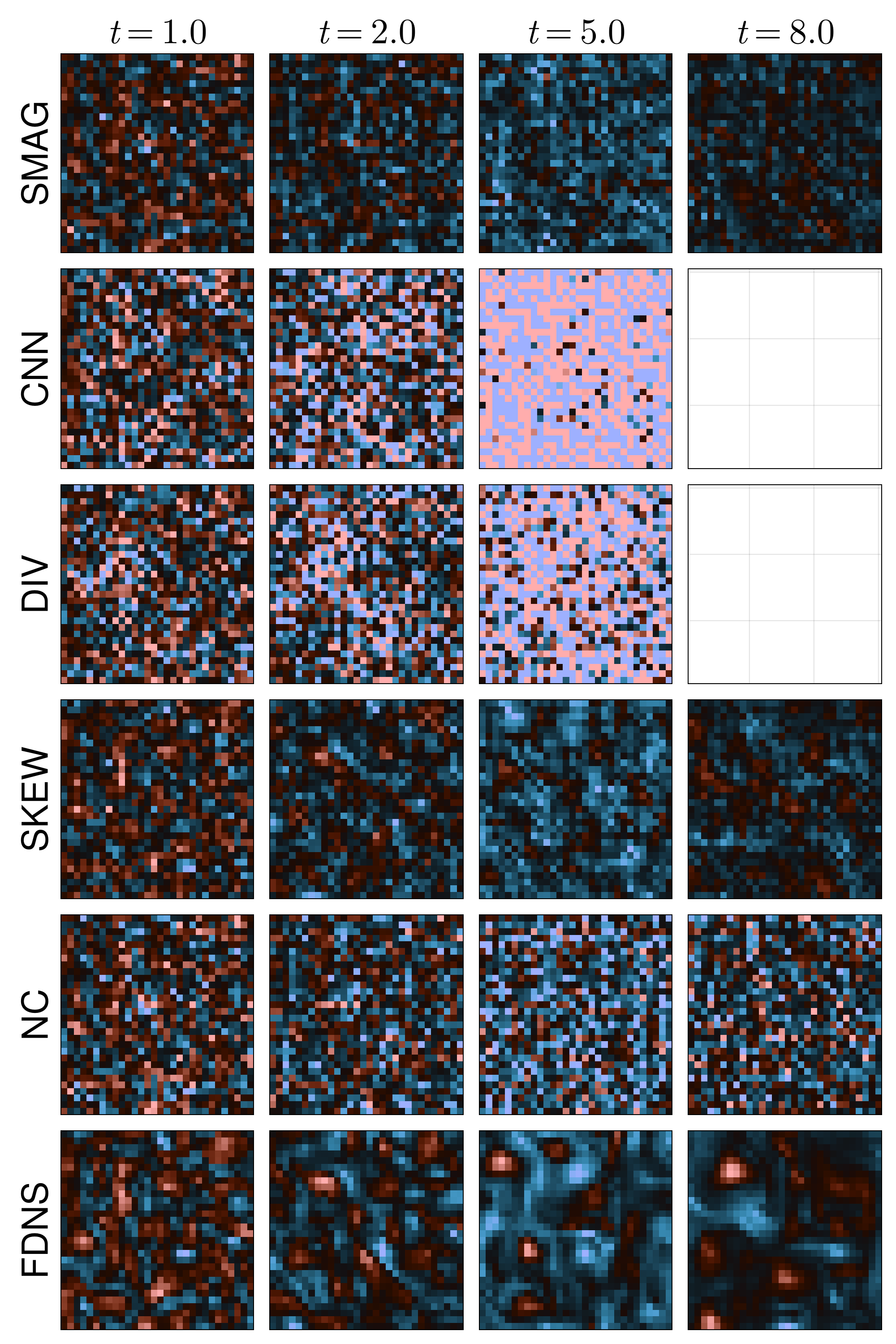}

\caption{Vorticity fields at each point in time for each of the closure models on a $32 \times 32$ grid. Simulations correspond to the decaying turbulence test case. Blank boxes indicate an unstable simulation.}\label{fig:32_32}
\end{figure}

\begin{figure}
    \centering
\includegraphics[width = 1.0\textwidth]{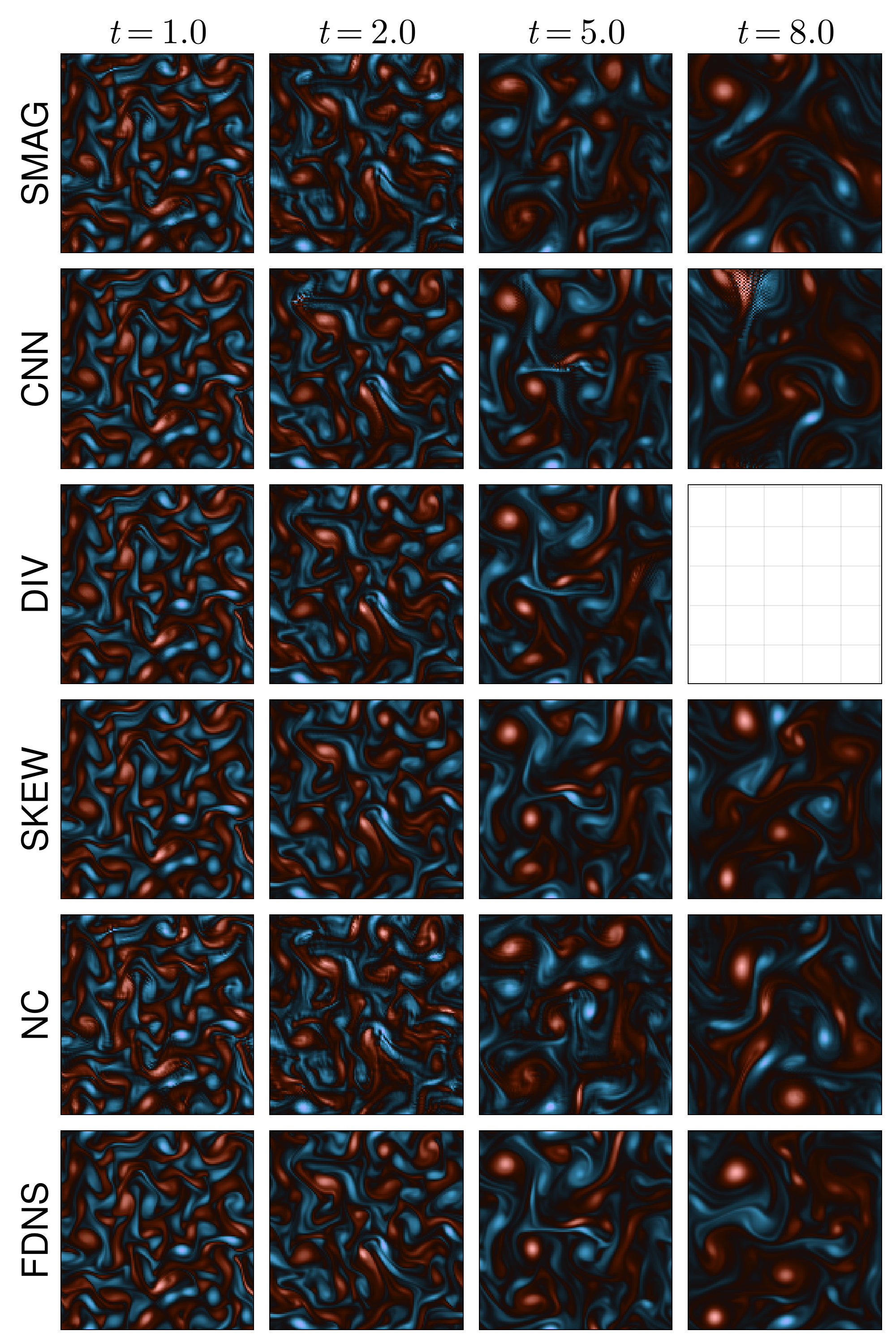}

\caption{Vorticity fields at each point in time for each of the closure models on a $128 \times 128$ grid. Simulations correspond to the decaying turbulence test case. Blank boxes indicate an unstable simulation.}\label{fig:128_128}
\end{figure}

In Figure \ref{fig:32_32} and Figure \ref{fig:128_128} we display the vorticity fields of the decaying turbulence simulation, using the trained closure models.

\end{document}